# Towards a Theoretical Understanding of Word and Relation Representation

*Carl S Allen*

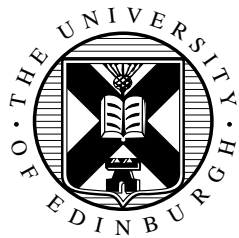

Doctor of Philosophy
School of Informatics
University of Edinburgh

2022

# Abstract

Representing words by vectors of numbers, known as *word embeddings*, enables computational reasoning over words and is foundational to automating tasks involving natural language. For example, by crafting word embeddings so that *similar* words have similar valued embeddings, often thought of as nearby points in a *semantic space*, word similarity can be readily assessed using a variety of metrics. In contrast, judging whether two words are similar from more common representations, such as their English spelling, is often impossible (e.g. *cat/feline*); and to predetermine and store all similarities between all words is prohibitively time-consuming, memory intensive and subjective. As a succinct means of representing words – or, perhaps, the *concepts* that words themselves represent – word embeddings also relate to information theory and cognitive science.

Numerous algorithms have been proposed to learn word embeddings from different data sources, such as large text corpora, document collections and "knowledge graphs" – compilations of facts in the form ⟨*subject entity, relation, object entity*⟩, e.g. ⟨*Edinburgh, capital_of, Scotland*⟩. The broad aim of these algorithms is to capture information from the data in the components of each word embedding that is useful for a certain task or suite of tasks, such as detecting sentiment in text, identifying the topic of a document, or predicting whether a given fact is true or false. In this thesis, we focus on word embeddings learned from text corpora and knowledge graphs.

Several well-known algorithms learn word embeddings from text on an *unsupervised* (or, more recently, *self-supervised*) basis by learning to predict *context* words that occur around each word, e.g. *word2vec* (Mikolov et al., 2013a,b) and *GloVe* (Pennington et al., 2014). The parameters of word embeddings learned in this way are known to reflect word co-occurrence statistics, but how they capture semantic meaning has been largely unclear.

Knowledge graph representation models learn representations both of *entities*, which include words, people, places, etc., and binary *relations* between them. Representations are typically learned by training the model to predict known true facts of the knowledge graph in a *supervised* manner. Despite steady improvements in the accuracy with which knowledge graph representation models are able to predict facts, both seen and *unseen* during training, little is understood of the latent structure that allows them to do so.

This limited understanding of how latent semantic structure is encoded in the geometry of word embeddings and knowledge graph representations makes a principled direction for improving their performance, reliability or interpretability unclear. To address this:

1. we theoretically justify the empirical observation that particular *geometric* relationships between word embeddings learned by algorithms such as *word2vec* and *GloVe* correspond to *semantic* relations between words; and

2. we extend this correspondence between semantics and geometry to the entities and relations of knowledge graphs, providing a model for the latent structure of knowledge graph representation linked to that of word embeddings.

We first give a probabilistic explanation for why word embeddings of *analogies* – phrases of the form "man is to king as woman is to queen" – often appear to approximate a parallelogram. This "analogy phenomenon" has generated much intrigue since word embeddings are not trained to achieve it, yet it allows many analogies to be "solved" simply by adding and subtracting their embeddings, e.g. $\boldsymbol{w}_{queen} \approx \boldsymbol{w}_{king} - \boldsymbol{w}_{man} +$ $\boldsymbol{w}_{woman}$. Similar probabilistic rationale is given to explain how semantic relations such as *similarity* and *paraphrase* are encoded in the relative geometry of word embeddings.

Lastly, we extend this correspondence, between semantics and embedding geometry, to the specific relations of knowledge graphs. We derive a hierarchical categorisation of relation *types* and, for each type, identify the notional geometric relationship between *word embeddings* of related entities. This gives a theoretical basis for relation representation against which we can contrast a range of knowledge graph representation models. By analysing properties of their representations and their relation-by-relation performance, we show that the closer the agreement between how a model represents a relation and our theoretically-inspired basis, the better the model performs. Indeed, a knowledge graph representation model inspired by this research achieved state-of-the-art performance (Balažević et al., 2019b).

# Lay Summary

Humans effortlessly perform many tasks involving words with which they are familiar, e.g. we know instantly whether two words refer to *similar* concepts, or if their meanings are *related* in other ways, such as being opposites, or one being a part of the other (e.g. *wheel/car*). Indeed, we perform far more complex word-based tasks, such as forming coherent sentences to describe scenarios, or translating from one language to another. Despite our ability at these tasks, often from childhood, they are typically hard to automate. This is in good part because the way in which we *represent* words, e.g. their spelling in a given language, does not facilitate these tasks. For example, deciding if two words have similar meaning cannot generally be achieved by considering their spelling alone, e.g. *van/truck*, or *eagle/falcon*.

As a result, to automate language tasks, words are instead represented by vectors (or lists) of numbers, known as *embeddings*, that can be thought of similar to co-ordinates in space. The idea is that by assigning numerical values to words, their semantic properties, e.g. whether a noun/verb/adjective/etc., should be identifiable from the location of their embedding; and semantic relationships between words, e.g. *is_similar_to* or *is_bigger_than*, are reflected in the *relative* locations of their embeddings.

Several methods have been developed to produce word embeddings with these semantic properties, in particular so that embeddings of similar words are near one another, so that identifying similar words simply requires calculating the distance between their embeddings. An intriguing finding is that these embeddings can often be (approximately) added and subtracted intuitively, e.g. **man**+**royal** = **king** or **US**+**currency** = **dollar**, and an *analogy* question like "man is to king as woman is to ...?" is solvable by finding the closest embedding to **king** − **man** + **woman**, which turns out to be that of "queen". These embeddings were previously known to reflect statistics of the data they are learned from. We explain the described phenomena by considering interactions between those statistics to show how semantic relationships between words lead to spatial relationships between their embeddings.

Other methods have been developed to learn facts of the form ⟨*subject entity, relation, object entity*⟩, e.g. ⟨*Edinburgh, capital_of, Scotland*⟩. Collections of such facts are called *knowledge graphs* (or *knowledge bases*). As with word embeddings, the *entities* of knowledge graphs can be represented by vectors of numbers and the relation between them can be seen as a *transformation* or *mapping* from subject entity embeddings to object entity embeddings. By representing entities and relations numerically in this way, new facts can be predicted that were not in the original knowledge graph. To try to understand how this works, we extend the connection established between semantic relationships and the spatial arrangement of word embeddings. From this we propose templates of how embeddings are spatially arranged relative to one another for different types of semantic relations between words, as found in knowledge graphs.

By improving our understanding of how word embeddings and knowledge graph representations reflect the semantics of words in this thesis, we hope that improved embedding methods can be developed in future, e.g. that perform better or that are more interpretable. Indeed, a state-of-the-art method for representing knowledge graph entities and relations was developed in the course of this research.

# Acknowledgements

I would like to thank my supervisor Tim Hospedales for his enthusiasm and support throughout this thesis and bearing with the significant changes of course since our early discussions.

I am grateful for helpful comments, discussions and proofreads from many over the course of this work, particularly Iain Murray (my second supervisor), Michael Gutmann, Chris Williams, Rik Sarkar, Jonathan Mallinson, Ed Fincham, Artur Bekasov, Nick Hoernle and Henry Gouk, and for enlightening conversations with Nigel de Sousa and Ben Hartley regarding the "box". I would especially like to thank Ivan Titov for knowledgeable, constructive and positive comments along the way.

I am grateful to my collaborator Benedek Rozemberczki for his contagious enthusiasm for graph embeddings and dedicated hard work, and to friends from my programme, office and elsewhere who made the experience fun and memorable, despite half of it in lockdown. I am very lucky and grateful to have met my "collaborator" Ivana. We have worked together and mutually supported one another through many submissions, reviews and other dark moments. This PhD, and life, would not be the same without her. Lastly, I want to deeply thank my Mum (despite refusing to read this!) for her love and support over the years, not least during the rather belated and uncertain career change that led here.

This work was supported by the Centre for Doctoral Training in Data Science, funded by EPSRC (grant EP/L016427/1) and the University of Edinburgh.

# Declaration

I declare that this thesis was composed by myself, that the work contained herein is my own except where explicitly stated otherwise in the text, and that this work has not been submitted for any other degree or professional qualification except as specified.

*(Carl S Allen)*

To Anna-Jane, you said I should do it... turns out you were right.
I wish you were still here to see it.

# Table of Contents

# Chapter 1

# Introduction

Humans are able to perform many tasks involving words with which they are familiar with apparent ease, often from a young age, e.g. we generally know instantly whether two words refer to *similar* concepts, or if their meanings are *related* in other ways, such as being opposites, or one being a part of the other (e.g. *wheel/car*). Indeed, we can perform far more complex word-based tasks, such as forming coherent sentences to describe a real or imagined scenario, or translating from one language to another. Despite our ability at these tasks, they are typically non-trivial to automate. This is in good part because our familiar *representations* of words, in particular their spelling in a given language, are not well suited to such tasks. For example, even the relatively simple task of deciding whether words have similar meaning cannot be achieved in general by considering their spelling alone, e.g. *van/truck*, or *eagle/falcon*.

The prevailing approach to overcome this is to (re-)represent words by vectors of real numbers, known as *embeddings*, that are often thought of as co-ordinates in a *semantic space*. By assigning numerical values to words, it is found that semantic properties can be identified by rules, e.g. by classification models for part-of-speech tagging or named entity recognition; and semantic relationships between words can be encoded in numerical relationships between their embeddings, interpretable as geometric relationships in semantic space. These word embeddings can also be incorporated into more complex models, e.g. for automated text generation or machine translation.

The question then is which values, and how many of them (the dimensionality of the embedding), to assign to each word so that the desired semantic properties are present, i.e. how to create word embeddings. For many languages, the vast number of words suggests that this process itself is automated, which also avoids a multitude of subjective human judgements. Many algorithms have been proposed to generate word embeddings from different sources of data, most commonly large text corpora, collections of documents and "knowledge graphs". Text is readily available from web sources such as Wikipedia, document collections are commonly obtained from news sources and knowledge graphs (or knowledge bases) are curated lists of facts representing every day knowledge, in the form ⟨*subject entity, relation, object entity*⟩, e.g. ⟨*Edinburgh, capital_of, Scotland*⟩.

Word embeddings are often learned from text corpora as the parameters of a classification model trained to predict the *context words* observed around each word (e.g. Mikolov et al., 2013a,b; Pennington et al., 2014). Representations of entities and re-lations of knowledge graphs are likewise often learned within a classification model trained to identify true facts in the knowledge graph (e.g. Nickel et al., 2011; Yang et al., 2015; Bordes et al., 2014; Trouillon et al., 2016; Balažević et al., 2019c,b). A good number of algorithms exist in each case with impressive performance on tasks of interest. However, despite a few inroads (e.g. Levy and Goldberg, 2014b), both lack a firm mathematical basis for how *semantic* properties and relations of words/entities are captured in the *geometry* of their embeddings; in short, how these embeddings "work".

The existence of successful representation algorithms allows the broad question of *how to generate useful representations*, to be approached more concretely by analysing *what such algorithms learn*, and deciphering *why that is useful for semantic tasks*. In the case of knowledge graphs, it is not immediately clear how to determine what KGR models learn or how to mathematically model the latent structure of the data that they capture. However, certain word embedding models are known to capture specific statistical relationships in the data, leaving the question of how those statistics relate to semantics. We thus consider how the semantic relations *similarity, relatedness, paraphrase and analogy* manifest in observed geometric patterns between word embeddings. The identified correspondence between semantics and geometry is then extended to knowledge graphs on the premise that similar latent semantic structure may underpin both word embeddings and knowledge graphs, after all the same words/entities, with their same semantic meanings and relationships, can arise in either.

Beyond the natural scientific interest in establishing a firmer theoretical understanding of word/entity and relation representations, including intriguing properties such as the "analogy phenomenon" where word embeddings of analogies often approximate a parallelogram, other reasons for doing so include:

1. that it may foster algorithms producing embeddings that perform better on downstream tasks, are more interpretable aiding *explainability*, mitigate against unwanted *bias* in the data and/or enable *confidence* assessments of their predictions;

2. that it may have broader application since word embedding algorithms have been applied in many other domains, e.g. to represent members of social networks;

3. that it may extend to larger scale embeddings, such as of phrases, sentences or documents, e.g. algorithms for generating these embeddings sometimes perform little better than the mean of their word embeddings (Wieting and Kiela, 2018);

4. although focus has recently turned to large language models and *contextualised embeddings* (e.g. Devlin et al., 2019; Brown et al., 2020), understanding "simpler" un-contextualised embeddings may provide an essential foundation for understanding these more complex models;

5. not only can many words/entities appear in both text corpora and knowledge graphs, they may arise in other data, e.g. in speech or as class labels in image classification tasks. A clearer understanding of word embeddings may lead to a principled basis for *multi-modal embeddings* learned jointly across such domains.

## 1.1 Personal Motivation

This thesis began with an early investigation into the representation of knowledge graph data and the surprising ability of knowledge graph representation (KGR) models to predict many previously unknown facts from known true facts. This early research led to the development of two improved KGR models: *TuckER* (Balažević et al., 2019c) based on Tucker tensor decomposition and capable of *multi-task learning* across relations; and *HypER* (Balažević et al., 2019a), which simplified yet outperformed the then state-of-the-art convolutional model *ConvE* (Dettmers et al., 2018). However, this investigation revealed little understanding of how KGR models work, i.e. how the latent semantic structure they capture enables unobserved facts to be predicted, leaving no clear path to their further improvement or to understand when they fail. One clear pattern was that KGR models typically represent *entities* as vectors (embeddings) and *relations* between entity pairs as parametric transformations (e.g. matrix multiplication) that map embeddings of related entities "close together" (e.g. by Euclidean distance or dot product, as determined by the loss function). As such, KGR models represent entities comparably to how word embedding algorithms represent words, i.e. as vectors in a *semantic space*, hinting that perhaps a common latent structure may underpin both. This thesis is based on following this intuition to explore the latent structure of word embeddings, both as an end in itself (§3, §4) and as a step towards deciphering the latent structure of knowledge graph representation (§5).

## 1.2 Thesis Structure

This thesis consists of three main chapters, each centred around a published conference paper:

- Chapter 3 explains why the embeddings of words that form an analogy, e.g. "man is to king as woman is to queen", often approximate a parallelogram, based on:

  **Analogies Explained: Towards Understanding Word Embeddings.** C. Allen and T. Hospedales, *International Conference on Machine Learning*, 2019 (Honourable Mention).

- Chapter 4 explores further the relationship between word embedding geometry and word semantics, extending it to other semantic relationships and explaining common heuristics, based on:

  **What the Vec? Towards Probabilistically Grounded Embeddings.** C. Allen, I. Balažević and T. Hospedales. *Advances in Neural Information Processing Systems*, 2019.

- Chapter 5 extends the findings for word embeddings into a correspondence between the semantic relations of knowledge graphs and geometric relationships between entity embeddings learned by knowledge graph representation models, based on:

  **Interpreting Knowledge Graph Relation Representation from Word Embeddings.** C. Allen*, I. Balažević*, and T. Hospedales. *International Conference on Learning Representations*, 2021.

# Chapter 2

# Background and Related Work

## 2.1 Word Embedding

Representing words by vectors, or *embeddings*, that in some way capture a word's semantic meaning is essential to almost every natural language processing (NLP) task, such as part-of speech (POS) tagging, named-entity recognition (NER), sentiment analysis and machine translation.

### 2.1.1 Count-based Embeddings

Early word embeddings involved human judgement to determine both the meaning of each dimension, and where a word should fall along them (e.g. Osgood et al., 1957). The subjectivity involved was reduced in subsequent *count-based embeddings*, generated automatically from *co-occurrence counts* extracted from large text corpora (e.g. Schütze, 1992; Lund and Burgess, 1996; Landauer and Dumais, 1997). This approach was inspired by the *distributional hypothesis* that words with similar meaning appear in similar contexts (Wittgenstein, 1953; Harris, 1954), reflected in Firth's maxim "you shall know a word by the company it keeps" (Firth, 1957).

Given a dictionary of words to embed $\mathcal{D}$, the $j^{th}$ component of the embedding of $w \in \mathcal{D}$ is determined by the number of times $c_j$, the $j^{th}$ word of a context vocabulary $\mathcal{E}$, is observed within a defined *neighbourhood*, or *context*, of $w$ across the corpus. The "neighbourhood" around each instance of $w$ can be defined in several ways, e.g. as a fixed number of words (or tokens) either side, as all words in the same document of a collection, or according to a particular syntactic rule (Terra and Clarke, 2003). The set of embedded words $\mathcal{D}$ need not be the same as the context vocabulary $\mathcal{E}$, the size of which defines the embedding length $m = |\mathcal{E}|$. Since a large context vocabulary is required to capture the wide spectrum of semantic meaning, raw count vectors can be large, yet sparse since many words do not co-occur at all within a given corpus. As such, many algorithms include a dimensionality reduction step, such as *principal component analysis* (PCA), so that embeddings capture sufficient semantic meaning but remain succinct and efficient to use (e.g. Schütze, 1992; Landauer and Dumais, 1997).

Embedding components need not contain direct co-occurrence counts and several statistics derived from such counts have been explored (Lee, 2001; Bullinaria and Levy, 2007; Turney and Pantel, 2010). Of these, *Point-wise Mutual Information* (**PMI**) from in-

$$\text{An example of } \underbrace{\text{context and target}}_{context} \overbrace{\text{words}}^{target} \underbrace{\text{for a context}}_{context} \text{ window of size 6.}$$

Figure 2.1: An example of context and target words for a context window of size $l = 6$ (3 words each side of the target word). The target location steps through each token in the corpus with the context window moving relatively.

formation theory, defined as

$$\text{PMI}(w, c) = \log \frac{p(w, c)}{p(w)p(c)} = \log \frac{p(w|c)}{p(w)}, \tag{2.1}$$

has been found to perform well (Church and Hanks, 1990; Terra and Clarke, 2003). PMI quantifies the independence of two random variables: a positive value indicates that the word $w$ occurs more frequently in the presence of word $c$ than otherwise (i.e. under its marginal probability); a negative value indicates $w$ becomes less frequent in the presence of $c$; and zero indicates that $w$ and $c$ occur independently. A closely related statistic, *positive PMI* (**PPMI**), where $\text{PPMI}(w, c) = \max\{0, \text{PMI}(w, c)\}$, is also often found to perform well (Turney and Pantel, 2010). For a more comprehensive review of count-based embeddings we defer to Bullinaria and Levy (2007); Turney and Pantel (2010); Baroni and Lenci (2010).

### 2.1.2 Neural Embeddings

More recently, word embeddings have been generated from the parameters of *neural networks*, in particular following their application to *language modelling* (e.g. Bengio et al., 2000, 2003; Emami et al., 2003; Morin and Bengio, 2005). In language modelling, a key task is to learn the probability distribution over word sequences, which is often approached by predicting each word from its preceding words. Since the number of word sequences is exponential in the vocabulary size, those observed form an extremely sparse subset of the full space of sequences, hence models that directly capture such statistics, e.g. *n-gram* models (Katz, 1987; Jelinek and Mercer, 1980; Chen and Goodman, 1999), suffer the "curse of dimensionality" (Bengio et al., 2003). Neural networks were introduced to tackle this by learning *distributed representations* of words used to predict sequence statistics. The intuition is that if similar representations can be learned for semantically similar words, then learning to predict the statistics of one (observed) sequence should partially improve predictions for all semantically similar sequences (even if unobserved). Early neural language models gave promising results, especially in combination with n-gram models (Bengio et al., 2003; Emami et al., 2003), but suffered from long training times, e.g. 3 weeks for 5 epochs (Bengio et al., 2003). Subsequent neural language models focused on improved training times (e.g. Schwenk and Gauvain, 2005; Mnih and Hinton, 2007; Mnih and Teh, 2012).

Interestingly, the distributed representations learned by language models were found to improve performance when used as word features for natural language processing applications, such as semantic role labelling (Collobert and Weston, 2008), sentiment analysis (Maas and Ng, 2010), named entity recognition (Turian et al., 2010), and parsing (Socher et al., 2011). This finding led to algorithms that resemble earlier language models, but whose sole purpose is to generate word embeddings useful in NLP applications. These include **word2vec** (Mikolov et al., 2013a), the log bi-linear language model (**LBL**) of Mnih and Kavukcuoglu (2013) and **GloVe** (Pennington et al., 2014). Similar to many earlier count-based approaches, these models consider in turn each token of a large corpus (a *target* word) together with $l$ nearby tokens in its *context window* (see Figure 2.1). The vocabularies of embedded words and context words are typically the same (i.e. $\mathcal{D} = \mathcal{E}$), referred to simply as *the* dictionary $\mathcal{E}$. The word embeddings output by these algorithms, often known as *neural* or *dense* embeddings, have been shown to significantly improve performance on down stream NLP tasks, relative to count-based embeddings (Baroni et al., 2014). In this thesis, we focus on these neural embeddings and their close derivatives (e.g Ling et al., 2015; Jameel et al., 2019), referred to henceforth as simply *word embeddings*.

### 2.1.2.1 Word2vec

The *word2vec* model (Mikolov et al., 2013a,b) has two forms: the *Continuous Bag-of-Words* model (**CBOW**) and the *Continuous Skip-Gram* model (**Skip-Gram**). The former learns to predict a target word from its context words; the latter predicts context words from a target word, treating them as conditionally independent given the target word. Both models learn two $d$-dimensional vectors per dictionary word, one relating to it as the target word, the other as a context word. Typically the former are taken to be the output word embeddings, the latter discarded.

As initially proposed, CBOW and Skip-Gram make predictions using a *softmax* function of the general form

$$p(y|x) = \exp \phi(x,y)/ \sum_{y' \in \mathcal{Y}} \exp \phi(x, y'), \tag{2.2}$$

where $\phi$ is some function of the arguments $x$, $y$; and $\mathcal{Y}$ is the domain of the predicted random variable y. Since the denominator sums over all possible values of y, predicting a word, e.g. the target word in CBOW, with the softmax function involves a sum over all dictionary words, which is computationally expensive for a typical dictionary size of order $\geq 10^6$ words. Initially, this was addressed using a *hierarchical softmax* approach (Morin and Bengio, 2005; Mnih and Hinton, 2008) and later, for the Skip-Gram model, by introducing **negative sampling** (Mikolov et al., 2013b), a variation on **noise contrastive estimation** (Gutmann and Hyvärinen, 2010, 2012) (see panel), as had been recently applied to language modelling (Mnih and Teh, 2012).

In an approach similar to NCE, Skip-Gram with negative sampling (**SGNS**) learns to classify the context words observed for each target word from amongst $k$ (e.g. 5 – 20) times as many words drawn from a noise distribution (*negative samples*). Negative sampling primarily differs from NCE in that the noise distribution is sampled from but, together with $k$, not "factored out" in the loss function. As such, the noise distribution need not necessarily be known analytically, but remains a component of the statistics learned by the model (discussed further in §2.1.5).

Good results were obtained by setting the noise distribution to the marginal (uni-gram) word distribution raised to the power 0.75 and re-normalised. Other empirically-driven algorithm details include *sub-sampling* – discarding target words according to their global frequency, subject to a threshold $\tau$; and *down-weighting* the contribution of context words according to the number of tokens between them and the target word.

> **Noise Contrastive Estimation** (**NCE**) is a technique for estimating the parameters $\theta$ of a statistical model $p_\theta$ of a data distribution $p_d$, given samples from $p_d$. Under NCE, a logistic classification model learns to distinguish true data samples from samples drawn randomly from an analytically defined *noise distribution* $p_n$.
>
> A logistic regression model learns to predict a binary random variable $z \in \{0,1\}$ associated with a random variable $x \in \mathcal{X}$ given samples $\{(x,z)\}$ drawn from a joint distribution, such that $\sigma(f(x)) \approx p(z=1|x=x)$, where $\sigma$ is the standard *sigmoid function*, $\sigma(x) = (1+e^{-x})^{-1}$, and $f: \mathcal{X} \to \mathbb{R}$ is a function of the data. Implicitly, $f(x)$ learns to approximate the "log odds" $\log \frac{p(z=1|x)}{p(z=0|x)} = \log \frac{p(x|z=1)}{p(x|z=0)\kappa}$, a log ratio of class conditional distributions with $\kappa = \frac{p(z=0)}{p(z=1)}$ the class probability ratio.
>
> In NCE, z is chosen to indicate whether a sample $x$ is from the data distribution (z = 1) or noise distribution (z = 0), hence the logistic classifier learns the log ratio $f(x) \approx \log \frac{p_d(x)}{p_n(x)k}$, with $k$ the ratio of noise to data samples. Since both $k$ and $p_n$ are known, $f(x)$ can be chosen $f(x) = \log \frac{p_\theta(x)}{p_n(x)k}$, whereby $p_\theta$ learns to approximate the true data distribution $p_d$.
>
> Unlike *maximum likelihood estimation*, NCE does not require the statistical model $p_\theta$ to be normalised since the normaliser can also be learned. Mnih and Teh (2012) apply NCE to predict the next word given a set of preceding words in a language model by calculating the conditional probability distribution over all words as in Equation 2.2 but avoiding the costly denominator.

The loss function of SGNS is given by:

$$\ell^{SGNS} = \sum_{w,c} \log \sigma(\boldsymbol{w}^\top \boldsymbol{c}) + \sum_{r=1}^{k} \mathbb{E}_{c' \sim p_n}[\log \sigma(-\boldsymbol{w}^\top \boldsymbol{c}')], \qquad (2.3)$$

where the outer summation is over all target-context word pairs $(w,c)$ observed in a training corpus (post sub-sampling and down-weighting). The expectation is over all negatively sampled words $c'$ drawn from the noise distribution $p_n$. Enumerating all words in the dictionary $\mathcal{E}$, the $i^{th}$ word is denoted $w_i$ when considered a target word and $c_i$ when a context word. Target and context word embeddings $\boldsymbol{w}_i, \boldsymbol{c}_j \in \mathbb{R}^d$ represent words $w_i, c_j$ respectively, and can be viewed as the $i/j^{th}$ columns of embedding matrices $\boldsymbol{W}, \boldsymbol{C} \in \mathbb{R}^{d \times |\mathcal{E}|}$. For clarity, $w_i$ and $c_i$ always refer to the same word ($i \in \{1,...,|\mathcal{E}|\}$), whereas embeddings $\boldsymbol{w}_i$ and $\boldsymbol{c}_i$ may differ.

SGNS word embeddings are not only learned more quickly than using hierarchical softmax, they are found to give significantly improved performance on tasks such as predicting word *similarity* and *analogical reasoning* (see §2.1.4) (Mikolov et al., 2013b).

#### 2.1.2.2 Log Bi-linear Language Model

The LBL model (Mnih and Kavukcuoglu, 2013) has two forms analogous to those of word2vec: *vector LBL* (**vLBL**), which predicts target words from a set of context words (*cf* CBOW); and its *inverse* (**ivLBL**) that predicts context words from the target word, assuming conditional independence given the target word (*cf* Skip-Gram). The main differences between LBL models and their word2vec counterparts are the full adoption of NCE, and the inclusion of *position-dependent weights* and biases $b_i$ for each word (the authors go on to show that position-dependent weights in fact tend to impede performance and we omit them here). Since NCE is used rather than negative sampling, the loss functions of vLBL and ivLBL are minimised, respectively, if:

$$\log p(w_i|c_{j_1}, ..., c_{j_l}) = \boldsymbol{w}_i^\top \sum_{r=1}^{l} \boldsymbol{c}_{j_r} + b_i, \tag{2.4}$$

$$\log p(c_j|w_i) = \boldsymbol{w}_i^\top \boldsymbol{c}_j + b_j, \tag{2.5}$$

where $l$ is the length of the context window. The LBL models are shown to outperform their word2vec counterparts trained using hierarchical softmax (Mnih and Kavukcuoglu, 2013) but not when negative sampling is used (Mikolov et al., 2013b). This suggests that negative sampling, although a simplification of NCE, is preferable to using NCE in full.

### 2.1.2.3  GloVe: Global Vectors for Word Representation

Inspired by word2vec, the GloVe model (Pennington et al., 2014) has a similar underlying architecture: two matrices of target and context word embeddings that interact via a dot product. The GloVe loss function is given by

$$\ell^{GloVe} = \sum_{i,j} f_{i,j} \big(\boldsymbol{w}_i^\top \boldsymbol{c}_j + b_i + b_j - \log p(w_i, c_j)\big)^2, \tag{2.6}$$

where weighting terms $f_{i,j} = \min(1, \frac{p(w_i, c_j)}{\tau})^{0.75}$, with threshold $\tau$, act comparably to the sub-sampling in word2vec. The main differences between the GloVe and SGNS algorithms are that GloVe: (i) has bias terms $b_i, b_j$ for each word, (ii) uses pre-computed co-occurrence counts, and (iii) has a *weighted least squares* loss function.

An initial claim by Pennington et al. (2014) that GloVe embeddings outperform those of word2vec has been questioned in several subsequent evaluations (Lai et al., 2016; Levy et al., 2015; Schnabel et al., 2015; Wang et al., 2019). A cursory meta-analysis of these results suggests that the performance of these two popular word embedding algorithms is broadly comparable over a range of downstream tasks and datasets, perhaps slightly in favour of word2vec.

### 2.1.3  Other Word Embedding Models

Many other algorithms can be used to generate vector representations of words, e.g. *Latent Dirichlet Allocation* (Blei et al., 2001, 2003), a hierarchical probabilistic model that views documents as distributions over topics and topics as distributions over words; *FastText* (Bojanowski et al., 2017), which considers *sub-word* information, i.e. the spelling of a word; and *Eigenwords* (Dhillon et al., 2015), a spectral approach based on *canonical correlation analysis*. We focus, however, on word embeddings produced by the methods previously described, e.g. SGNS and GloVe, as the first neural embeddings to exhibit the semantic structure we aim to understand, in particular the *analogy phenomenon* (§2.1.4), due to their simplicity yet strong performance and their widespread use and popularity (including application to other domains).

Recently, much attention has turned towards large *language models*, such as *BERT* (Devlin et al., 2019) and *GPT1-3* (Radford et al., 2018, 2019; Brown et al., 2020). These highly parameterised models take into account a word's immediate context within a passage of text when considering its meaning in each instance, offering clear advantages in problems such as *word sense disambiguation* or complex tasks such as *text generation*. We do not, however, consider these models or any word representations learned within their parameters for two reasons: (i) our related interest in *knowledge graph representation*, where entities are typically represented by vectors without considering contextual information, e.g. from other elements of a triple or the wider knowledge graph; and (ii) it follows a natural approach of tackling the least complex models first.

### 2.1.4 Semantic Properties of Word Embeddings

An ideal set of word embeddings might be expected to enable the successful automation of an arbitrary NLP task. Since the breadth of such tasks is vast, a subset is chosen to evaluate the relative performance of different word embeddings. Common evaluation tasks are either full "down-stream" NLP tasks, i.e. practical use-cases of embeddings (termed *extrinsic* evaluation (Galliers and Jones, 1993; Jones and Galliers, 1995)); or test for a particular semantic property believed indicative of useful word embeddings (termed *intrinsic* evaluation).

Of the two, extrinsic evaluation might be considered the more practical indicator of embedding performance since it directly involves NLP applications. However, extrinsic tasks typically require word embeddings to be incorporated into a task-specific model, e.g. a softmax classifier for *part-of-speech tagging* or *named entity recognition*, or a recurrent neural network for *language modelling*. As such, the performance of word embeddings is entangled with that of the task-specific model, their compatibility with that model, and the complexity of the evaluation task. This makes it difficult to theoretically analyse why certain word embeddings perform better than others or how they capture semantic properties. In contrast, intrinsic evaluation tasks directly estimate specific semantic properties using relatively simple functions of the embeddings, without having to "embed them in a complete NLP system" (Linzen, 2016). For this reason we, together with much prior research (discussed in §2.1.5), focus on intrinsic evaluation tasks to gain theoretical insight into the latent semantic structure captured by word embeddings. We note that the relationship between performance on extrinsic and intrinsic evaluation tasks is non-trivial (e.g. see Wang et al., 2019, Figure 1) and understanding one may not readily explain the other, but the latter appears to offer a simpler place to start.

Two common intrinsic evaluation tasks are:

- predicting the semantic **similarity** or **relatedness** of a word pair $w_i, w_j$. These are typically estimated by the **cosine similarity** of their (target) embeddings,

$$\text{cos\_sim}(w_i, w_j) = \cos \theta_{i,j} = \frac{\boldsymbol{w}_i^\top \boldsymbol{w}_j}{\|\boldsymbol{w}_i\|\|\boldsymbol{w}_j\|}, \quad (2.7)$$

  where $\theta_{i,j}$ is the interior angle between $\boldsymbol{w}_i$ and $\boldsymbol{w}_j$ (Figure 2.2a). The overall evaluation result is given by the correlation between such predictions and human judgements.

- solving **analogies** (or *analogical reasoning*), i.e. predicting the missing word $b^*$ from an expression of the form "$a$ is to $a^*$ as $b$ is to ...?" (also denoted ($a$ :

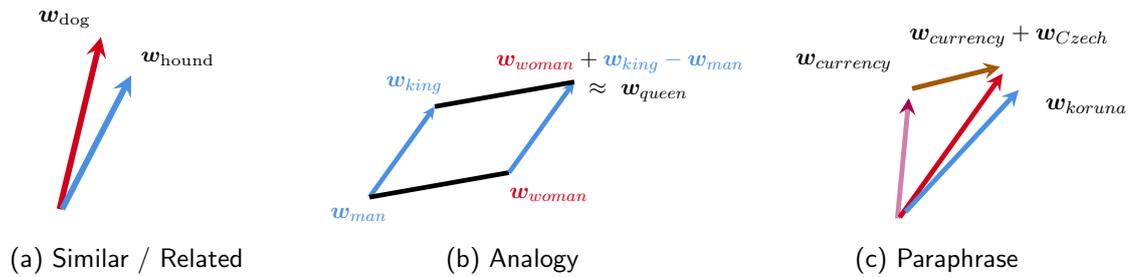

(a) Similar / Related  (b) Analogy  (c) Paraphrase

Figure 2.2: Illustration of the geometric relationships between word embeddings corresponding to different semantic relationships (a) similarity and relatedness; (b) paraphrases; and (c) *analogies*, or phrases of the form "$a$ is to $a^*$ as $b$ is to $b^*$". "Closeness" of embeddings is typically measured by cosine similarity, the angle subtended at the origin.

$a^* :: b :?$)), such as "*man* is to *king* as *woman* is to ...?". This task arises from the empirical observation that the solution is often identified by finding the closest embedding to $\boldsymbol{w}_{a^*} - \boldsymbol{w}_a + \boldsymbol{w}_b$, measuring closeness by cosine similarity and excluding the embeddings of $a$, $a^*$ and $b$, often referred to as the **vector offset** method (Mikolov et al., 2013b,c). This surprising finding, which we refer to as the **analogy phenomenon**, suggests that the embeddings of an analogy approximate a parallelogram (Figure 2.2b).

Further to analogies, Mikolov et al. (2013b) also observed that word2vec embeddings can be "meaningfully combined" by vector addition, e.g. $\boldsymbol{w}_{Czech} + \boldsymbol{w}_{currency} \approx \boldsymbol{w}_{koruna}$. Originally called "additive compositionality", we refer to this as **paraphrasing** (folowing Gittens et al., 2017), i.e. *koruna* paraphrases the word set {*Czech, currency*} (Figure 2.2c). Identifying paraphrases from word embeddings is not a common evaluation task, but we include it here as a further simple relationship between semantics and word embedding geometry, comparable to analogies and similarity.

### 2.1.5 Prior theoretical analysis of word embeddings

At a high level, the neural word embedding models that we consider, word2vec, LBL and GloVe (§2.1.2), take a similar approach. Each learns to predict the context words surrounding a target word (or vice versa) using a *linear* underlying model from which word embeddings are extracted. Here, we review works that try to understand how the embeddings of these algorithms reflect semantics by analysing them and/or their embeddings. This problem can be broken down into understanding:

(i) what values embedding parameters learn when an algorithm's loss function is minimised; and

(ii) how such values correspond to word semantics.

#### 2.1.5.1 What word embedding parameters learn

**LBL**: vLBL and ivLBL are trained using NCE to satisfy:[1]

$$\boldsymbol{w}_i^\top \sum_{r=1}^{l} \boldsymbol{c}_{j_r} + b_i = \log p(w_i | c_{j_1}, ..., c_{j_l}) \qquad (2.4)$$

$$\boldsymbol{w}_i^\top \boldsymbol{c}_j + b_j = \log p(c_j | w_i) \qquad (2.5)$$

**GloVe**: The loss function of GloVe (Equation 2.6) can be seen to be minimised if:

$$\boldsymbol{w}_i^\top \boldsymbol{c}_j + b_i + b_j = \log p(w_i, c_j) \qquad (2.8)$$

**Word2vec**: By setting its derivative to zero, Levy and Goldberg (2014b) show that the SGNS loss function (Equation 2.3) is minimised if:

$$\begin{aligned} \boldsymbol{w}_i^\top \boldsymbol{c}_j &= \log \frac{p(w_i, c_j)}{k\, p(w_i) p(c_j)} \\ &\doteq \mathrm{PMI}(w_i, c_j) - \log k \end{aligned} \qquad (2.9)$$

We note that the marginal distribution $p(c_j)$ in the denominator of Equation 2.9 reflects the negative sampling distribution and appears in the learned statistics due to not being "factored out" as it would under NCE (see §2.1.2.1). Under the original softmax formulation, Skip-Gram embeddings would instead satisfy $\boldsymbol{w}_i^\top \boldsymbol{c}_j = \log p(c_j | w_i)$, very similar to ivLBL (Equation 2.5). Thus, although negative sampling was implemented as a means of improving efficiency of the softmax computation, it in fact changes the solution of the Skip-Gram algorithm, i.e. what its embeddings learn.

This relationship to PMI links SGNS to earlier *general embedding methods*, used for exploratory data analysis and visualisation. Globerson et al. (2004) propose a method to represent discrete heterogeneous data, $x$, $y$ (with embeddings $\boldsymbol{x}$, $\boldsymbol{y}$) based on their co-occurrence statistics. Conditional probabilities are modelled as

$$p(y|x) = \frac{p(y)}{Z(x)} \exp\{-\|\boldsymbol{x} - \boldsymbol{y}\|^2\}, \qquad (2.10)$$

for a partition function $Z(x) = \sum_{y'} p(y') e^{-\|\boldsymbol{x} - \boldsymbol{y}'\|^2}$, whereby embeddings satisfy

$$-\|\boldsymbol{x} - \boldsymbol{y}\|^2 = \mathrm{PMI}(x, y) - \log Z(x), \qquad (2.11)$$

where $-\log Z(x)$ can be viewed as a bias specific to $x$. Here, PMI statistics are encoded in the *Euclidean distance* rather than dot product between embeddings, but since $-\|\boldsymbol{x} - \boldsymbol{y}\|^2 = 2\boldsymbol{x}^\top \boldsymbol{y} - \|\boldsymbol{y}\|^2 - \|\boldsymbol{x}\|^2$, Equation 2.11 compares closely to the SGNS solution (Equation 2.9) with $\ell_2$ regularisation of the embeddings.

Equations 2.4, 2.5, 2.8 and 2.9 define solutions to the word embedding algorithms, specifying relationships that the parameters achieve, or *learn*, when the loss function is minimised. A few algorithm details are omitted for simplicity: raising negative sampling probabilities ($p(c_j)$ in the denominator of Equation 2.9) to the power 0.75 and down-weighting more distant target-context pairs in SGNS. However, the equations capture the core aspects of the algorithms and can be modified to include the omitted details.

---

[1]Equations 2.4, 2.5 repeated here for ease of reference.

Considered over all target and context words, several of these solutions can be interpreted as **factorisation of a matrix** of co-occurrence statistics $\boldsymbol{M} \in \mathbb{R}^{|\mathcal{E}| \times |\mathcal{E}|}$,

$$\boldsymbol{W}^\top \boldsymbol{C}, = \boldsymbol{M}, \tag{2.12}$$

where $\boldsymbol{W}, \boldsymbol{C}$ are embedding matrices in which word embeddings are columns.

- GloVe: embeddings and biases can be concatenated, $\boldsymbol{w}'_i{}^\top = [\, b_i \,|\, \cdots \boldsymbol{w}_i^\top \cdots \,|\, 1\,]$, $\boldsymbol{c}'_j{}^\top = [\, 1 \,|\, \cdots \boldsymbol{c}_j^\top \cdots \,|\, b_j \,]$, and stacked into matrices $\boldsymbol{W}, \boldsymbol{C} \in \mathbb{R}^{(d+2) \times |\mathcal{E}|}$, whereby Equation 2.8 is equivalent to Equation 2.12 with $\boldsymbol{M}_{i,j} = \log p(w_i, c_j)$.
- SGNS: arranging target and context embeddings as columns of $\boldsymbol{W}, \boldsymbol{C} \in \mathbb{R}^{d \times |\mathcal{E}|}$, Equation 2.9 is equivalent to Equation 2.12 with $\boldsymbol{M}_{i,j} = \mathrm{PMI}(w_i, c_j) - \log k$, known as **shifted PMI** ("shift" refers to the $\log k$ term).
- ivLBL: concatenating embeddings and biases (as for GloVe), Equation 2.5 is equivalent to Equation 2.12 with $\boldsymbol{M}_{i,j} = \log p(c_j | w_i)$.

CBOW and vLBL can not be interpreted as matrix factorisation (for $l > 1$), since they learn probabilities with $l + 1$ variables that form $(l + 1)$-order tensors.

From the perspective of matrix factorisation, it is clear that for a solution to be reached, the embedding dimension $d$ must at least equal the rank of $\boldsymbol{M}$. In practice, $d$ is in the range $100\text{--}1000$, whereas $\boldsymbol{M}$ typically has full rank equal to the dictionary size, of order $> 10^5$. Thus, the solutions are only achieved *approximately* with errors that depend on (i) the embedding dimension; (ii) the loss function, e.g. least squares or binary cross entropy; and (iii) the weighting of each loss component, e.g. the $f_{i,j}$ terms in GloVe or the probability of sampling each target-context word pair in word2vec.

The realisation that SGNS, in particular, implicitly factorises a matrix of co-occurrence statistics (Levy and Goldberg, 2014b) is significant because it relates this effective but initially uninterpretable algorithm to earlier work on count-based methods that explicitly factorise such matrices. Empirical evidence supporting the theoretical connection between SGNS and PMI shows that SGNS performs comparably on several tasks to count-based embeddings based on both (a) shifted PMI and (b) explicit factorisation (SVD) of a PPMI matrix (Levy and Goldberg, 2014b; Levy et al., 2015).

#### 2.1.5.2 Relating word embedding parameters to semantics

Here we discuss prior theoretical research into the relationship between word embeddings and semantics, much of which focuses on the analogy phenomenon (§2.1.4), aiming to explain why embeddings of words that form an analogy often approximate a parallelogram. This phenomenon has piqued particular interest since word embeddings are not trained to achieve it. It is also of practical use for solving analogical queries by the vector offset method (Mikolov et al., 2013c). By contrast, *similar* words are expected to have similar distributions of context words under the distributional hypothesis, hence finding that their embeddings are nearby seems relatively intuitive. Although analyses typically consider SGNS, their findings may extend to other word embedding models due to the similarity between what their embeddings learn (§2.1.5.1).

---

Before the connection between SGNS and PMI was discovered (Levy and Goldberg, 2014b), explanations for how embeddings capture semantics were largely qualitative:

**Mikolov et al. (2013c)** justify the additive compositionality of embeddings to find paraphrases on the basis that embeddings "represent" their context distributions logarithmically "so the sum of two word vectors is related to the product of the two context distributions."

**Levy and Goldberg (2014a)** show empirically that many analogies can also be solved by the vector offset method using high-dimensional count-based embeddings based on positive PMI. An alternative way to combine the word embeddings of analogies ("3CosMul") is also proposed. Having introduced a notion of word "aspects", e.g. *king* has aspects *man* and *royal*, the authors suggest that (i) "*relational similarities* [between word pairs] can be viewed as a composition of *attributional similarities* [of words]" that relate to aspects; and (ii) "solving the analogy question involves identifying the relevant aspects, and trying to change one of them while preserving the other". The intuition is that a word is (in some sense) the *sum* of its aspects, e.g. *king* = *man*+*royal*, *queen* = *woman*+*royal*. As such the analogy "*man* is to *king* as *woman* is to *queen*" is solved by identifying the difference between *man* and *king*, i.e. *royal*, and adding this to *woman*, while preserving the *woman* aspect, to give *queen*. Context words related to an aspect are identified by point-wise multiplying count-based embeddings of words sharing that aspect. (Note that this does not relate to neural embeddings and does not explain how *aspects* are quantified or "added".)

**Pennington et al. (2014)** also refer to "aspects of meaning" and that words co-occur based on mutual aspects in justifying the GloVe loss function (Equation 2.6).

---

The connection between SGNS embeddings and PMI (Levy and Goldberg, 2014b) spurred several quantitative analyses.

**Arora et al. (2016)** propose a *generative latent variable model* for language in which a *discourse* vector takes a *slow random walk* in a latent space containing word embeddings. Under the model, words are emitted over a series of time-steps with log probability proportional to the dot product between their embedding and the position of the discourse vector at the current step. The model has an appealing intuition but requires several strong and unexplained assumptions (as noted by Gittens et al., 2017), e.g. that directions of word embeddings are *isotropically* distributed under a spherical Gaussian; that embedding magnitudes are upper bounded by a fixed constant $\kappa$; and that the stationary distribution of the random walk is uniform over the unit sphere. Furthermore, the generative model assumes each word has only one embedding, whereas the embedding models it aims to explain, such as SGNS and GloVe, have two; and the square of embedding norms are expected to relate linearly to log marginal word probabilities (see their Equation 2.4), which is not observed in practice for models such as SGNS and GloVe (Arora et al., 2015, Fig. 6).

**Hashimoto et al. (2016)** relate word embedding to *metric recovery* (see panel) on the premise that word embeddings exist in an innate Euclidean *semantic space* from where they are *recoverable*. To support this notion, studies from psycho-metrics and cognitive science are referenced that: (a) show reasoning problems such as analogies can be solved with word embeddings derived from human word similarity judgements (Rumelhart and Abrahamson, 1973); and (b) propose that human reasoning is consistent with embeddings in a Euclidean space, e.g. human subjects solve reasoning problems by "finding the word closest (in semantic space) to an ideal point", e.g. as the vertex of a

> **Metric Recovery** is a form of *manifold learning*. Manifolds (or, loosely speaking, surfaces) in high dimensional space can be approximated by joining nearby points on the manifold by straight lines to form a *graph*. Shortest paths over the graph approximate *geodesics* over the manifold. Metric recovery methods recover the relative co-ordinates and probability distribution of the data by considering *random walks* over a neighbourhood graph (see Hashimoto et al., 2015).

parallelogram for analogies, and as points in a line for series (e.g. "penny, nickel, dime, ?") (Sternberg and Gardner, 1983). The connection to an assumed Euclidean semantic space is extended to word embeddings that learn PMI co-occurrence statistics by showing that such statistics (from several text corpora) correlate with human judgements and approximately satisfy two tests indicating that they are "mostly consistent with a Euclidean hypothesis". Several word embedding algorithms, including GloVe and Skip-Gram, are thus framed as metric recovery methods in which passages of text are considered *random walks* over a graph in semantic space where each word is represented by a node. This explanation involves several strong assumptions:

(i) that an innate semantic space exists, without justification of its origin, the meaning attributed to different dimensions, or how those meanings are chosen;

(ii) that each word is represented by a single embedding, despite several of the considered word embedding models generating two (e.g. Skip-Gram and GloVe); and

(iii) that, in the underlying theory (Hashimoto et al., 2015), context windows are asymptotically large, whereas several of the considered word embedding models (e.g. Skip-Gram and GloVe) use small windows in practice (e.g. $l = 5$).

We note also that the observed additive compositionality of embeddings is not explained, rather it is an assumed property of the underlying semantic space.

**Paperno and Baroni (2016)** consider the addition of (PMI) count-based embeddings for *phrases* "ab" formed of words $a, b$. It is shown that, for each embedding dimension corresponding to any context word $c$:

$$\text{PMI}(ab, c) = \underbrace{\log \frac{p(a|c)}{p(a)}}_{\text{PMI}(a,c)} + \underbrace{\log \frac{p(b|c)}{p(b)}}_{\text{PMI}(b,c)} + \underbrace{\log \frac{p(ab|a \wedge c)}{p(b|c)}}_{\text{``PMI}(ab|c)\text{''}} - \underbrace{\log \frac{p(ab|a)}{p(b)}}_{\text{``PMI}(ab)\text{''}}. \quad (2.13)$$

$$\underbrace{\phantom{\log \frac{p(ab|a \wedge c)}{p(b|c)} - \log \frac{p(ab|a)}{p(b)}}}_{\Delta_c}$$

Equation 2.13 shows that adding (PMI) count-based embeddings should predict "phrase vectors", subject to a "correction" term $\Delta_c$ that captures "how $[c]$ changes the tendency of $[a, b]$ to form a phrase". The authors find that for adjective-noun phrases, $\Delta_c$ is very often negative (92% of examples considered) whereby adding embeddings $\text{PMI}(a, c), \text{PMI}(b, c)$ over-estimates $\text{PMI}(ab, c)$ the true PMI vector of the noun phrase. Prediction accuracy of the noun phrase PMI vector was found to increase by weighting embedding components.

**Gittens et al. (2017)** introduce a probabilistic definition of *paraphrase*: a word $c$ is said to paraphrase a set of words $\mathcal{C} = \{c_1, ..., c_m\}$ if $p(w|\mathcal{C}) \approx p(w|c)$ for every word $w$. Analogies are then explained in terms of paraphrases: for an analogy "$a$ is to $a^*$ as $b$ is to $b^*$", the relationship between $a$ and $a^*$ is assumed to mean that $a$ paraphrases $\{a^*\} \cup \mathcal{R}$ for some set of words $\mathcal{R}$ that reflect the relationship. The same relationship is assumed to hold between $b$ and $b^*$, whereby $b$ paraphrases $\{b^*\}\cup\mathcal{R}$. A mathematical justification is given for why word embeddings can be added to find the embedding of their paraphrase and, thereby, for the linear relationship of analogy embeddings. There are, however, notable limitations to the proposed argument:

(i) context words are assumed conditionally independent given the target word, a strong assumption in practice;

(ii) negative sampling in SGNS is overlooked, hence embeddings are assumed to satisfy $p(w_i|c_j) \propto \exp \boldsymbol{w}_i^\top \boldsymbol{c}_j$, or $\boldsymbol{w}_i^\top \boldsymbol{c}_j \approx \log p(w_i|c_j) - \log Z_j$ (for a proportionality constant $Z_j$), as opposed to $\boldsymbol{w}_i^\top \boldsymbol{c}_j \approx \mathrm{PMI}(w_i, c_j)$ per Equation 2.9;

(iii) in explaining embedding compositionality, word frequencies are assumed to be *uniform* for all words, in significant contrast to the observation that most words occur rarely and relatively few occur very frequently (Zipf's law); and

(iv) by defining paraphrases in terms of minimising a KL divergence, the quantitative difference between $p(w|\mathcal{C})$ and $p(w|c)$ for each word $w$ is overlooked, whereas some words may form "good" paraphrases with small KL divergence but others may have minimal but large KL divergence, meaning that $c$ bares little semantic relationship to the paraphrased word set $\mathcal{C}$.

**Hakami et al. (2018)** analyse a *bi-linear operator* for representing relations between words, but word embeddings are pre-supposed to exist in a semantic space with independent dimensions rather than explained. Furthermore, relations between words are assumed to be independent, whereas they can be strongly inter-dependent in practice, e.g. being equivalent, the inverse or one another, or transitive. Weaker dependencies may also hold, such as one relation simply making another more likely.

#### 2.1.5.3 Other analysis of word embeddings

An extensive further body of work explores properties of word embedding algorithms or the embeddings they produce. Here, we review those most relevant to this thesis.

Several works look to **situate word embedding algorithms within established mathematical frameworks**. For example, Cotterell et al. (2017) relate the loss function of Skip-Gram to *exponential principal component analysis* (**EPCA**). EPCA generalises standard PCA, associated with Gaussian distributed data, to all exponential family distributions, e.g. Bernoulli and Multinomial. Landgraf and Bellay (2017) extend this work to account for negative sampling to show that the SGNS loss function is equivalent to weighted logistic PCA. Separately, Melamud and Goldberger (2017) draw a connection between SGNS and information theory, relating its loss function to a quantity based on *mutual information*. Each of these works offers a description of the overall embedding algorithm, but do not explain the observed semantic properties of their embeddings or why those are useful in downstream tasks.

Of works that **explore empirical properties of word embeddings**, Mimno and Thompson (2017) compare SGNS and GloVe embeddings, concluding for the former that: (i) embeddings have a large positive inner product with their mean, i.e. "point in roughly the same direction"; (ii) target and context embeddings tend to have negative inner products, i.e. "point away" from each other; and (iii) the arrangement of embeddings is strongly affected by the number of negative samples $k$.

A notable body of research **investigates the performance of the vector offset method for analogies with different types of semantic relations**.

- Levy and Goldberg (2014a) observe wide performance variability (15-91%) for SGNS embeddings across the analogies found in common benchmark datasets.

- Köper et al. (2015) note that most relations in common datasets are *morpho-syntactic*, e.g. comparatives and superlatives; and find that performance drops significantly for *paradigmatic* relations, such as *synonyms*, *antonyms* (opposites), *meronyms* (parts of, e.g. *aeroplane:cockpit*) and *hypernyms* (categories, e.g. *animal:dog*), concluding that such relations require "deeper semantic knowledge".

- Similarly, Vylomova et al. (2016) compare:

    (i) *lexical semantic* relations, such as hypernyms, meronyms, quality/action (e.g. *cloud:rain*) and cause/purpose (e.g. *cook:eat*);

    (ii) *morpho-syntactic* relations, such as noun-to-plural (e.g. *year:years*) and present-to-past tense (e.g. *know:knew*); and

    (iii) *morpho-semantic* relations, such as "light verb constructions" (e.g. *give:approval*) and collective nouns (e.g. *army:ants*).

    The vector offset method is not evaluated, but lexical semantic relations are shown to be captured less well than other types, consistent with Köper et al. (2015).

- Gladkova et al. (2016) categorise relations between *inflection* (e.g. noun-to-plural, infinitive-to-past), *derivation* (e.g. noun+"less", "un"+adjective), *lexicographic* (e.g. hypernyms, meronyms) and *encyclopedic* (e.g. animal-to-sound, country-to-language). Consistent with earlier findings, performance on morpho-syntactic inflections tends to significantly exceed that of other relation types.

- Linzen (2016) compares several variations of the vector offset method across different relation types. To solve analogy ($a:a^* :: b:?$), rather than take the closest embedding by cosine similarity to $\boldsymbol{w}_b + \boldsymbol{w}_{a^*} - \boldsymbol{w}_a$, variations include taking the closest to $\boldsymbol{w}_b$, $\boldsymbol{w}_b + \boldsymbol{w}_{a^*}$ or $\boldsymbol{w}_b - \boldsymbol{w}_{a^*} + \boldsymbol{w}_a$. Performing comparably to 3CosMul (§2.1.5.2), vector offset is seen to outperform all proposed baselines across all relation types considered. Although performance margins vary by relation, this suggests that:

    (i) the vector offset method is better than simply taking the nearest neighbour of $\boldsymbol{w}_b$, and

    (ii) adding the relational vector $\boldsymbol{w}_{a^*} - \boldsymbol{w}_a$ to $\boldsymbol{w}_b$ tends to "point in the right direction".

    The ability to predict an analogy is also seen to correlate well with that of its "reverse" ($a^*:a :: b^*:?$), supporting the notion that the vector offset *represents* the semantic relation, which can be viewed in either direction. Excluding other analogy words $a, a^*, b$ as possible solutions is shown to be essential to identifying $b^*$.

- Drozd et al. (2016) propose *supervised* methods to solve analogies using sets of commonly-related word pairs $\{(a_i, a_i^*)\}_i$ (requiring additional information):

- adding the *mean* vector $\hat{\boldsymbol{r}} = \langle \boldsymbol{w}_{a_i^*} - \boldsymbol{w}_{a_i} \rangle_i$ to $\boldsymbol{w}_b$ (*cf* Bollegala et al., 2015), rather than any individual $\boldsymbol{w}_{a_i^*} - \boldsymbol{w}_{a_i}$. Improved performance relative to vector offset is put down to avoiding "idiosyncrasies of individual words".

- using logistic regression to classify *target words* $\{a_i^*\}$ of a given relation set from their embeddings. Candidate solutions $b^*$ for a new *source word* $b$ are ranked by the product of their classification score and cosine similarity to $b$. Performance ranges from comparable to significantly better than vector offset or mean vector offset.

This study suggests that at least some semantic relations correspond to more complex geometric relationships between word embeddings than a vector offset.

More recently, the variable performance of solving analogies by vector offset has led to questions over its suitability as an intrinsic evaluation task for word embeddings (e.g. Rogers et al., 2017; Schluter, 2018). This seems somewhat justified since, as an evaluation metric, performance on the analogy task is intended to indicate "useful" word embeddings. However, if the vector offset serves as a poor representation of some analogy relations, then the results may be skewed arbitrarily and not fully reliable.

In this thesis, we take a more optimistic view that, even though performance varies, the vector offset method clearly captures *some* semantic relational structure for the word embeddings considered, as evidenced by it outperforming a nearest neighbour approach (Drozd et al., 2016). It is this positive presence of latent semantic structure in the geometry of word embeddings, for similarity, relatedness, paraphrases and analogies, that we wish to explain and in doing so justify the variable performance that arises.

## 2.2 Knowledge Graph Representation

Knowledge Graphs (or *Knowledge Bases*) are collections of known true **facts**, or **triples**, of the form ⟨*subject entity, relation, object entity*⟩, e.g. ⟨*London, captial_of, U.K.*⟩. $\mathcal{E}$ and $\mathcal{R}$ denote the sets of all **entities** and **relations**, respectively. Since knowledge graphs often have many entities and relations, they typically contain only a subset of all facts that are actually true amongst the vast number of combinations of entities and relations, referred to as being *incomplete*. Table 2.1 shows the statistics of popular datasets used to evaluate knowledge graph representation (**KGR**) models.

| Dataset | | # Entities ($n_e$) | # Relations ($n_r$) |
|---|---|---|---|
| FB15k | (Bordes et al., 2013) | 14,951 | 1,345 |
| FB15k-237 | (Toutanova et al., 2015) | 14,541 | 237 |
| WN18 | (Bordes et al., 2013) | 40,943 | 18 |
| WN18RR | (Dettmers et al., 2018) | 40,943 | 11 |
| NELL-995 | (Xiong et al., 2017) | 75,492 | 200 |

Table 2.1: Statistics of popular data sets for evaluating knowledge graph representations.

KGR models aim to numerically represent entities and relations of a knowledge graph so that (i) known facts can be recalled, e.g. for *question answering*; and (ii) missing facts can be inferred, known as *link prediction* or *knowledge base completion*.

A knowledge graph is commonly viewed as either:

- a *graph* $\mathcal{G}$ in which *nodes* represent entities and relations between entities are indicated by *typed, directed edges* (Figure 2.3, *left*); or

- a *binary tensor* $\mathbf{B} \in \{0,1\}^{|\mathcal{E}| \times |\mathcal{R}| \times |\mathcal{E}|}$, comparable to a graph *adjacency matrix*, where $\mathbf{B}_{s,r,o} = 1$ if subject and object entities $e_s, e_o \in \mathcal{E}$ are related by relation $r \in \mathcal{R}$, otherwise $\mathbf{B}_{s,r,o} = 0$ (Figure 2.3, *right*).

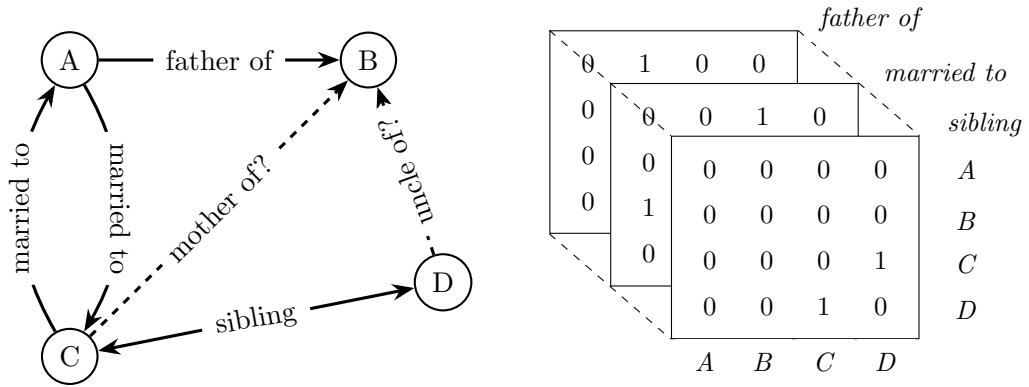

**Knowledge Graph** $\mathcal{G} = \{⟨A, father\ of, B⟩, ⟨A, married\ to, C⟩, ...\}$
**Entities** $\mathcal{E} = \{A,\ B,\ C,\ D,\ ...\}$ **Relations** $\mathcal{R} = \{married\ to,\ father\ of,\ ...\}$

Figure 2.3: Interpretations of Knowledge Graph data: (left) as a *graph* where nodes represent entities and typed, directed edges represent relations; (right) as a *binary tensor* of dimension $|\mathcal{E}| \times |\mathcal{E}| \times |\mathcal{R}|$ (cf a graph adjacency matrix). *Link prediction* aims to predict unknown facts based on those in the knowledge graph, e.g. ⟨*B, mother of, C*⟩.

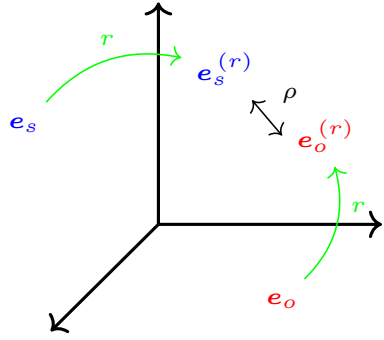

Figure 2.4: An illustration of how many knowledge graph representation can be interpreted: subject and object entities are represented by vectors $e_s, e_o \in \mathbb{R}^d$; a relation $r$ is represented by transformations that act on the entity embeddings to give relation-transformed embeddings $e_s^{(r)}, e_o^{(r)} \in \mathbb{R}^d$; the score $\phi_{s,r,o}$ is given by applying a proximity function $\rho$.

Many knowledge graph representation models have been proposed, with a significant increase in recent years. We focus on a subset of models that each achieved state-of-the-art performance and exhibit structural properties common to many KGR models:

(i) each entity $e \in \mathcal{E}$ is represented by a vector, or *embedding*, in $\mathbb{R}^d$, denoted $e_s$ when the subject entity, $e_o$ when the object entity.

(ii) each relation $r \in \mathcal{R}$ is represented by a parameterised *transformation* applied to the subject and/or object entity embedding (to give $e_s^{(r)}, e_o^{(r)} \in \mathbb{R}^d$, respectively);

(iii) a *score* $\phi_{s,r,o} = \rho(e_s^{(r)}, e_o^{(r)})$ is obtained by applying a *proximity function* $\rho$, e.g. dot product or negative Euclidean distance, to relation-transformed entity embeddings (see Figure 2.4); and

(iv) a fact prediction is derived from the score, e.g. by applying the sigmoid function, $\text{Prob}[\langle e_s, r, e_o \rangle \text{ is True}] = \sigma(\phi_{s,r,o})$.

Some models directly treat scores as predictions, using a least squares loss function to trained to train scores to 1 for true facts, 0 otherwise (e.g Nickel et al., 2011). An obvious drawback is that predictions may exceed 1 or be less than 0. Other models take a *margin-based* approach, training scores of true triples to differ, by a fixed margin, from those of mostly false random triples (e.g. Bordes et al., 2011; Yang et al., 2015). This induces an ordering over scores with high scoring triples more likely to be true, but the *relative ranking* prevents triples being independently predicted true or false. In all cases, scores are trained to be higher for true triples than false triples. Applying the sigmoid function to the score has become more common, it gives probabilistically interpretable predictions and avoids the issues of other approaches.

We categorise KGR models according to their loss function: as *additive*, *multiplicative*, or *both*. Table 2.2 summarises their score functions.

### 2.2.1 Additive KGR Models

**TransE** (Bordes et al., 2013) is the simplest additive KGR model. Each relation is represented by a vector $r \in \mathbb{R}^d$, which is added to the subject entity embedding (the relation transformation). The score $\phi_{s,r,o} = -\|e_s + r - e_o\|$ is given by the negative Euclidean distance (the proximity function) between the relation-transformed subject entity and the object entity. Fact predictions are given by applying the sigmoid function to the score. Many derivations of TransE have been proposed, but we restrict focus to the simplest architecture.

Table 2.2: Score functions for additive (A), multiplicative (M) and both (A+M) KGR models. $d$, $d_r$ are entity and relation embedding dimensionalities. $\overline{e} \in \mathbb{C}^d$ is the complex conjugate of $e$, $\otimes_n$ denotes the tensor product along the $n$-th mode, $\mathbf{C} \in \mathbb{R}^{d \times d_r \times d}$ is the core tensor of a Tucker decomposition.

|   | Model |   | Score function | Relation parameters |
|---|---|---|---|---|
| A | TransE | Bordes et al. (2013) | $-\|e_s + r - e_o\|$ | $r \in \mathbb{R}^d$ |
| M | RESCAL | Nickel et al. (2011) | $e_s^\top R e_o$ | $R \in \mathbb{R}^{d \times d}$ |
|   | DistMult | Yang et al. (2015) | $e_s^\top R e_o$ | $R = \text{diag}[r], r \in \mathbb{R}^d$ |
|   | ComplEx | Trouillon et al. (2016) | $\text{Re}(e_s^\top R \overline{e}_o)$ | $R = \text{diag}[r], r \in \mathbb{C}^d$ |
|   | TuckER | Balažević et al. (2019c) | $\mathbf{C} \otimes_1 e_s \otimes_2 r \otimes_3 e_o$ | $r \in \mathbb{R}^{d_r}$ |
| A+M | MuRE | Balažević et al. (2019b) | $-\|R e_s + r - e_o\|_2^2 + b_s + b_o$ | $r \in \mathbb{R}^{d_r}, R \in \mathbb{R}^{d \times d}$ |

### 2.2.2 Multiplicative KGR Models

In multiplicative models, relations are represented by matrices $R \in \mathbb{R}^{d \times d}$ and the score includes a *bi-linear product* with the entity embeddings, $e_s^\top R e_o$. This can be viewed as matrix multiplication of the subject entity embedding (the relation transformation) followed by a dot product with the object embedding (the proximity function). Fact predictions are obtained by applying the sigmoid function to the score. Multiplicative models relate to tensor decomposition of $\mathbf{B}$.

**RESCAL** (Nickel et al., 2011) is perhaps the earliest multiplicative model in which each relation is represented by a distinct relation matrix.

**DistMult** (Yang et al., 2015) is similar to RESCAL, but relation matrices are restricted to diagonal matrices. DistMult is found to outperform RESCAL despite its symmetry preventing asymmetric relations being accurately represented (any triple $\langle s,r,o \rangle$ is assigned the same score as its reverse $\langle o,r,s \rangle$). The higher performance of DistMult relative to RESCAL, despite being less expressive, is indicative of a tendency for KGR models to *over-fit* due to a scarcity of data relative to the number of model parameters.

**ComplEx** (Trouillon et al., 2016) has the same architecture as DistMult, but entity embeddings and relation matrices are in the complex domain. This breaks the symmetry allowing ComplEx to better represent asymmetric relations and so outperform DistMult for the same parameter number.

**TuckER** (Balažević et al., 2019c) is similar to RESCAL but the relation matrices are linear combinations of a common (low rank) set of matrices in a *core tensor* $\mathbf{C} \in \mathbb{R}^{d \times d_r \times d}$, with relation-specific coefficient vectors $r \in \mathbb{R}^{d_r}$. TuckER has fewer parameters than RESCAL and enables *multi-task learning* of relations, which is of particular benefit to scarce relations. It is shown that TuckER subsumes the DistMult and ComplEx models under specific dimensionality and parameter choices.

### 2.2.3  Additive & Multiplicative Models

**MuRE** (Balažević et al., 2019b) is a recent model that combines aspects of additive and multiplicative models, with relations parameterised by both a matrix $\boldsymbol{R} \in \mathbb{R}^{d \times d}$ and a vector $\boldsymbol{r} \in \mathbb{R}^d$.

---

The multiplicative KGR models considered in §2.2.2 can each be interpreted as different forms of low-rank *tensor factorisation*, in some cases subject to a sigmoid function relating to low *sign-rank* factorisation (Alon et al., 1985; Bouchard et al., 2015). However, KGR models that contain an additive component, e.g. TransE and MuRE, cannot readily be viewed in this way. That MuRE is found to outperform these multiplicative models on the popular WN18RR dataset (Balažević et al., 2019b), brings into question whether KGR representation is optimally addressed by tensor factorisation methods.

The approach of this thesis is to investigate knowledge graph representation by first understanding how semantic relations can be encoded geometrically between word embeddings, and, from that, consider how knowledge graph relations might be encoded similarly. This aims to avoid preconceived ideas of knowledge graph relation representation, such as using tensor factorisation, and instead build an understanding from first principles grounded in identified mathematical relationships between semantic relations, word co-occurrence statistics and word embedding geometry.

# Chapter 3

# Analogies Explained: Towards Understanding Word Embeddings

The central contribution of this chapter is the paper "*Analogies Explained: Towards Understanding Word Embeddings*" (**Analogies Explained**, **AE**), published at the *International Conference on Machine Learning* in June 2019 where it received an Honourable Mention for Best Paper. We first outline the motivation for this work (§3.1), followed by the paper itself (§3.2), the impact that it has had so far (§3.3) and a discussion (§3.4).

## 3.1 Motivation

The motivation for this thesis is to work towards an understanding of the latent structure of knowledge graph representation, where entities are represented by vectors and relations by transformations between them. In principle, this might be tackled by:

(i) constructing a mathematical model of the latent structure of knowledge graph data, or

(ii) analysing representations of KGR models to decipher what they learn.

It is not immediately clear how to proceed in either case: there is no obvious basis for modelling entities and relations of a knowledge graph, and it is unclear where to begin or what to look for amongst the plethora of KGR models. We therefore take a tangential approach inspired by the observation that *linear* structure, as seen in the loss functions of additive and multiplicative knowledge graph representation models (§2.2), also arises in the relationships between word embeddings of well-known algorithms, e.g. the additive compositionality of paraphrases and the vector offset of analogies (§2.1).

In fact, a strong parallel can be seen between analogy relations and knowledge graph relations (as noted, e.g., by Hakami et al., 2018): both involve *common binary semantic relations* between words/entities. In their respective settings, *relation* and *analogy* serve as "catch-all" terms to refer to many *heterogeneous* semantic relations. Knowledge graph relations are often considered in logical or structural terms, e.g. symmetry, asymmetry and transitivity, 1-to-1, 1-to-many, etc. (e.g. Ali et al., 2020); those of analogies tend to be differentiated semantically, e.g. as morpho-syntactic or paradigmatic (Köper et al., 2015; Vylomova et al., 2016; Gladkova et al., 2016). These different perspectives arise from separate research communities and can be unified to categorise relations in many ways in either setting.

In essence, word analogies and knowledge graph relations both involve the same range of semantic relations, simply in different formats. This is supported empirically by the KGR model *TransE* (Bordes et al., 2013). Just as the vector offset often successfully represents the relation of an analogy, *TransE* often successfully represents knowledge graph relations as vectors from subject entity embedding to object entity embedding.

The key difference between word analogy and knowledge graph relations is that each analogy $(a : a^* :: b : b^*)$ is considered individually with word pair $(a, a^*)$ providing a *single* training example of the implicit relation and $(b, b^*)$ effectively the test set; whereas, knowledge graph relations typically have multiple subject-object entity pairs as training instances. The singular training instance of analogies restricts the scope for representing their relations to relatively simple "one-size-fits-all" functions of $\boldsymbol{w}_a, \boldsymbol{w}_{a^*}$, e.g. the vector offset $\boldsymbol{w}_{a^*} - \boldsymbol{w}_a$. The multiple training samples of knowledge graphs (or *labelled* analogy datasets (e.g. Bollegala et al., 2015; Drozd et al., 2016)) allow more bespoke relation representations to be learned, e.g. with relation-specific parameters.

Thus, considering words and entities as broadly similar concepts and recognising that the same semantic relations between them can arise as analogies or in knowledge graphs, we proceed on the premise that similar latent semantic structure may underpin representations in both cases. As such, we look to understand the latent semantic structure of KGR models, our ultimate goal, by investigating how semantic relations between words are encoded in geometric relationships between word embeddings. This side-step from knowledge graphs to word embeddings seems promising for several reasons:

- it avoids the potential difficulty of deciphering what knowledge graph relation representations learn in practice (approach (ii) above) – instead, if a certain *geometric* relationship between embeddings is found to correspond to a particular *semantic* relation between words, then it indicates *one* way to represent that relation (option (i));

- certain semantic relations, *similarity*, *relatedness*, *paraphrase* and *analogy*, are known to be reflected in spatial relationships between embeddings (§2.1.4); and

- the word embeddings produced by several algorithms are known to reflect co-occurrence statistics, providing a quantitative basis to consider them (§2.1.5.1).

This explains how our core motivation leads to a consideration of word analogies. Of course, the analogy phenomenon is of particular interest in itself, not least due to its natural intrigue: word embeddings are not trained to form the (approximate) parallelogram structure that often seems to arise. That a classifier trained to identify context words surrounding each word produces word embeddings that can be seemingly added and subtracted intuitively may appear a profound anomaly, yet it reflects experimental findings from developmental psychology and human cognition that may give its understanding far greater significance (Rumelhart and Abrahamson, 1973; Sternberg and Gardner, 1983; Peterson et al., 2020). Accordingly, multiple works have sought to explain this phenomenon since its initial observation (§2.1.5.2).

The approach taken in *Analogies Explained* is inspired initially by Gittens et al. (2017), in particular their definition of *paraphrase* (§2.1.5.2), which we develop into a probabilistic identity (*AE*, Equation 5). This identity can also be seen to extend the work of Paperno and Baroni (2016), in particular Equation 2.13 (§2.1.5.2), that relates the distribution of context words around a *phrase* to those around each word of the phrase.[1] Equation 2.13 can be aligned more clearly with Equation 5 of the paper by re-stating it as:

$$\text{PMI}(\{a,b\},c) = \log\left\{\frac{p(\{a,b\}|c)}{p(\{a,b\})} \underbrace{\frac{p(a|c)p(b|c)}{p(a|c)p(b|c)} \frac{p(a)p(b)}{p(a)p(b)}}_{1}\right\}$$

$$= \underbrace{\log\frac{p(a|c)}{p(a)}}_{\text{PMI}(a,c)} + \underbrace{\log\frac{p(b|c)}{p(b)}}_{\text{PMI}(b,c)} + \underbrace{\log\frac{p(\{a,b\}|c)}{p(a|c)p(b|c)} - \log\frac{p(\{a,b\})}{p(a)p(b)}}_{\Delta_c}$$

## 3.2 The Paper

**Author Contributions**

The paper is co-authored by myself and Timothy Hospedales. As the lead author, I conceived and developed the theory behind *Analogies Explained* and wrote the paper. Tim provided helpful discussions and suggestions throughout and helped with editing the paper into its final form.

---

[1] We were unaware of this work at the time of writing *AE*, hence its omission from the paper.

# Analogies Explained: Towards Understanding Word Embeddings

Carl Allen [1]  Timothy Hospedales [1]

## Abstract

Word embeddings generated by neural network methods such as *word2vec* (W2V) are well known to exhibit seemingly linear behaviour, e.g. the embeddings of analogy *"woman is to queen as man is to king"* approximately describe a parallelogram. This property is particularly intriguing since the embeddings are not trained to achieve it. Several explanations have been proposed, but each introduces assumptions that do not hold in practice. We derive a probabilistically grounded definition of *paraphrasing* that we re-interpret as *word transformation*, a mathematical description of "$w_x$ is to $w_y$". From these concepts we prove existence of linear relationships between W2V-type embeddings that underlie the analogical phenomenon, identifying explicit error terms.

## 1. Introduction

The vector representation, or *embedding*, of words underpins much of modern machine learning for natural language processing (e.g. Turney & Pantel (2010)). Where, previously, embeddings were generated explicitly from word statistics, neural network methods are now commonly used to generate *neural embeddings* that are of low dimension relative to the number of words represented, yet achieve impressive performance on downstream tasks (e.g. Turian et al. (2010); Socher et al. (2013)). Of these, *word2vec*[2] (W2V) (Mikolov et al., 2013a) and *Glove* (Pennington et al., 2014) are amongst the best known and on which we focus.

Interestingly, such embeddings exhibit seemingly linear behaviour (Mikolov et al., 2013b; Levy & Goldberg, 2014a), e.g. the respective embeddings of *analogies*, or word relationships of the form "$w_a$ is to $w_{a^*}$ as $w_b$ is to $w_{b^*}$", often satisfy $\mathbf{w}_{a^*} - \mathbf{w}_a + \mathbf{w}_b \approx \mathbf{w}_{b^*}$, where $\mathbf{w}_i$ is the embedding of word $w_i$. This enables analogical questions such as "*man is to king as woman is to ..?*" to be solved by vector addition and subtraction. Such high order structure is surprising since word embeddings are trained using only pairwise word co-occurrence data extracted from a text corpus.

We first show that where embeddings factorise *pointwise mutual information* (PMI), it is *paraphrasing* that determines when a linear combination of embeddings equates to that of another word. We say $king$ paraphrases $man$ and $royal$, for example, if there is a semantic equivalence between $king$ and $\{man, royal\}$ combined. We can measure such equivalence with respect to probability distributions over nearby words, in line with Firth's maxim "*You shall know a word by the company it keeps*" (Firth, 1957). We then show that paraphrasing can be reinterpreted as *word transformation* with additive *parameters* (e.g. from $man$ to $king$ by adding $royal$) and generalise to also allow subtraction. Finally, we prove that by interpreting an analogy "$w_a$ is to $w_{a^*}$ as $w_b$ is to $w_{b^*}$" as word transformations $w_a$ to $w_{a^*}$ and $w_b$ to $w_{b^*}$ sharing the same parameters, the linear relationship observed between word embeddings of analogies follows (see overview in Fig 4). Our key contributions are:

- to derive a probabilistic definition of *paraphrasing* and show that it governs the relationship between one (PMI-derived) word embedding and any sum of others;
- to show how paraphrasing can be generalised and interpreted as the *transformation* from one word to another, giving a mathematical formulation for "$w_x$ is to $w_{x^*}$";
- to provide the first rigorous proof of the linear relationship between word embeddings of analogies, including explicit, interpretable error terms; and
- to show how these relationships materialise between vectors of PMI values, and so too in word embeddings that factorise the PMI matrix, or approximate such a factorisation e.g. W2V and *Glove*.

## 2. Previous Work

Intuition for the presence of linear analogical relationships, or *linguistic regularity*, amongst word embeddings was first suggested by Mikolov et al. (2013a;b) and Pennington et al. (2014), and has been widely discussed since (e.g. Levy & Goldberg (2014a); Linzen (2016)). More recently, several theoretical explanations have been proposed:

---

[1]School of Informatics, University of Edinburgh. Correspondence to: Carl Allen <carl.allen@ed.ac.uk>.

[2]Throughout, we refer to the more commonly used *Skipgram* implementation of W2V with negative sampling (SGNS).

**Analogies Explained: Towards Understanding Word Embeddings**

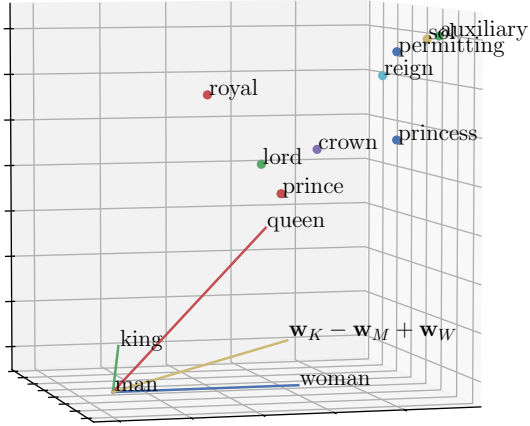

*Figure 1.* The relative locations of word embeddings for the analogy "*man* is to *king* as *woman* is to ..?". The closest embedding to the linear combination $\mathbf{w}_K - \mathbf{w}_M + \mathbf{w}_W$ is that of *queen*. We explain why this occurs and interpret the difference between them.

- Arora et al. (2016) propose a latent variable model for language that contains several strong *a priori* assumptions about the spatial distribution of word vectors, discussed by Gittens et al. (2017), that we do not require. Also, the two embedding matrices of W2V are assumed equal, which we show to be false in practice.

- Gittens et al. (2017) refer to *paraphrasing*, from which we draw inspiration, but make several assumptions that fail in practice: (i) that words follow a uniform distribution rather than the (highly non-uniform) Zipf distribution; (ii) that W2V learns a conditional distribution – violated by negative sampling (Levy & Goldberg, 2014b); and (iii) that joint probabilities beyond pairwise co-occurrences are zero.

- Ethayarajh et al. (2018) offer a recent explanation based on *co-occurrence shifted PMI*, however that property lacks motivation and several assumptions fail, e.g. it requires more than for opposite sides to have equal length to define a parallelogram in $\mathbb{R}^d$, $d > 2$ (their Lemma 1).

To our knowledge, no previous work mathematically interprets analogies so as to rigorously explain why if *"$w_a$ is to $w_{a^*}$ as $w_b$ is to $w_{b^*}$"* then a linear relationship manifests between corresponding word embeddings.

## 3. Background

The **Word2Vec** algorithm considers a set of word pairs $\{(w_{i_k}, c_{j_k})\}_k$ generated from a (typically large) text corpus, by allowing the *target* word $w_i$ to range over the corpus, and the *context* word $c_j$ to range over a context window (of size $l$) symmetric about the target word. For each observed word pair (*positive sample*), $k$ random word pairs (*negative samples*) are generated according to monogram distributions. The 2-layer "neural network" architecture simply multiplies two weight matrices $\mathbf{W}, \mathbf{C} \in \mathbb{R}^{d \times n}$, subject to a non-linear (sigmoid) function, where $d$ is the embedding dimensionality and $n$ is the size of $\mathcal{E}$ the dictionary of unique words in the corpus. Conventionally, $\mathbf{W}$ denotes the matrix closest to the input target words. Columns of $\mathbf{W}$ and $\mathbf{C}$ are the *embeddings* of words in $\mathcal{E}$: $\mathbf{w}_i \in \mathbb{R}^d$ ($i^{\text{th}}$ column of $\mathbf{W}$) corresponds to $w_i$ the $i^{th}$ word in $\mathcal{E}$ observed as a target word; and $\mathbf{c}_i \in \mathbb{R}^d$ ($i^{\text{th}}$ column of $\mathbf{C}$) corresponds to $c_i$, the same word when observed as a context word.

Levy & Goldberg (2014b) identified that the objective function for W2V is optimised if:

$$\mathbf{w}_i^\top \mathbf{c}_j \;=\; \text{PMI}(w_i, c_j) - \log k \,, \quad (1)$$

where $\text{PMI}(w_i, c_j) = \log \frac{p(w_i, c_j)}{p(w_i)p(c_j)}$ is known as *pointwise mutual information*. In matrix form, this equates to:

$$\mathbf{W}^\top \mathbf{C} \;=\; \mathbf{SPMI} \;\in \mathbb{R}^{n \times n} \,, \quad (2)$$

where $\mathbf{SPMI}_{i,j} = \text{PMI}(w_i, c_j) - \log k$, (*shifted* PMI).

*Glove* (Pennington et al., 2014) has the same architecture as W2V. Its embeddings perform comparably and also exhibit linear analogical structure. *Glove*'s loss function is optimised when:

$$\mathbf{w}_i^\top \mathbf{c}_j = \log p(w_i, c_j) - b_i - b_j + \log Z \quad (3)$$

for biases $b_i, b_j$ and normalising constant $Z$. (3) generalises (1) due to the biases, giving *Glove* greater flexibility than W2V and a potentially wider range of solutions. However, we will show that it is factorisation of the PMI matrix that causes linear analogical structure in embeddings, as approximately achieved by W2V (1). We conjecture that the same rationale underpins analogical structure in *Glove* embeddings, perhaps more weakly due to its increased flexibility.

## 4. Preliminaries

We consider pertinent aspects of the relationship between word embeddings and co-occurrence statistics (1, 2) relevant to the linear structure between embeddings of analogies:

**Impact of the *Shift*** As a chosen hyper-parameter, reflecting nothing of word properties, any effect on embeddings of $k$ appearing in (1) is arbitrary. Comparing typical values of $k$ with empirical PMI values (Fig 2), shows that the so-called *shift* ($-\log k$) may also be material. Further, it is observed that adjusting the W2V algorithm to avoid any direct impact of the *shift* improves embedding performance (Le, 2017). We conclude that the *shift* is a detrimental artefact of the W2V algorithm and, unless stated otherwise, consider embeddings that factorise the *unshifted* PMI matrix:

$$\mathbf{w}_i^\top \mathbf{c}_j \;=\; \text{PMI}(w_i, c_j) \quad \text{or} \quad \mathbf{W}^\top \mathbf{C} = \mathbf{PMI} \,. \quad (4)$$

**Analogies Explained: Towards Understanding Word Embeddings**

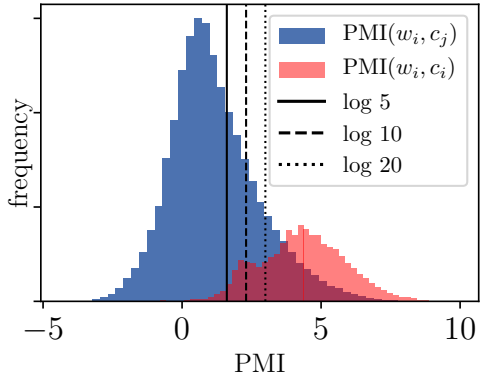

*Figure 2.* Histogram of PMI$(w_i, c_j)$ for word pairs randomly sampled from text (blue) with PMI$(w_i, c_i)$ for the *same word* overlaid (red, scale enlarged). The *shift* is material for typical values of $k$.

**Reconstruction Error** In practice, (2) and (4) hold only *approximately* since $\mathbf{W}^\top \mathbf{C} \in \mathbb{R}^{n \times n}$ is rank-constrained (rank $r \ll d < n$) relative to the factored matrix $\mathbf{M}$, e.g. $\mathbf{M} = \mathbf{PMI}$ in (4). Recovering elements of $\mathbf{M}$ from $\mathbf{W}$ and $\mathbf{C}$ is thus subject to *reconstruction error*. However, we rely throughout on linear relationships in $\mathbb{R}^n$, requiring only that they are sufficiently maintained when projected "down" into $\mathbb{R}^d$, the space of embeddings. To ensure this, we assume:

**A1.** $\mathbf{C}$ *has full row rank.*

**A2.** *Letting $\mathbf{M}_k$ denote the $k^{th}$ column of factored matrix $\mathbf{M} \in \mathbb{R}^{n \times n}$, the projection $f : \mathbb{R}^n \to \mathbb{R}^d$, $f(\mathbf{M}_i) = \mathbf{w}_i$ is approximately homomorphic with respect to addition, i.e. $f(\mathbf{M}_i + \mathbf{M}_j) \approx f(\mathbf{M}_i) + f(\mathbf{M}_j)$.*

A1 is reasonable since $d \ll n$ and $d$ is chosen. A2 means that, whatever the factorisation method used (e.g. analytic, W2V, *Glove*, weighted matrix factorisation (Srebro & Jaakkola, 2003)), linear relationships between columns of $\mathbf{M}$ are sufficiently preserved by columns of $\mathbf{W}$, i.e. the embeddings $\mathbf{w}_i$. For example, minimising a least squares loss function gives the linear projection $\mathbf{w}_i = f_{LSQ}(\mathbf{M}_i) = \mathbf{C}^\dagger \mathbf{M}_i$ for which A2 holds exactly (where $\mathbf{C}^\dagger = (\mathbf{C}\mathbf{C}^\top)^{-1}\mathbf{C}$, the *Moore-Penrose pseudo-inverse* of $\mathbf{C}^\top$, which exists by A1);[1] whereas for W2V, $\mathbf{w}_i = f_{W2V}(\mathbf{M}_i)$ is non-linear.[2]

**Zero Co-occurrence Counts** The co-occurrence of rare words are often unobserved, thus their empirical probability estimates zero and PMI estimates undefined. However, for a fixed dictionary $\mathcal{E}$, such zero counts decline as the corpus or context window size increase (the latter can be arbitrarily large if more distant words are down-weighted, e.g. Pennington et al. (2014)). Here, we consider small word sets $\mathcal{W}$ and assume the corpus and context window to be of sufficient size that the *true* values of considered probabilities are non-zero and their PMI values well-defined, i.e.:

**A3.** $p(\mathcal{W}) > 0, \ \forall \mathcal{W} \subseteq \mathcal{E}, |\mathcal{W}| < l,$

where (throughout) "$|\mathcal{W}| < l$" means $|\mathcal{W}|$ *sufficiently less than $l$.*

**The Relationship between W and C** Several works (e.g. Hashimoto et al. (2016); Arora et al. (2016)) assume embedding matrices $\mathbf{W}$ and $\mathbf{C}$ to be equal, i.e. $\mathbf{w}_i = \mathbf{c}_i \ \forall i$. The assumption is convenient as the number of parameters is halved, equations simplify and consideration of how to use $\mathbf{w}_i$ and $\mathbf{c}_i$ falls away. However, this implies $\mathbf{W}^\top \mathbf{W} = \mathbf{PMI}$, requiring $\mathbf{PMI}$ to be positive semi-definite, which is not true for typical corpora. Thus $\mathbf{w}_i$, $\mathbf{c}_i$ are not equal and modifying W2V to enforce them to be would unnecessarily constrain and may well worsen the low-rank approximation.

## 5. Paraphrases

Following a similar approach to Gittens et al. (2017), we consider a small set of target words $\mathcal{W} = \{w_1, \ldots, w_m\} \subseteq \mathcal{E}$, $|\mathcal{W}| < l$; and the sum of their embeddings $\mathbf{w}_\mathcal{W} = \sum_i \mathbf{w}_i$. In practice, we say word $w_* \in \mathcal{E}$ *paraphrases* $\mathcal{W}$ if $w_*$ and $\mathcal{W}$ are semantically interchangeable within the text, i.e. in circumstances where *all* $w_i \in \mathcal{W}$ appear, $w_*$ could appear instead. This suggests a relationship between the probability distributions $p(c_j|\mathcal{W})$ and $p(c_j|w_*)$, $\forall c_j \in \mathcal{E}$. We refer to such conditional distributions over all context words as the *distribution induced* by $\mathcal{W}$ or $w_*$, respectively.

### 5.1. Defining a Paraphrase

Let $\mathcal{C}_\mathcal{W} = \{c_{j_1}, \ldots, c_{j_t}\}$ be a sequence of words (with repetition) observed in the context of $\mathcal{W}$.[3] A *paraphrase word* $w_* \in \mathcal{E}$ can be thought of as that which *best explains* the observation of $\mathcal{C}_\mathcal{W}$. From a maximum likelihood perspective we have $w_*^{(1)} = \mathrm{argmax}_{w_i \in \mathcal{E}} \, p(\mathcal{C}_\mathcal{W}|w_i)$. Assuming $c_j \in \mathcal{C}_\mathcal{W}$ to be independent draws from $p(c_j|\mathcal{W})$, gives:

$$w_*^{(1)} = \underset{w_i}{\mathrm{argmax}} \ \prod_{c_j \in \mathcal{E}} p(c_j|w_i)^{\#_j}$$
$$\to \underset{w_i}{\mathrm{argmax}} \ \sum_{c_j \in \mathcal{E}} p(c_j|\mathcal{W}) \log p(c_j|w_i) \ ,$$

as $|\mathcal{C}_\mathcal{W}| \to \infty$, where $\#_j$ denotes the count of $c_j$ in $\mathcal{C}_\mathcal{W}$. It follows that $w_*^{(1)}$ minimises the Kullback-Leibler (KL) divergence $\Delta_{KL}^{\mathcal{W}, w_*}$ between the induced distributions, i.e.:

$$\Delta_{KL}^{\mathcal{W}, w_*} = D_{KL}[\, P(c_j|\mathcal{W}) \,||\, P(c_j|w_*) \,]$$
$$= \sum_j p(c_j|\mathcal{W}) \log \frac{p(c_j|\mathcal{W})}{p(c_j|w_*)} \ .$$

---

[1] *w.l.o.g.* we write $f(\cdot) = \mathbf{C}^\dagger(\cdot)$ throughout (except in specific cases) to emphasise linearity of the relationship.

[2] It is beyond the scope of this work to show A2 is satisfied when the W2V loss function is minimised (4). We instead prove existence of linear relationships in the full rank space of PMI columns, thus in linear projections thereof, and assume A2 holds sufficiently for W2V embeddings given (2) and empirical observation of linearity.

[3] By symmetry, $\mathcal{C}_\mathcal{W}$ is the set of target words for which all $w_i \in \mathcal{W}$ are simultaneously observed in the context window.

Alternatively, we might consider $w_*^{(2)}$, the target word whose set of associated context words $\mathcal{C}_{w_*}$ is best explained by $\mathcal{W}$, in the sense that $w_*^{(2)}$ minimises KL divergence $\Delta_{KL}^{w_*,\mathcal{W}} = D_{KL}[P(c_j|w_*) \| P(c_j|\mathcal{W})]$ (where, in general, $\Delta_{KL}^{\mathcal{W},w_*} \neq \Delta_{KL}^{w_*,\mathcal{W}}$). Interpretations of $w_*^{(1)}$ and $w_*^{(2)}$ are discussed in Appendix A. In each case, the KL divergence lower bound (zero) is achieved *iff* the induced distributions are equal, providing a theoretical basis for:

**Definition D1.** *We say word $w_* \in \mathcal{E}$ paraphrases word set $\mathcal{W} \subseteq \mathcal{E}$, $|\mathcal{W}| < l$, if the paraphrase error $\boldsymbol{\rho}^{\mathcal{W},w_*} \in \mathbb{R}^n$ is (element-wise) small, where:*

$$\rho_j^{\mathcal{W},w_*} = \log \frac{p(c_j|w_*)}{p(c_j|\mathcal{W})}, c_j \in \mathcal{E}.$$

Note that $\mathcal{W}$ and $w_*$ need not appear similarly often for $w_*$ to paraphrase $\mathcal{W}$, only amongst the same context words. We now connect paraphrasing, a semantic relationship, to relationships between word embeddings.

### 5.2. Paraphrase = Embedding Sum + Error

**Lemma 1.** *For any word $w_* \in \mathcal{E}$ and word set $\mathcal{W} \subseteq \mathcal{E}$, $|\mathcal{W}| < l$:*

$$\mathbf{PMI}_* = \sum_{w_i \in \mathcal{W}} \mathbf{PMI}_i + \boldsymbol{\rho}^{\mathcal{W},w_*} + \boldsymbol{\sigma}^{\mathcal{W}} - \tau^{\mathcal{W}} \mathbf{1}, \quad (5)$$

*where $\mathbf{PMI}_\bullet$ is the column of $\mathbf{PMI}$ corresponding to $w_\bullet \in \mathcal{E}$, $\mathbf{1} \in \mathbb{R}^n$ is a vector of 1s, and error terms $\sigma_j^{\mathcal{W}} = \log \frac{p(\mathcal{W}|c_j)}{\prod_i p(w_i|c_j)}$ and $\tau^{\mathcal{W}} = \log \frac{p(\mathcal{W})}{\prod_i p(w_i)}$.*

*Proof.* (See Appendix B.) As Lem 1 is central to what follows, we sketch its proof: a correspondence is drawn between the product of distributions induced by each $w_i \in \mathcal{W}$ (I) and the distribution induced by $w_*$ (II), by comparison to the distribution induced by joint event $\mathcal{W}$ (III), i.e. observing *all* $w_i \in \mathcal{W}$ in the context window. I relates to III by the (in)dependence of $w_i \in \mathcal{W}$ (i.e. by $\sigma_j^{\mathcal{W}}, \tau^{\mathcal{W}}$).[4] II relates to III by the paraphrase error $\rho_j^{\mathcal{W},w_*}$.   □

Following immediately from Lem 1 we have:

**Theorem 1** (Paraphrase). *For any word $w_* \in \mathcal{E}$ and word set $\mathcal{W} \subseteq \mathcal{E}$, $|\mathcal{W}| < l$:*

$$\mathbf{w}_* = \mathbf{w}_{\mathcal{W}} + \mathbf{C}^{\dagger}(\boldsymbol{\rho}^{\mathcal{W},w_*} + \boldsymbol{\sigma}^{\mathcal{W}} - \tau^{\mathcal{W}} \mathbf{1}), \quad (6)$$

*where $\mathbf{w}_{\mathcal{W}} = \sum_{w_i \in \mathcal{W}} \mathbf{w}_i$.*

*Proof.* Multiply (5) by $\mathbf{C}^{\dagger}$.   □

Thm 1 shows that an embedding (of $w_*$) and a sum of embeddings (of $\mathcal{W}$) differ by the paraphrase error $\rho^{\mathcal{W},w_*}$ between $w_*$ and $\mathcal{W}$; and $\boldsymbol{\sigma}^{\mathcal{W}}, \tau^{\mathcal{W}}$ (collectively *dependence error*) reflecting relationships within $\mathcal{W}$ (unrelated to $w_*$):

- $\boldsymbol{\sigma}^{\mathcal{W}}$ is a vector reflecting conditional dependencies within $\mathcal{W}$ given each $c_j \in \mathcal{E}$; $\sigma_j^{\mathcal{W}} = 0$ *iff* all $w_i \in \mathcal{W}$ are conditionally independent given each and every $c_j \in \mathcal{E}$;

- $\tau^{\mathcal{W}}$ is a scalar measure of mutual independence of $w_i \in \mathcal{W}$ (thus constant $\forall c_j \in \mathcal{E}$); $\tau^{\mathcal{W}} = 0$ *iff* $w_i \in \mathcal{W}$ are mutually independent.

**Corollary 1.1.** *A word set $\mathcal{W}$ has no associated dependence error iff $w_i \in \mathcal{W}$ are both mutually independent and conditionally independent given each context word $c_j \in \mathcal{E}$.*

Thm 1, which holds for all words $w_*$ and word sets $\mathcal{W}$, explains why and when a paraphrase (e.g. of $\{man, royal\}$ by $king$) can be identified by embedding addition ($\mathbf{w}_{man} + \mathbf{w}_{royal} \approx \mathbf{w}_{king}$). The phenomenon occurs due to a relationship between PMI vectors in $\mathbb{R}^n$ that holds for embeddings in $\mathbb{R}^d$ under projection by $\mathbf{C}^{\dagger}$ (by A1, A2). The vector error $\mathbf{w}_* - \mathbf{w}_{\mathcal{W}}$ depends on both the paraphrase relationship between $w_*$ and $\mathcal{W}$; and statistical dependencies within $\mathcal{W}$.

**Corollary 1.2.** *For word $w_* \in \mathcal{E}$ and word set $\mathcal{W} \subseteq \mathcal{E}$, $\mathbf{w}_* \approx \mathbf{w}_{\mathcal{W}}$ if $w_*$ paraphrases $\mathcal{W}$ and $w_i \in \mathcal{W}$ are materially independent (i.e. net dependence error is small).*

### 5.3. Do Linear Relationships Identify Paraphrases?

The converse of Cor 1.2 is false: $\mathbf{w}_* \approx \mathbf{w}_{\mathcal{W}}$ does not imply $w_*$ paraphrases $\mathcal{W}$. Specifically, *false positives* arise if: (i) paraphrase and dependence error terms are material but happen to cancel, i.e. *total error* $\boldsymbol{\rho}^{\mathcal{W},w_*} + \boldsymbol{\sigma}^{\mathcal{W}} - \tau^{\mathcal{W}} \mathbf{1} \approx \mathbf{0}$; or (ii) material components of the total error fall within the high $(n-d)$ dimensional null space of $\mathbf{C}^{\dagger}$ and project to a small vector difference between $\mathbf{w}_*$ and $\mathbf{w}_{\mathcal{W}}$. Case (i) can arise in PMI vectors (Lem 1) and thus lower rank embeddings also (Thm 1), but is highly unlikely in practice due to the high dimensionality $(n)$. Case (ii) can arise only in lower rank embeddings (Thm 1) and might be minimised by a good choice of factorisation or projection method.

### 5.4. Paraphrasing in *Explicit* Embeddings

Lem 1 applies to full rank **PMI** vectors, without reconstruction error or case (ii) false positives (Sec 5.3), explaining the linear relationships observed by Levy & Goldberg (2014a).

**Corollary 1.3.** *Thm 1 holds for* explicit *word embeddings, i.e. columns of* **PMI**.

*Proof.* Choose factorisation $\mathbf{W} = \mathbf{PMI}$, $\mathbf{C} = \mathbf{I}$ (the identity matrix) in Thm 1.   □

### 5.5. Paraphrasing in W2V Embeddings

Thm 1 extends to W2V embeddings by substituting $\mathbf{w}_i^{\top} \mathbf{c}_j = \text{PMI}(w_i, c_j) - \log k$ and $f_{W2V}$:

**Corollary 1.4.** *Under conditions of Thm 1, W2V embeddings satisfy:*

$$\mathbf{w}_* = \mathbf{w}_{\mathcal{W}} + f_{W2V}\big(\boldsymbol{\rho}^{\mathcal{W},w_*} + \boldsymbol{\sigma}^{\mathcal{W}} - \tau^{\mathcal{W}} \mathbf{1} + \log k(|\mathcal{W}|-1)\mathbf{1}\big). \quad (7)$$

---
[4] Analogous to a product of marginal probabilities relating to their joint probability subject to independence.

Comparing (6) and (7) shows that paraphrases correspond to linear relationships in W2V embeddings with an additional error term linear in $|\mathcal{W}|$, and hence with less accuracy if $|\mathcal{W}|>1$, than for embeddings that factorise **PMI**.

## 6. Analogies

An *analogy* is said to hold for words $w_a, w_{a^*}, w_b, w_{b^*} \in \mathcal{E}$ if, in some sense, *"$w_a$ is to $w_{a^*}$ as $w_b$ is to $w_{b^*}$"*. Since in principle the same relationship may extend further ("... as $w_c$ is to $w_{c^*}$" etc), we characterise a general analogy $\mathfrak{A}$ by a set of ordered word pairs $S_\mathfrak{A} \subseteq \mathcal{E} \times \mathcal{E}$, where $(w_x, w_{x^*}) \in S_\mathfrak{A}$, $w_x, w_{x^*} \in \mathcal{E}$, iff *"$w_x$ is to $w_{x^*}$ as ... [all other analogical pairs]"* under $\mathfrak{A}$. Our aim is to explain why respective word embeddings often satisfy:

$$\mathbf{w}_{b^*} \approx \mathbf{w}_{a^*} - \mathbf{w}_a + \mathbf{w}_b, \qquad (8)$$

or why in the more general case:

$$\mathbf{w}_{x^*} - \mathbf{w}_x \approx \mathbf{u}_\mathfrak{A}, \qquad (9)$$

$\forall (w_x, w_{x^*}) \in S_\mathfrak{A}$ and vector $\mathbf{u}_\mathfrak{A} \in \mathbb{R}^n$ specific to $\mathfrak{A}$.

We split the task of understanding why analogies give rise to Equations 8 and 9 into: **Q1**) understanding conditions under which word embeddings can be added and subtracted to approximate other embeddings; **Q2**) establishing a mathematical interpretation of *"$w_x$ is to $w_{x^*}$"*; and **Q3**) drawing a correspondence between those results. We show that all of these can be answered with paraphrasing by generalising the notion to word sets.

### 6.1. Paraphrasing Word Sets

**Definition D2.** *We say word set $\mathcal{W}_* \subseteq \mathcal{E}$ paraphrases word set $\mathcal{W} \subseteq \mathcal{E}$, $|\mathcal{W}|, |\mathcal{W}_*| < l$, if paraphrase error $\boldsymbol{\rho}^{\mathcal{W},\mathcal{W}_*} \in \mathbb{R}^n$ is (element-wise) small, where:*

$$\boldsymbol{\rho}_j^{\mathcal{W},\mathcal{W}_*} = \log \frac{p(c_j|\mathcal{W}_*)}{p(c_j|\mathcal{W})}, c_j \in \mathcal{E}.$$

D2 generalises D1 such that the paraphrase term $\mathcal{W}_*$, previously $w_*$, can be more than one word.[5] Analogously to D1, word sets paraphrase one another if they induce equivalent distributions over context words. Note that paraphrasing under D2 is both reflexive and symmetric (since $|\boldsymbol{\rho}^{\mathcal{W},\mathcal{W}_*}| = |\boldsymbol{\rho}^{\mathcal{W}_*,\mathcal{W}}|$), thus "$\mathcal{W}_*$ paraphrases $\mathcal{W}$" and "$\mathcal{W}$ paraphrases $\mathcal{W}_*$" are equivalent and denoted $\mathcal{W} \approx_\text{P} \mathcal{W}_*$.

Analogues of Lem 1 and Thm 1 follow:

**Lemma 2.** *For any word sets $\mathcal{W}, \mathcal{W}_* \subseteq \mathcal{E}, |\mathcal{W}|, |\mathcal{W}_*| < l$:*

$$\sum_{w_i \in \mathcal{W}_*} \text{PMI}_i = \sum_{w_i \in \mathcal{W}} \text{PMI}_i + \boldsymbol{\rho}^{\mathcal{W},\mathcal{W}_*} + \boldsymbol{\sigma}^{\mathcal{W}} - \boldsymbol{\sigma}^{\mathcal{W}_*}$$
$$- (\tau^{\mathcal{W}} - \tau^{\mathcal{W}_*})\mathbf{1}. \qquad (10)$$

*Proof.* (See Appendix C.) $\square$

---
[5]Equivalently, D1 is a special case of D2 with $|\mathcal{W}_*| = 1$, hence we reuse terms without ambiguity.

**Theorem 2** (Generalised Paraphrase). *For any word sets $\mathcal{W}, \mathcal{W}_* \subseteq \mathcal{E}, |\mathcal{W}|, |\mathcal{W}_*| < l$:*

$$\mathbf{w}_{\mathcal{W}_*} = \mathbf{w}_\mathcal{W} + \mathbf{C}^\dagger(\boldsymbol{\rho}^{\mathcal{W},\mathcal{W}_*} + \boldsymbol{\sigma}^{\mathcal{W}} - \boldsymbol{\sigma}^{\mathcal{W}_*} - (\tau^{\mathcal{W}} - \tau^{\mathcal{W}_*})\mathbf{1}).$$

*Proof.* Multiply (10) by $\mathbf{C}^\dagger$. $\square$

Note that $|\mathcal{W}_*| = 1$ recovers Lem 1 and Thm 1. With analogies in mind, we restate Thm 2 as:

**Corollary 2.1.** *For any words $w_x, w_{x^*} \in \mathcal{E}$ and word sets $\mathcal{W}^+, \mathcal{W}^- \subseteq \mathcal{E}, |\mathcal{W}^+|, |\mathcal{W}^-| < l - 1$:*

$$\mathbf{w}_{x^*} = \mathbf{w}_x + \mathbf{w}_{\mathcal{W}^+} - \mathbf{w}_{\mathcal{W}^-} + \mathbf{C}^\dagger(\boldsymbol{\rho}^{\mathcal{W},\mathcal{W}_*} + \boldsymbol{\sigma}^{\mathcal{W}} - \boldsymbol{\sigma}^{\mathcal{W}_*}$$
$$- (\tau^{\mathcal{W}} - \tau^{\mathcal{W}_*})\mathbf{1}), \qquad (11)$$

*where $\mathcal{W} = \{w_x\} \cup \mathcal{W}^+$, $\mathcal{W}_* = \{w_{x^*}\} \cup \mathcal{W}^-$.*

*Proof.* Set $\mathcal{W} = \{w_x\} \cup \mathcal{W}^+$, $\mathcal{W}_* = \{w_{x^*}\} \cup \mathcal{W}^-$ in Thm 2. $\square$

Cor 2.1 shows how any word embedding $\mathbf{w}_{x^*}$ relates to a linear combination of other embeddings ($\mathbf{w}_\Sigma = \mathbf{w}_x + \mathbf{w}_{\mathcal{W}^+} - \mathbf{w}_{\mathcal{W}^-}$), due to an equivalent relationship between columns of **PMI**. Analogously to one-word (D1) paraphrases, the vector difference $\mathbf{w}_{x^*} - \mathbf{w}_\Sigma$ depends on the paraphrase error that reflects the relationship <u>between</u> the two word sets $\mathcal{W}_*, \mathcal{W}$; and the dependence error that reflects statistical dependence between words <u>within</u> each of $\mathcal{W}$ and $\mathcal{W}_*$.

**Corollary 2.2.** *For terms as defined above, $\mathbf{w}_{x^*} \approx \mathbf{w}_x + \mathbf{w}_{\mathcal{W}^+} - \mathbf{w}_{\mathcal{W}^-}$ if $\mathcal{W}_* \approx_\text{P} \mathcal{W}$ and $w_i \in \mathcal{W}$ and $w_i \in \mathcal{W}_*$ are materially independent or dependence terms materially cancel.*

False positives can arise as discussed in Sec 5.3.

### 6.2. From Paraphrases to Analogies

A special case of Cor 2.1 gives:

**Corollary 2.3.** *For any $w_a, w_{a^*}, w_b, w_{b^*} \in \mathcal{E}$:*

$$\mathbf{w}_{b^*} = \mathbf{w}_{a^*} - \mathbf{w}_a + \mathbf{w}_b + \mathbf{C}^\dagger(\boldsymbol{\rho}^{\mathcal{W},\mathcal{W}_*} + \boldsymbol{\sigma}^{\mathcal{W}} - \boldsymbol{\sigma}^{\mathcal{W}_*}$$
$$- (\tau^{\mathcal{W}} - \tau^{\mathcal{W}_*})\mathbf{1}), \qquad (12)$$

*where $\mathcal{W} = \{w_b, w_{a^*}\}$ and $\mathcal{W}_* = \{w_{b^*}, w_a\}$.*

*Proof.* Set $w_x = w_b$, $w_{x^*} = w_{b^*}$, $\mathcal{W}^+ = \{w_{a^*}\}$, $\mathcal{W}^- = \{w_a\}$ in Cor 2.1. $\square$

Thus we see that (8) holds *if* $\{w_{b^*}, w_a\} \approx_\text{P} \{w_b, w_{a^*}\}$ and those word pairs exhibit *similar dependence* (Sec 6.6). More generally, by Cor 2.1 we see that (9) is satisfied by $\mathbf{u}_\mathfrak{A} \approx \mathbf{w}_{\mathcal{W}^+} - \mathbf{w}_{\mathcal{W}^-}$ *if* $\{w_{x^*}, \mathcal{W}^-\} \approx_\text{P} \{w_x, \mathcal{W}^+\}$ $\forall (w_x, w_{x^*}) \in S_\mathfrak{A}$ for *common* word sets $\mathcal{W}^+, \mathcal{W}^- \subseteq \mathcal{E}$ and each pair of paraphrasing word sets exhibit similar dependence.

This establishes sufficient conditions for the linear relationships observed in analogy embeddings (8, 9) in terms of

**Analogies Explained: Towards Understanding Word Embeddings**

semantic relationships, answering Q1. However, those relationships are *paraphrases*, with no obvious connection to the "$w_x$ *is to* $w_{x^*}$..." relationships of analogies. We now show that paraphrases sufficient for (8, 9) correspond to analogies by introducing the concept of *word transformation*.

### 6.3. Word Transformation

The paraphrase of a word set $\mathcal{W}$ by word $w_*$ (D1) has, so far, been considered in terms of an equivalence between $\mathcal{W}$ and $w_*$ by reference to their induced distributions. Alternatively, that paraphrase can be interpreted as a *transformation* from an arbitrary $w_s \in \mathcal{W}$ to $w_*$ by adding words $\mathcal{W}^+ = \{w_i \in \mathcal{W}, w_i \neq w_s\}$. Notionally, $\mathcal{W}^+$ can be considered "words that make $w_s$ more like $w_*$". More precisely, $w_i \in \mathcal{W}^+$ *add context* to $w_s$: we move from a distribution induced by $w_s$ alone to one induced by the *joint* event of simultaneously observing $w_s$ and all $w_i \in \mathcal{W}^+$, a *contextualised* occurrence of $w_s$ with an induced distribution closer that of $w_*$. A similar view can be taken of the associated embedding addition: starting with $\mathbf{w}_s$, add $\mathbf{w}_i \ \forall w_i \in \mathcal{W}^+$ to approximate $\mathbf{w}_*$. Note that only *addition* applies.

Moving to D2, the paraphrase of one word set $\mathcal{W}$ by another $\mathcal{W}_*$ can be interpreted additively as starting with some $w_x \in \mathcal{W}$, $w_{x^*} \in \mathcal{W}_*$, and adding $\mathcal{W}^+ = \{w_i \in \mathcal{W}, w_i \neq w_x\}$, $\mathcal{W}^- = \{w_i \in \mathcal{W}_*, w_i \neq w_{x^*}\}$, respectively, such that the resulting sets $\mathcal{W}$ and $\mathcal{W}_*$ induce similar distributions, i.e. paraphrase. In effect, context is added to both $w_x$ and $w_{x^*}$ until their contextualised cases $\mathcal{W}$ and $\mathcal{W}_*$ paraphrase (Fig 3a). Note $\mathcal{W}$ and $\mathcal{W}_*$ may have no intuitive meaning and need not correspond to a single word, unlike D1 paraphrases. Alternatively, such a paraphrase can be interpreted as a transformation from $w_x \in \mathcal{W}$ to $w_{x^*} \in \mathcal{W}^*$ by adding $w_i \in \mathcal{W}^+$ and *subtracting* $w_i \in \mathcal{W}^-$. "Subtraction" is effected by *adding words to the other side*, i.e. to $w_{x^*}$.[6] Just as adding words to $w_x$ adds or *narrows* its context, subtracting words removes or *broadens* context. Context is thus added and removed to transform from $w_x$ to $w_{x^*}$, in which the paraphrase between $\mathcal{W}$ and $\mathcal{W}_*$ effectively serves as an intermediate step (Fig 3b). We refer to $\mathcal{W}^+$, $\mathcal{W}^-$ as *transformation parameters*, which can be thought of as *explaining the difference* between $w_x$ and $w_{x^*}$ with a "richer dictionary" than that available to D1 paraphrases by including *differences* between words. More precisely, transformation parameters align the induced distributions to create a paraphrase.

This interpretation show equivalence between a paraphrase $\mathcal{W} \approx_\text{P} \mathcal{W}_*$ and a word transformation – a relationship between $w_x \in \mathcal{W}$ and $w_{x^*} \in \mathcal{W}_*$ based on the addition and subtraction of context that is mirrored in the addition and subtraction of embeddings. Mathematical equivalence of the perspectives is reinforced by an alternate proof of Cor 2.1

---
[6]Analogous to standard algebra: if $x < y$, equality is achieved either by adding to $x$ or by subtracting from $y$.

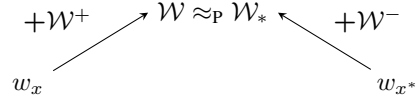

(a) Adding context to each of $w_x$ and $w_{x^*}$ to reach a paraphrase.

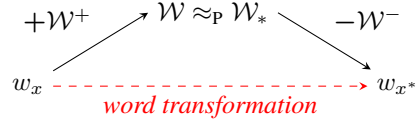

(b) Adding and subtracting context to *transform* $w_x$ to $w_{x^*}$.

*Figure 3.* Perspectives of the paraphrase $\mathcal{W} \approx_\text{P} \mathcal{W}_*$.

in Appendix D that begins with terms in only $w_x$ and $w_{x^*}$, highlighting that *any* words $\mathcal{W}^+$, $\mathcal{W}^-$ can be introduced, but only certain choices form the necessary paraphrase.

**Definition D 3.** *There exists a word transformation from $w_x \in \mathcal{E}$ to $w_{x^*} \in \mathcal{E}$ with transformation parameters $\mathcal{W}^+$, $\mathcal{W}^- \subseteq \mathcal{E}$ iff $\{w_x\} \cup \mathcal{W}^+ \approx_\text{P} \{w_{x^*}\} \cup \mathcal{W}^-$.*

Note that transformation parameters may not be unique and always (trivially) include $\mathcal{W}^+ = \{w_{x^*}\}$, $\mathcal{W}^- = \{w_x\}$.

### 6.4. Interpreting "*a is to a* as b is to b**"

With word transformation as a means of describing semantic difference between words, we mathematically interpret analogies. Specifically, we consider "$w_x$ *is to* $w_{x^*}$" to refer to a transformation from $w_x$ to $w_{x^*}$ and an analogy to require an equivalence between such word transformations.

**Definition D 4.** *We say "$w_a$ is to $w_{a^*}$ as $w_b$ is to $w_{b^*}$" for $w_a, w_b, w_{a^*}, w_{b^*} \in \mathcal{E}$ iff there exist parameters $\mathcal{W}^+, \mathcal{W}^- \subseteq \mathcal{E}$ that simultaneously transform $w_a$ to $w_{a^*}$ and $w_b$ to $w_{b^*}$.*

We show that the linear relationships between word embeddings of analogies (8, 9) follow from D4.

**Lemma 3.** *If "$w_a$ is to $w_{a^*}$ as $w_b$ is to $w_{b^*}$" by D4 with transformation parameters $\mathcal{W}^+, \mathcal{W}^- \subseteq \mathcal{E}$, then:*

$$\begin{aligned}
\text{PMI}_{b^*} = \text{PMI}_{a^*} &- \text{PMI}_a + \text{PMI}_b \\
&+ \boldsymbol{\rho}^{\mathcal{W}^b, \mathcal{W}^b_*} - \boldsymbol{\rho}^{\mathcal{W}^a, \mathcal{W}^a_*} \\
&+ (\boldsymbol{\sigma}^{\mathcal{W}^b} - \boldsymbol{\sigma}^{\mathcal{W}^b_*}) - (\boldsymbol{\sigma}^{\mathcal{W}^a} - \boldsymbol{\sigma}^{\mathcal{W}^a_*}) \\
&- ((\boldsymbol{\tau}^{\mathcal{W}^b} - \boldsymbol{\tau}^{\mathcal{W}^b_*}) - (\boldsymbol{\tau}^{\mathcal{W}^a} - \boldsymbol{\tau}^{\mathcal{W}^a_*}))\mathbf{1}, \quad (13)
\end{aligned}$$

*where $\mathcal{W}^x = \{w_x\} \cup \mathcal{W}^+$, $\mathcal{W}^x_* = \{w_{x^*}\} \cup \mathcal{W}^-$ for $x \in \{a, b\}$ and $\boldsymbol{\rho}^{\mathcal{W}^b, \mathcal{W}^b_*}, \boldsymbol{\rho}^{\mathcal{W}^a, \mathcal{W}^a_*}$ are small.*

*Proof.* Let $\mathcal{W} = \mathcal{W}^x$, $\mathcal{W}_* = \mathcal{W}^x_*$ for $x \in \{a, b\}$ in instances of Cor 2.1 and take the difference. $\mathcal{W}^x$ paraphrases $\mathcal{W}^x_*$ for $x \in \{a, b\}$ by D3 and D4. □

$$\begin{array}{ccccccc}
\text{``}w_a \text{ is to } w_{a^*} & & w_a \xrightarrow[\mathcal{W}^-]{\mathcal{W}^+} w_{a^*} & & \{w_a, \mathcal{W}^+\} \approx_{\text{P}} \{w_{a^*}, \mathcal{W}^-\} & & \mathbf{w}_{a^*} - \mathbf{w}_a \\
\text{as} & \Longleftrightarrow & \wedge & \Longleftrightarrow & \wedge & \Longrightarrow & \approx \\
w_b \text{ is to } w_{b^*}\text{''} & & w_b \xrightarrow[\mathcal{W}^-]{\mathcal{W}^+} w_{b^*} & & \{w_b, \mathcal{W}^+\} \approx_{\text{P}} \{w_{b^*}, \mathcal{W}^-\} & & \mathbf{w}_{b^*} - \mathbf{w}_b
\end{array}$$

*Figure 4.* Summary of steps to prove the relationship between analogies and word embeddings (omitting dependence error). $w_x \xrightarrow[\mathcal{W}^-]{\mathcal{W}^+} w_{x^*}$ denotes a word transformation $w_x$ to $w_{x^*}$ with parameters $\mathcal{W}^+, \mathcal{W}^- \subseteq \mathcal{E}$.

**Theorem 3** (Analogies). *If "$w_a$ is to $w_{a^*}$ as $w_b$ is to $w_{b^*}$" by D4 with $\mathcal{W}^+, \mathcal{W}^- \subseteq \mathcal{E}$, then:*

$$\begin{aligned}
\mathbf{w}_{b^*} = \ & \mathbf{w}_{a^*} - \mathbf{w}_a + \mathbf{w}_b \\
& + \mathbf{C}^\dagger (\boldsymbol{\rho}^{\mathcal{W}^b, \mathcal{W}_*^b} - \boldsymbol{\rho}^{\mathcal{W}^a, \mathcal{W}_*^a}) \\
& + (\boldsymbol{\sigma}^{\mathcal{W}^b} - \boldsymbol{\sigma}^{\mathcal{W}_*^b}) - (\boldsymbol{\sigma}^{\mathcal{W}^a} - \boldsymbol{\sigma}^{\mathcal{W}_*^a}) \\
& - ((\tau^{\mathcal{W}^b} - \tau^{\mathcal{W}_*^b}) - (\tau^{\mathcal{W}^a} - \tau^{\mathcal{W}_*^a}))\mathbf{1}).
\end{aligned}$$

*with terms as defined in Lem 3.*

*Proof.* Multiply (13) by $\mathbf{C}^\dagger$. □

More generally, if D4 applies for a set of ordered word pairs $S = \{(w_x, w_{x^*})\}$, i.e. "$w_a$ is to $w_{a^*}$ as $w_b$ is to $w_{b^*}$" $\forall (w_a, w_{a^*}), (w_b, w_{b^*}) \in S$ with transformation parameters $\mathcal{W}^+, \mathcal{W}^- \subseteq \mathcal{E}$, then each set $\{w_{x^*}, \mathcal{W}^-\}$ must paraphrase $\{w_x, \mathcal{W}^+\}$ by D3, and (11) holds with small paraphrase error. By this and Thm 3 we know that word embeddings of an analogy $\mathbf{w}_a, \mathbf{w}_b, \mathbf{w}_{a^*}, \mathbf{w}_{b^*}$ satisfy linear relationships (8, 9), subject to dependence error.

A few questions remain: how to find appropriate transformation parameters; and, given non-uniqueness, which to choose? Addressing these in reverse order:

**Transformation Parameter Equivalence**

By Lem 3, if "$w_a$ is to $w_{a^*}$ as $w_b$ is to $w_{b^*}$" then, subject to dependence error:

$$\text{PMI}_{b^*} - \text{PMI}_b \approx \text{PMI}_{a^*} - \text{PMI}_a \ . \quad (14)$$

If parameters $\mathcal{W}_2^+, \mathcal{W}_2^-$ exist that (*w.l.o.g.*) transform $w_a$ to $w_{a^*}$ then (13) holds by suitably redefining $\mathcal{W}^x, \mathcal{W}_*^x$, in which $\boldsymbol{\rho}^{\mathcal{W}^a, \mathcal{W}_*^a}$ is small but nothing is known of $\boldsymbol{\rho}^{\mathcal{W}^b, \mathcal{W}_*^b}$. Thus, subject to dependence error:

$$\text{PMI}_{b^*} - \text{PMI}_b \approx \text{PMI}_{a^*} - \text{PMI}_a + \boldsymbol{\rho}^{\mathcal{W}^b, \mathcal{W}_*^b} \ . \quad (15)$$

By (14), (15), subject to dependence error, $\boldsymbol{\rho}^{\mathcal{W}^b, \mathcal{W}_*^b}$ is also small and $\mathcal{W}_2^+, \mathcal{W}_2^-$ must also transform $w_b$ to $w_{b^*}$. Thus transformation parameters of any analogical pair transform all pairs and all applicable transformation parameters can be considered equivalent, up to dependence error.

**Corollary 3.1.** *For analogy $\mathfrak{A}$, if parameters $\mathcal{W}^+, \mathcal{W}^- \subseteq \mathcal{E}$ transform $w_x$ to $w_{x^*}$ for any $(w_x, w_{x^*}) \in S_\mathfrak{A}$, then $\mathcal{W}^+, \mathcal{W}^-$ simultaneously transform $w_x$ to $w_{x^*}$ $\forall (w_x, w_{x^*}) \in S_\mathfrak{A}$.*

**Identifying Transformation Parameters**

To identify "words that explain the difference between other words" might, in general, be non-trivial. However, by Cor 3.1, transformation parameters for analogy $\mathfrak{A}$ can simply be chosen as $\mathcal{W}^+ = \{w_{x^*}\}, \mathcal{W}^- = \{w_x\}$ for any $(w_x, w_{x^*}) \in S_\mathfrak{A}$.[7] Making an arbitrary choice, Thm 3 simplifies to:

**Corollary 3.2.** *If "$w_a$ is to $w_{a^*}$ as $w_b$ is to $w_{b^*}$" then:*

$$\begin{aligned}
\mathbf{w}_{b^*} = \ & \mathbf{w}_{a^*} - \mathbf{w}_a + \mathbf{w}_b + \mathbf{C}^\dagger (\boldsymbol{\rho}^{\mathcal{W},\mathcal{W}_*} + \boldsymbol{\sigma}^\mathcal{W} - \boldsymbol{\sigma}^{\mathcal{W}_*} \\
& - (\tau^\mathcal{W} - \tau^{\mathcal{W}_*})\mathbf{1}), \quad (16)
\end{aligned}$$

*where $\mathcal{W} = \{w_b, w_{a^*}\}, \mathcal{W}_* = \{w_{b^*}, w_a\}$ and $\boldsymbol{\rho}^{\mathcal{W},\mathcal{W}_*}$ is small.*

*Proof.* Let $\mathcal{W}^+ = \{w_{a^*}\}, \mathcal{W}^- = \{w_a\}$ in Thm 3. □

We arrive back at (12) but now link directly to analogies, proving that word embeddings of analogies satisfy linear relationships (8) and (9), subject to dependence error. Fig 4 shows a summary of all steps to prove Cor 3.2. D4 also provides a mathematical interpretation of what we mean when we say "$w_a$ is to $w_{a^*}$ as $w_b$ is to $w_{b^*}$".

**6.5. Example**

To demonstrate the concepts developed, we consider the canonical analogy $\mathfrak{A}^*$: *"man is to king as woman is to queen"*, for which $S_{\mathfrak{A}^*} = \{(man, king), (woman, queen)\}$. By D4, there exist parameters $\mathcal{W}^+, \mathcal{W}^- \subseteq \mathcal{E}$ that simultaneously transform *man* to *king* and *woman* to *queen*, which (by Cor 3.1) can be chosen to be $\mathcal{W}^+ = \{queen\}, \mathcal{W}^- = \{woman\}$. Thus $\mathfrak{A}^*$ implies that $\{man, queen\} \approx_\text{P} \{king, woman\}$ and $\{woman, queen\} \approx_\text{P} \{queen, woman\}$, the latter being trivially true. By Cor 2.1, $\mathfrak{A}^*$ therefore implies:

$$\mathbf{w}_Q = \mathbf{w}_K - \mathbf{w}_M + \mathbf{w}_W + \mathbf{C}^\dagger(\boldsymbol{\rho}^{\mathcal{W},\mathcal{W}_*} + \boldsymbol{\sigma}^\mathcal{W} - \boldsymbol{\sigma}^{\mathcal{W}_*} \\ - (\tau^\mathcal{W} - \tau^{\mathcal{W}_*})\mathbf{1}) \ ,$$

where we abbreviate words by their initials and, explicitly:

$$\boldsymbol{\rho}^{\mathcal{W},\mathcal{W}_*} = \log \tfrac{p(c_j|w_Q, w_M)}{p(c_j|w_W, w_K)} \quad \text{(which must be small)},$$

$$\boldsymbol{\sigma}^\mathcal{W} = \log \tfrac{p(w_W, w_K|c_j)}{p(w_W|c_j)p(w_K|c_j)}, \quad \tau^\mathcal{W} = \log \tfrac{p(w_W, w_K)}{p(w_W)p(w_K)},$$

$$\boldsymbol{\sigma}^{\mathcal{W}_*} = \log \tfrac{p(w_Q, w_M|c_j)}{p(w_Q|c_j)p(w_M|c_j)}, \quad \tau^{\mathcal{W}_*} = \log \tfrac{p(w_Q, w_M)}{p(w_Q)p(w_M)} \ .$$

---

[7] In the case of an analogical question *"$w_a$ is to $w_{a^*}$ as $w_b$ is to ... ?"*, there is only one choice: $\mathcal{W}^+ = \{w_{a^*}\}, \mathcal{W}^- = \{w_a\}$.

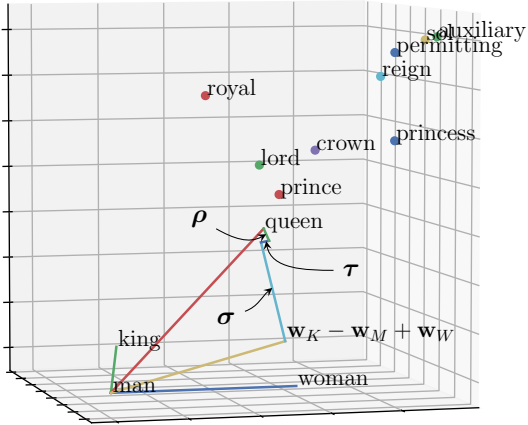

*Figure 5.* The plot shows the same embeddings of Fig 1, now with the difference between $\mathbf{w}_K - \mathbf{w}_M + \mathbf{w}_W$ and the embedding of *queen* explained (see connecting "zigzag") as the sum of conditional independence error ($\boldsymbol{\sigma}$), independence error ($\boldsymbol{\tau} = \tau\mathbf{1}$) and paraphrase error ($\boldsymbol{\rho}$). As anticipated, their sum is smallest for *queen*. Related words are seen nearby, with unrelated words clustered further away. Plot generated by fixing the $xy$ plane to contain *man*, *king*, *queen* and all other vectors plotted relatively, i.e. the $z$-axis captures any component off the $xy$-plane. Values are computed from the "text8" corpus (Mahoney, 2011).

Thus $\mathbf{w}_Q \approx \mathbf{w}_K - \mathbf{w}_M + \mathbf{w}_W$ subject to the accuracy with which $\{man, queen\}$ paraphrases $\{king, woman\}$ and statistical dependencies within those word pairs (see Fig 5).

### 6.6. Dependence error in analogies

Dependence error terms for analogies (13) bear an important distinction from those in one-word paraphrases (5). When a word set $\mathcal{W}$ is paraphrased by a single word $w_*$, the dependence error comprises a conditional independence term ($\boldsymbol{\sigma}^{\mathcal{W}}$) and a mutual independence term ($\tau^{\mathcal{W}}\mathbf{1}$) that bear no obvious relationship to one another and can only cancel by chance, which is low in high dimensions. However, (13) contains offsetting pairs of each component ($\boldsymbol{\sigma}^{\mathcal{W}}, \boldsymbol{\sigma}^{\mathcal{W}_*}, \tau^{\mathcal{W}}, \tau^{\mathcal{W}_*}$), i.e. terms of the same form that may cancel, thus word sets with *similar dependence terms* will paraphrase with small overall dependence error.

It is illustrative to consider the case $w_a = w_b$, $w_{a^*} = w_{b^*}$, corresponding to the trivial analogy "$w_a$ is to $w_{a^*}$ as "$w_a$ is to $w_{a^*}$", which holds true with zero total error for any word pair. Considering specific error terms: the paraphrase error is zero since $p(c_j|\{w_a, w_{a^*}\}) = p(c_j|\{w_{a^*}, w_a\})$, $\forall c_j \in \mathcal{E}$, thus the net dependence error is also zero. However, individual dependence error terms, e.g. $\log \frac{p(w_a, w_{a^*})}{p(w_a)p(w_{a^*})}$, are generally non-zero. This therefore proves existence of a case in which non-zero dependence error terms negate one another to give a negligible net dependence error.

### 6.7. Analogies in *explicit* embeddings

As with paraphrases, analogical relationships in embeddings stem from relationships between columns of $\mathbf{PMI}$.

**Corollary 3.3.** *Cor 3.2 applies to explicit (full-rank) embeddings, i.e. columns of $\mathbf{PMI}$, with $\mathbf{C} = \mathbf{I}$ (the identity matrix).*

### 6.8. Analogies in W2V embeddings

As with paraphrases (Sec 5.5), the results for analogies can be extended to W2V embeddings by including the *shift* term appropriately throughout. Since the transformation parameters for analogies are of equal size (i.e. $|\mathcal{W}^+| = |\mathcal{W}^-| = 1$), we find that all *shift* terms cancel.

**Corollary 3.4.** *Cor 3.2 applies to W2V embeddings replacing the projection $\mathbf{C}^\dagger(\cdot)$ with $f_{W2V}(\cdot)$.*

Thus, linear relationships between embeddings for analogies hold equally for W2V embeddings as for those derived without the *shift* distortion. Whilst perhaps surprising, this is corroborative since linear analogical relationships have been observed extensively in W2V embeddings (e.g. Levy & Goldberg (2014a)), as is now justified theoretically. Thus we know that analogies hold for W2V embeddings subject to higher order statistical relationships between words of the analogy as defined by the paraphrase and dependence errors.

## 7. Conclusion

In this work, we develop a probabilistically principled definition of *paraphrasing* by which equivalence is drawn between words and word sets by reference to the distributions they induce over words around them. We prove that, subject to statistical dependencies, paraphrase relationships give rise to linear relationships between word embeddings that factorise PMI (including columns of the PMI matrix), and thus others that approximate such a factorisation, e.g. W2V and *Glove*. By showing that paraphrases can be interpreted as *word transformations*, we enable analogies to be mathematically defined and, thereby, properties of semantics to be translated into properties of word embeddings. This provides the first rigorous explanation for the presence of linear relationships between the word embeddings of analogies.

In future work we aim to extend our understanding of the relationships between word embeddings to other applications of discrete object representation that rely on an underlying matrix factorisation, e.g. graph embeddings and recommender systems. Also, word embeddings are known to capture stereotypes present in corpora (Bolukbasi et al. (2016)) and future work may look at developing our understanding of embedding composition to foster principled methods to correct or *debias* embeddings.

**Acknowledgements**

We thank Ivana Balažević and Jonathan Mallinson for helpful comments on this manuscript. Carl Allen was supported by the Centre for Doctoral Training in Data Science, funded by EPSRC (grant EP/L016427/1) and the University of Edinburgh.

# Appendices

## A. The KL-divergence between induced distributions

We consider the words found by minimising the difference KL-divergences considered in Section 5. Specifically:

$$w_*^{(1)} = \underset{w_i \in \mathcal{E}}{\mathrm{argmin}}\, D_{KL}[\,p(c_j|\mathcal{W})\,||\,p(c_j|w_i)\,]$$
$$w_*^{(2)} = \underset{w_i \in \mathcal{E}}{\mathrm{argmin}}\, D_{KL}[\,p(c_j|w_i)\,||\,p(c_j|\mathcal{W})\,]$$

Minimising $D_{KL}[\,p(c_j|\mathcal{W})\,||\,p(c_j|w_i)\,]$ identifies the word that induces a probability distribution over context words closest to that induced by $\mathcal{W}$, in which probability mass is assigned to $c_j$ wherever it is for $\mathcal{W}$. Intuitively, $w_*^{(1)}$ is the word that most closely reflects *all* aspects of $\mathcal{W}$, and may occur in contexts where no word $w_i \in \mathcal{W}$ does.

Minimising $D_{KL}[\,p(c_j|w_i)\,||\,p(c_j|\mathcal{W})\,]$ finds the word that induces a distribution over context words that is closest to that induced by $\mathcal{W}$, in which probability mass is assigned as broadly as possible but *only* to those $c_j$ to which probability mass is assigned for $\mathcal{W}$. Intuitively, $w_*^{(2)}$ is the word that reflects as many aspects of $\mathcal{W}$ as possible, as closely as possible, but nothing additional, e.g. by having other meaning that $\mathcal{W}$ does not.

### A.1. Weakening the paraphrase assumption

For a given word set $\mathcal{W}$, we consider the relationship between embedding sum $\mathbf{w}_\mathcal{W}$ and embedding $\mathbf{w}_*$ for the word $w_* \in \mathcal{E}$ that minimises the KL-divergence (we illustrate with $\Delta_{KL}^{\mathcal{W},w_*}$). Exploring a weaker assumption than D1, tests whether D1 might exceed requirement, and explores the relationship between $\mathbf{w}_*$ and $\mathbf{w}_\mathcal{W}$ as paraphrase error increases.

**Theorem 4** (Weak paraphrasing). *For $w_* \in \mathcal{E}, \mathcal{W} \subseteq \mathcal{E}$, if $w_*$ minimises $\Delta_{KL}^{\mathcal{W},w_*} \doteq D_{KL}[\,p(c_j|\mathcal{W})\,||\,p(c_j|w_*)\,]$, then:*

$$\mathbf{w}_*^\top \hat{\mathbf{c}} = \mathbf{w}_\mathcal{W}^\top \hat{\mathbf{c}} - \Delta_{KL}^{\mathcal{W},w_*} + \hat{\sigma}^\mathcal{W} - \tau^\mathcal{W} \quad (17)$$

*where $\hat{\mathbf{c}} = \mathbb{E}_{j|\mathcal{W}}[\mathbf{c}_j]$, $\hat{\sigma}^\mathcal{W} = \mathbb{E}_{j|\mathcal{W}}[\sigma_j^\mathcal{W}]$ and $\mathbb{E}_{j|\mathcal{W}}[\cdot]$ denotes expectation under $p(c_j|\mathcal{W})$.*

*Proof.*

$$\begin{aligned}\Delta_{KL}^{\mathcal{W},w_*} &= \sum_j p(c_j|\mathcal{W}) \log \tfrac{p(c_j|\mathcal{W})}{p(c_j|w_*)} \\ &\stackrel{(5)}{=} \mathbb{E}_{j|\mathcal{W}}[\,\sum_i \mathrm{PMI}(w_i, c_j) \\ &\qquad - \mathrm{PMI}(w_*, c_j) + \sigma_j^\mathcal{W} - \tau^\mathcal{W}\,] \\ &= \mathbb{E}_{j|\mathcal{W}}[\mathbf{w}_\mathcal{W}^\top \mathbf{c}_j - \mathbf{w}_*^\top \mathbf{c}_j] + \hat{\sigma}^\mathcal{W} - \tau^\mathcal{W} \quad \square\end{aligned}$$

Thus, the weaker paraphrase relationship specifies a hyperplane containing $\mathbf{w}_*$ and so does not uniquely define $\mathbf{w}_*$

(as under D1) and cannot explain the observation of embedding addition for paraphrases (as suggested by Gittens et al. (2017)). A similar result holds for $\Delta_{KL}^{w_*,\mathcal{W}}$. In principle, Thm 4 could help locate embeddings of words that more loosely paraphrase $\mathcal{W}$, i.e. with increased paraphrase error.

## B. Proof of Lemma 1

**Lemma 1.** *For any word $w_* \in \mathcal{E}$ and word set $\mathcal{W} \subseteq \mathcal{E}$, $|\mathcal{W}| < l$:*

$$\mathrm{PMI}_* = \sum_{w_i \in \mathcal{W}} \mathrm{PMI}_i + \boldsymbol{\rho}^{\mathcal{W},w_*} + \boldsymbol{\sigma}^\mathcal{W} - \tau^\mathcal{W} \mathbf{1}\,, \quad (5)$$

*where $\mathrm{PMI}_\bullet$ is the column of $\mathbf{PMI}$ corresponding to $w_\bullet \in \mathcal{E}$, $\mathbf{1} \in \mathbb{R}^n$ is a vector of 1s, and error terms $\sigma_j^\mathcal{W} = \log \tfrac{p(\mathcal{W}|c_j)}{\prod_i p(w_i|c_j)}$ and $\tau^\mathcal{W} = \log \tfrac{p(\mathcal{W})}{\prod_i p(w_i)}$.*

*Proof.*

$$\begin{aligned}&\mathrm{PMI}(w_*, c_j) - \sum_{w_i \in \mathcal{W}} \mathrm{PMI}(w_i, c_j) \\ &= \log \tfrac{p(w_*|c_j)}{p(w_*)} - \log \prod_{w_i \in \mathcal{W}} \tfrac{p(w_i|c_j)}{p(w_i)} \\ &= \log \tfrac{p(w_*|c_j)}{\prod_\mathcal{W} p(w_i|c_j)} - \log \tfrac{p(w_*)}{\prod_\mathcal{W} p(w_i)} \\ &\quad + \log \tfrac{p(\mathcal{W}|c_j)}{p(\mathcal{W}|c_j)} + \log \tfrac{p(\mathcal{W})}{p(\mathcal{W})} \\ &= \log \tfrac{p(w_*|c_j)}{p(\mathcal{W}|c_j)} - \log \tfrac{p(w_*)}{p(\mathcal{W})} \\ &\quad + \log \tfrac{p(\mathcal{W}|c_j)}{\prod_\mathcal{W} p(w_i|c_j)} - \log \tfrac{p(\mathcal{W})}{\prod_\mathcal{W} p(w_i)} \\ &= \log \tfrac{p(c_j|w_*)}{p(c_j|\mathcal{W})} + \log \tfrac{p(\mathcal{W}|c_j)}{\prod_\mathcal{W} p(w_i|c_j)} \\ &\qquad\qquad - \log \tfrac{p(\mathcal{W})}{\prod_\mathcal{W} p(w_i)} \\ &= \boldsymbol{\rho}_j^{\mathcal{W},w_*} + \boldsymbol{\sigma}_j^\mathcal{W} - \tau^\mathcal{W}\,,\end{aligned}$$

where, unless stated explicitly, products are with respect to all $w_i$ in the set indicated. $\square$

Introduced terms are highlighted to show their evolution within the proof. At the step where terms are introduced, the existing error terms have no statistical meaning. This is resolved by introducing terms to which both error terms can be meaningfully related, through paraphrasing and independence.

## C. Proof of Lemma 2

**Lemma 2.** *For any word sets $\mathcal{W}, \mathcal{W}_* \subseteq \mathcal{E}$, $|\mathcal{W}|, |\mathcal{W}_*| < l$:*

$$\sum_{w_i \in \mathcal{W}_*} \text{PMI}_i = \sum_{w_i \in \mathcal{W}} \text{PMI}_i + \boldsymbol{\rho}^{\mathcal{W},\mathcal{W}_*} + \boldsymbol{\sigma}^{\mathcal{W}} - \boldsymbol{\sigma}^{\mathcal{W}_*} - (\tau^{\mathcal{W}} - \tau^{\mathcal{W}_*})\mathbf{1}. \quad (10)$$

*Proof.*

$$\sum_{w_i \in \mathcal{W}_*} \text{PMI}(w_i, c_j) - \sum_{w_i \in \mathcal{W}} \text{PMI}(w_i, c_j)$$

$$= \log \prod_{w_i \in \mathcal{W}_*} \frac{p(w_i|c_j)}{p(w_i)} - \log \prod_{w_i \in \mathcal{W}} \frac{p(w_i|c_j)}{p(w_i)}$$

$$= \log \frac{\prod_{\mathcal{W}_*} p(w_i|c_j)}{\prod_{\mathcal{W}} p(w_i|c_j)} - \log \frac{\prod_{\mathcal{W}_*} p(w_i)}{\prod_{\mathcal{W}} p(w_i)}$$

$$+ \log \frac{p(\mathcal{W}_*|c_j)}{p(\mathcal{W}_*|c_j)} + \log \frac{p(\mathcal{W}_*)}{p(\mathcal{W}_*)}$$

$$+ \log \frac{p(\mathcal{W}|c_j)}{p(\mathcal{W}|c_j)} + \log \frac{p(\mathcal{W})}{p(\mathcal{W})}$$

$$= + \log \frac{p(\mathcal{W}_*|c_j)}{p(\mathcal{W}|c_j)} - \log \frac{p(\mathcal{W}_*)}{p(\mathcal{W})}$$

$$+ \log \frac{\prod_{\mathcal{W}_*} p(w_i|c_j)}{p(\mathcal{W}_*|c_j)} - \log \frac{\prod_{\mathcal{W}_*} p(w_i)}{p(\mathcal{W}_*)}$$

$$+ \log \frac{p(\mathcal{W}|c_j)}{\prod_{\mathcal{W}} p(w_i|c_j)} - \log \frac{p(\mathcal{W})}{\prod_{\mathcal{W}} p(w_i)}$$

$$= + \log \frac{p(c_j|\mathcal{W}_*)}{p(c_j|\mathcal{W})}$$

$$+ \log \frac{p(\mathcal{W}|c_j)}{\prod_{\mathcal{W}} p(w_i|c_j)} - \log \frac{p(\mathcal{W}_*|c_j)}{\prod_{\mathcal{W}_*} p(w_i|c_j)}$$

$$- \log \frac{p(\mathcal{W})}{\prod_{\mathcal{W}} p(w_i)} + \log \frac{p(\mathcal{W}_*)}{\prod_{\mathcal{W}_*} p(w_i)}$$

$$= \boldsymbol{\rho}_j^{\mathcal{W},\mathcal{W}_*} + \boldsymbol{\sigma}_j^{\mathcal{W}} - \boldsymbol{\sigma}_j^{\mathcal{W}_*} - (\tau^{\mathcal{W}} - \tau^{\mathcal{W}_*}),$$

where, unless stated explicitly, products are with respect to all $w_i$ in the set indicated. $\square$

The proof is analogous to that of Lem 1, with more terms added (as highlighted) to an equivalent effect. A key difference to single-word (or *direct*) paraphrases (D1) is that the paraphrase is between two word sets $\mathcal{W}$ and $\mathcal{W}_*$ that need not correspond to any single word. The paraphrase error $\boldsymbol{\rho}^{\mathcal{W},\mathcal{W}_*}$ compares the induced distributions of the two sets, following the same principles as direct paraphrasing, but with perhaps less interpretatability.

## D. Alternate Proof of Corollary 2.1

**Corollary 2.1.** *For any words $w_x, w_{x^*} \in \mathcal{E}$ and word sets $\mathcal{W}^+, \mathcal{W}^- \subseteq \mathcal{E}$, $|\mathcal{W}^+|, |\mathcal{W}^-| < l - 1$:*

$$\mathbf{w}_{x^*} = \mathbf{w}_x + \mathbf{w}_{\mathcal{W}^+} - \mathbf{w}_{\mathcal{W}^-} + \mathbf{C}^{\dagger}(\boldsymbol{\rho}^{\mathcal{W},\mathcal{W}_*} + \boldsymbol{\sigma}^{\mathcal{W}} - \boldsymbol{\sigma}^{\mathcal{W}_*} - (\tau^{\mathcal{W}} - \tau^{\mathcal{W}_*})\mathbf{1}), \quad (11)$$

*where $\mathcal{W} = \{w_x\} \cup \mathcal{W}^+$, $\mathcal{W}_* = \{w_{x^*}\} \cup \mathcal{W}^-$.*

*Proof.*

$$\text{PMI}(w_{x^*}, c_j) - \text{PMI}(w_x, c_j)$$

$$= \log \frac{p(c_j|w_{x^*})}{p(c_j|w_x)} + \log \prod_{w_i \in \mathcal{W}^+} \frac{p(c_j|w_i)}{p(c_j|w_i)}$$

$$+ \log \prod_{w_i \in \mathcal{W}^-} \frac{p(c_j|w_i)}{p(c_j|w_i)}$$

$$= \sum_{w_i \in \mathcal{W}^+} \log p(c_j|w_i) - \sum_{w_i \in \mathcal{W}^-} \log p(c_j|w_i)$$

$$+ \log \frac{\prod_{\mathcal{W}_*} p(c_j|w_i)}{\prod_{\mathcal{W}} p(c_j|w_i)}$$

$$= \sum_{w_i \in \mathcal{W}^+} \text{PMI}(w_i, c_j) - \sum_{w_i \in \mathcal{W}^-} \text{PMI}(w_i, c_j)$$

$$+ \log \frac{\prod_{\mathcal{W}_*} p(w_i|c_j) \prod_{\mathcal{W}} p(w_i)}{\prod_{\mathcal{W}} p(w_i|c_j) \prod_{\mathcal{W}_*} p(w_i)}$$

$$= \sum_{w_i \in \mathcal{W}^+} \text{PMI}(w_i, c_j) - \sum_{w_i \in \mathcal{W}^-} \text{PMI}(w_i, c_j)$$

$$+ \log \frac{p(c_j|w_{x^*}, W^-)}{p(c_j|w_x, W^+)}$$

$$+ \log \frac{\prod_{\mathcal{W}_*} p(w_i|c_j)}{p(w_{x^*}, W^-|c_j)} \frac{p(w_x, W^+|c_j)}{\prod_{\mathcal{W}} p(w_i|c_j)}$$

$$- \log \frac{\prod_{\mathcal{W}_*} p(w_i)}{p(w_{x^*}, W^-)} \frac{p(w_x, W^+)}{\prod_{\mathcal{W}} p(w_i)}$$

$$= \sum_{w_i \in \mathcal{W}^+} \text{PMI}(w_i, c_j) - \sum_{w_i \in \mathcal{W}^-} \text{PMI}(w_i, c_j)$$

$$+ \boldsymbol{\rho}_j^{\mathcal{W},\mathcal{W}_*} + \boldsymbol{\sigma}_j^{\mathcal{W}} - \boldsymbol{\sigma}_j^{\mathcal{W}_*} - (\tau^{\mathcal{W}} - \tau^{\mathcal{W}_*}),$$

where, unless stated explicitly, products are with respect to all $w_i$ in the set indicated; and $\mathcal{W} = \{w_x\} \cup \mathcal{W}^+$, $\mathcal{W}_* = \{w_{x^*}\} \cup \mathcal{W}^-$ to lighten notation. Multiplying by $\mathbf{C}^{\dagger}$ completes the proof. $\square$

## 3.3 Impact

*Analogies Explained* received an honourable mention for best paper at ICML 2019 and has received 59 citations as of August 2021, according to Google Scholar. Of these, the only works that follow up on the specific findings of the paper are included in later chapters.

Two contemporaneous works claim similar findings to the paper, which we discuss.

**Seonwoo et al. (2019)** extend the work of Gittens et al. (2017) to take account of negative sampling, but maintain the assumption that context words are independent given a target word (within their Equation 5) and that paraphrase words ($c$ of a word set $\mathcal{C}$) are identified by minimising a KL divergence. The first assumption is violated in practice and together those assumptions abstract away the statistical differences between distributions $p(w|c)$ and $p(w|\mathcal{C})$, over all words $w \in \mathcal{E}$, i.e. the interpretable error terms $\boldsymbol{\rho}, \boldsymbol{\sigma}, \boldsymbol{\tau}$ that explain the difference between the embedding of $c$ and the sum of embeddings of $\mathcal{C}$ (*AE*, Equation 6).

**Ethayarajh et al. (2019)** claim an alternative theoretical explanation of the analogy phenomenon for SGNS and GloVe embeddings assuming:

(i) the dimension of embeddings matches the rank of the factorised matrix, e.g. the shifted PMI matrix in Skip-Gram;

(ii) for each analogy, corresponding rows of the factorised matrix (e.g. our *PMI vectors*) lie in a plane;

(iii) analogous word pairs satisfy several *co-occurrence shifted PMI* (csPMI) relationships, where $\text{csPMI}(x,y) = \log \frac{p(x,y)^2}{p(x)p(y)} = \log p(x|y)p(y|x)$;

(iv) all words co-occur with themselves, i.e. occur twice in close succession, with probability proportional to their marginal probability, i.e. $p(w_i, c_i) \propto p(w_i)$, $\forall i$; and

(v) embedding matrices $\boldsymbol{W}, \boldsymbol{C}$ are equivalent up to a multiplicative scalar $\lambda$.

These assumptions raise significant issues, undermining the paper's claim:

- (i) is invalid for typical word embeddings, where the dimensionality is far below the rank of the factorised matrix;

- (ii) and (iii) are strong unintuitive assumptions that require justification to avoid simply restating one unexplained phenomenon in terms of several others:

    - why high-dimensional rows of the factorised matrix lie in a 2-D plane is a key aspect of explaining why parallelograms arise;

    - "csPMI" is a novel statistic with no evident semantic meaning;

- dividing by $p(w_i)$, (iv) implies $p(c_i|w_i)$ is a constant for all words of any given language, i.e. that, having occurred once, every word occurs again within the context window with identical probability – this is clearly false in general since context windows are arbitrarily defined;

- (v) implies $\mathbf{PMI} = \boldsymbol{W}^\top \boldsymbol{C} = \lambda \boldsymbol{W}^\top \boldsymbol{W}$ and so $\lambda^{-1}\mathbf{PMI}$ is *positive semi-definite* with all diagonal elements positive – this is empirically false in general, e.g. typically some diagonal elements of the PMI matrix are positive, some negative.

The potential impact of *Analogies Explained* may go beyond word embeddings since similar analogy relationships have been observed in document level embeddings (Dai et al., 2015) and in supervised (Reed et al., 2015) and unsupervised (Radford et al., 2016) representations from computer vision. Furthermore, the SGNS algorithm has been widely applied to other real-word domains, such as materials science (Tshitoyan et al., 2019), bio-medicine (Moen and Ananiadou, 2013; Chen et al., 2018; Chiu and Baker, 2020) and social/technological networks (Perozzi et al., 2014; Grover and Leskovec, 2016; Rozemberczki et al., 2021). Since nothing in the paper is domain specific, should analogous pairs exist within these data domains, e.g. as explicitly observed by Tshitoyan et al. (2019), the analogy phenomenon would be expected to occur and explained by the same theory based on co-occurrence statistics.

## 3.4 Discussion

The paper builds on aspects of prior works to explain why and *how well* (as reflected in the error terms) word embeddings of analogies approximate parallelograms. We emphasise that the central probabilistic relationships (Lemmas 1-3) are *identities* that hold for *any* words, but error terms vary for different word combinations. The better words satisfy statistical co-occurrence relationships that correspond to analogies, the smaller the error terms and the better their word embeddings approximate a parallelogram.

From this, it is unlikely that any word embeddings form a *precise* parallelogram since that requires strong independence conditions to hold, hence deviations from a parallelogram are a necessary statistical consequence rather than a "fault" of the embeddings. Thus, the presence of error terms is important in justifying why deviations from parallelograms are variable, which may, in part, explain the observed variable performance of the vector offset method (§2.1.5.3). Computing these error terms explicitly, however, is problematic since they are higher order (triple) co-occurrence terms and so particularly sparse and subject to sampling error.

We note known limitations of the paper that may merit further investigation to improve understanding of the analogy phenomenon or the latent semantic structure learned by word embedding algorithms.

- Several aspects or heuristics of word embedding algorithms remain unexplained, e.g. the sub-sampling of frequent words, the marginal noise distribution being raised to the power 0.75 in SGNS, the choice of context window size and that "average" embeddings $\frac{1}{2}(\boldsymbol{w}_i + \boldsymbol{c}_i)$ are found to improve performance (Pennington et al., 2014; Levy et al., 2015).

- Geometric relationships between PMI vectors are assumed to be *sufficiently preserved* when projected to the far lower dimensionality of word embeddings, but the projection of word embedding algorithms, which may be *probability-weighted* and/or *non-linear*, and the chosen number of dimensions are not analysed.

- No explanation is given for why other words of the analogy, $a$, $a^*$, $b$, must be excluded, which is problematic in practice where they may be valid answers to the analogy, e.g. "*man* is to *doctor* as *woman* is to ..." (Nissim et al., 2020).

- It is unclear how the embedding $\boldsymbol{w}_{b^*}$ of a candidate solution to an analogy should be compared to $\boldsymbol{w}_b + \boldsymbol{w}_{a^*} - \boldsymbol{w}_a$ to evaluate or mitigate (projected) error terms (*AE*, Equation 6) for which common heuristics, e.g. cosine similarity, remain unjustified. This should account for the observation that net dependence error terms tend to be negative, causing systematic error in paraphrases (Baroni and Lenci, 2010).

- Although the identified error terms explain inaccuracies in the parallelogram relationship of embeddings, they do not explain the observed performance variability of the vector offset method for analogies with different semantic relation types. The paper explains those analogies that satisfy the probabilistic definition based on word transformations, but not all analogies necessarily fit that semantic template.

# Chapter 4

# What the Vec? Towards Probabilistically Grounded Embeddings

The main contribution of this chapter is the paper *What the Vec? Towards Probabilistically Grounded Embeddings* (**What the Vec**), which was published at the *Annual Conference on Neural Information Processing Systems* in December 2019. We first outline the motivation for this work (§4.1), before including the paper itself (§4.2), followed by a summary of its impact so far (§4.3) and a discussion (§4.4).

## 4.1 Motivation

The motivation for *What the Vec* is to build on *Analogies Explained* to develop a fuller understanding of the latent semantic space of word embeddings that factorise a PMI matrix (or similar), as learned by algorithms such as SGNS and GloVe (**PMI-based embeddings**). Where *Analogies Explained* offers a mathematically rigorous but terse explanation of the analogy phenomenon, which treats paraphrases as an intermediate step, *What the Vec* takes a more holistic view of PMI-based embeddings and the semantic relations of *similarity*, *relatedness*, *paraphrase* and *analogy*. The paper puts previous findings into clearer perspective and fills gaps left by *Analogies Explained* by explicitly analysing the SGNS loss function, exploring the geometry of the space of PMI vectors and considering the relationship between the two embedding matrices $\boldsymbol{W}$ and $\boldsymbol{C}$.

## 4.2 The Paper

**Author Contributions**

The paper is co-authored by myself, Ivana Balažević and Timothy Hospedales. As the lead author, I developed the theoretical findings of the paper, designed the experiments, implemented some, and wrote the paper. Ivana implemented and ran some of the experiments, and proofread the paper. Tim provided useful discussions and suggestions during its development, and helped editing the final paper.

# What the Vec?
# Towards Probabilistically Grounded Embeddings

Carl Allen[1]     Ivana Balažević[1]     Timothy Hospedales[1,2]
[1] School of Informatics, University of Edinburgh, UK
[2] Samsung AI Centre, Cambridge, UK
`{carl.allen, ivana.balazevic, t.hospedales}@ed.ac.uk`

## Abstract

Word2Vec (W2V) and GloVe are popular, fast and efficient word embedding algorithms. Their embeddings are widely used and perform well on a variety of natural language processing tasks. Moreover, W2V has recently been adopted in the field of graph embedding, where it underpins several leading algorithms. However, despite their ubiquity and relatively simple model architecture, a theoretical understanding of *what* the embedding parameters of W2V and GloVe learn and *why* that is useful in downstream tasks has been lacking. We show that different interactions between *PMI vectors* reflect semantic word relationships, such as similarity and paraphrasing, that are encoded in low dimensional word embeddings under a suitable projection, theoretically explaining why embeddings of W2V and GloVe work. As a consequence, we also reveal an interesting mathematical interconnection between the considered semantic relationships themselves.

## 1  Introduction

Word2Vec[1] (W2V) [25] and GloVe [29] are fast, straightforward algorithms for generating *word embeddings*, or vector representations of words, often considered points in a *semantic space*. Their embeddings perform well on downstream tasks, such as identifying word similarity by vector comparison (e.g. cosine similarity) and solving analogies, such as the well known "*man* is to *king* as *woman* is to *queen*", by the addition and subtraction of respective embeddings [26, 27, 19].

In addition, the W2V algorithm has recently been adopted within the growing field of *graph embedding*, where the typical aim is to represent graph nodes in a common latent space such that their relative positioning can be used to predict edge relationships. Several state-of-the-art models for graph representation incorporate the W2V algorithm to learn node embeddings based on random walks over the graph [13, 30, 31]. Furthermore, word embeddings often underpin embeddings of word sequences, e.g. sentences. Although sentence embedding models can be complex [8, 17], as shown recently [38] they sometimes learn little beyond the information available in word embeddings.

Despite their relative ubiquity, much remains unknown of the W2V and GloVe algorithms, perhaps most fundamentally we lack a theoretical understanding of (i) *what is learned* in the embedding parameters; and (ii) *why that is useful* in downstream tasks. Answering such core questions is of interest in itself, particularly since the algorithms are unsupervised, but may also lead to improved embedding algorithms, or enable better use to be made of the embeddings we have. For example, both algorithms generate two embedding matrices, but little is known of how they relate or should interact. Typically one is simply discarded, whereas empirically their mean can perform well [29] and elsewhere they are assumed identical [14, 4]. As for embedding interactions, a variety of heuristics are in common use, e.g. cosine similarity [26] and *3CosMult* [19].

---

[1] We refer exclusively, throughout, to the more common implementation *Skipgram* with negative sampling.

Of works that seek to theoretically explain these embedding models [20, 14, 4, 9, 18], Levy and Goldberg [20] identify the loss function minimised (implicitly) by W2V and, thereby, the relationship between W2V word embeddings and the *Pointwise Mutual Information* (PMI) of word co-occurrences. More recently, Allen and Hospedales [2] showed that this relationship explains the linear interaction observed between embeddings of analogies. Building on these results, our key contributions are:

- to show how particular semantic relationships correspond to linear interactions of high dimensional *PMI vectors* and thus to equivalent interactions of low dimensional word embeddings generated by their *linear* projection, thereby explaining the semantic properties exhibited by embeddings of W2V and GloVe;

- to derive a relationship between embedding matrices proving that they must differ, justifying the heuristic use of their mean and enabling word embedding interactions – including the widely used cosine similarity – to be semantically interpreted; and

- to establish a novel hierarchical mathematical inter-relationship between relatedness, similarity, paraphrase and analogy (Fig 2).

## 2  Background

**Word2Vec** [25, 26] takes as input word pairs $\{(w_{i_r}, c_{j_r})\}_{r=1}^{D}$ extracted from a large text corpus, where target word $w_i \in \mathcal{E}$ ranges over the corpus and context word $c_j \in \mathcal{E}$ ranges over a window of size $l$, symmetric about $w_i$ ($\mathcal{E}$ is the dictionary of distinct words, $n = |\mathcal{E}|$). For each observed word pair, $k$ random pairs (*negative samples*) are generated from unigram distributions. For embedding dimension $d$, W2V's architecture comprises the product of two weight matrices $\mathbf{W}, \mathbf{C} \in \mathbb{R}^{d \times n}$ subject to the logistic sigmoid function. Columns of $\mathbf{W}$ and $\mathbf{C}$ are the *word embeddings*: $\mathbf{w}_i \in \mathbb{R}^d$, the $i^{th}$ column of $\mathbf{W}$, represents the $i^{th}$ word in $\mathcal{E}$ observed as the target word ($w_i$); and $\mathbf{c}_j \in \mathbb{R}^d$, the $j^{th}$ column of $\mathbf{C}$, represents the $j^{th}$ word in $\mathcal{E}$ observed as a context word ($c_j$).

Levy and Goldberg [20] show that the loss function of W2V is given by:

$$\ell_{W2V} = -\sum_{i=1}^{n}\sum_{j=1}^{n} \#(w_i, c_j) \log \sigma(\mathbf{w}_i^\top \mathbf{c}_j) + \tfrac{k}{D}\#(w_i)\#(c_j) \log(\sigma(-\mathbf{w}_i^\top \mathbf{c}_j)), \tag{1}$$

which is minimised if $\mathbf{w}_i^\top \mathbf{c}_j = \mathbf{P}_{i,j} - \log k$, where $\mathbf{P}_{i,j} = \log \frac{p(w_i, c_j)}{p(w_i)p(c_j)}$ is *pointwise mutual information* (PMI). In matrix form, this equates to factorising a *shifted* PMI matrix $\mathbf{S} \in \mathbb{R}^{n \times n}$:

$$\mathbf{W}^\top \mathbf{C} = \mathbf{S}. \tag{2}$$

**GloVe** [29] has the same architecture as W2V, but a different loss function, minimised when:

$$\mathbf{w}_i^\top \mathbf{c}_j = \log p(w_i, c_j) - b_i - b_j + \log Z, \tag{3}$$

for biases $b_i$, $b_j$ and normalising constant $Z$. In principle, the biases provide flexibility, broadening the family of statistical relationships that GloVe embeddings can learn.

**Analogies** are word relationships, such as the canonical "*man* is to *king* as *woman* is to *queen*", that are of particular interest because their word embeddings appear to satisfy a linear relationship [27, 19]. Allen and Hospedales [2] recently showed that this phenomenon follows from relationships between *PMI vectors*, i.e. rows of the (unshifted) PMI matrix $\mathbf{P} \in \mathbb{R}^{n \times n}$. In doing so, the authors define (i) the *induced distribution* of an observation $\circ$ as $p(\mathcal{E}|\circ)$, the probability distribution over all context words observed given $\circ$; and (ii) that a word $w_*$ *paraphrases* a set of words $\mathcal{W} \subset \mathcal{E}$ if the induced distributions $p(\mathcal{E}|w_*)$ and $p(\mathcal{E}|\mathcal{W})$ are (elementwise) similar.

## 3  Related Work

While many works explore empirical properties of word embeddings (e.g. [19, 23, 5]), we focus here on those that seek to theoretically explain why W2V and GloVe word embeddings capture semantic properties useful in downstream tasks. The first of these is the previously mentioned derivation by Levy and Goldberg [20] of the loss function (1) and the PMI relationship that minimises it (2). Hashimoto et al. [14] and Arora et al. [4] propose generative language models to explain the structure found in word embeddings. However, both contain strong *a priori* assumptions of an underlying geometry that we do not require (further, we find that several assumptions of [4] fail in practice (Appendix D)). Cotterell et al. [9] and Landgraf and Bellay [18] show that W2V performs *exponential (binomial) PCA* [7], however this follows from the (binomial) negative sampling and so describes the algorithm's mechanics, not *why* it works. Several works focus on the linearity of analogy embeddings [4, 12, 2, 10], but only [2] rigorously links semantics to embedding geometry (S.2).

To our knowledge, no previous work explains how the semantic properties of relatedness, similarity, paraphrase and analogy are all encoded in the relationships of PMI vectors and thereby manifest in the low dimensional word embeddings of W2V and GloVe.

## 4 PMI: linking geometry to semantics

The derivative of W2V's loss function (1) with respect to embedding $\mathbf{w}_i$, is given by:

$$\tfrac{1}{D}\nabla_{\mathbf{w}_i}\ell_{W2V} = \sum_{j=1}^n \big(\underbrace{p(w_i,c_j) + kp(w_i)p(c_j)}_{\mathbf{d}_j^{(i)}}\big)\big(\underbrace{\sigma(\mathbf{S}_{i,j}) - \sigma(\mathbf{w}_i^\top \mathbf{c}_j)}_{\mathbf{e}_j^{(i)}}\big)\mathbf{c}_j = \mathbf{C}\,\mathbf{D}^{(i)}\mathbf{e}^{(i)}\,, \quad (4)$$

for diagonal matrix $\mathbf{D}^{(i)} = diag(\mathbf{d}^{(i)}) \in \mathbb{R}^{n \times n}$; $\mathbf{d}^{(i)}, \mathbf{e}^{(i)} \in \mathbb{R}^n$ containing the probability and error terms indicated; and all probabilities estimated empirically from the corpus. This confirms that (1) is minimised if $\mathbf{W}^\top \mathbf{C} = \mathbf{S}$ (2), since all $\mathbf{e}_j^{(i)} = 0$, but that requires $\mathbf{W}$ and $\mathbf{C}$ to each have rank at least that of $\mathbf{S}$. In the general case, including the typical case $d \ll n$, (1) is minimised when probability weighted error vectors $\mathbf{D}^{(i)}\mathbf{e}^{(i)}$ are orthogonal to the rows of $\mathbf{C}$. As such, embeddings $\mathbf{w}_i$ can be seen as a non-linear (due to the sigmoid function $\sigma(\cdot)$) *projection* of rows of $\mathbf{S}$, induced by the loss function. (Note that the distinction between $\mathbf{W}$ and $\mathbf{C}$ is arbitrary: embeddings $\mathbf{c}_j$ can also be viewed as projections onto the rows of $\mathbf{W}$.)

Recognising that the $\log k$ shift term is an artefact of the W2V algorithm (see Appendix A), whose effect can be evaluated subsequently (as in [2]), we exclude it and analyse properties and interactions of word embeddings $\mathbf{w}_i$ that are projections of $\mathbf{p}^i$, the corresponding rows of $\mathbf{P}$ (*PMI vectors*). We aim to identify the properties of PMI vectors that capture semantics and are then preserved in word embeddings under the low-rank projection induced by a suitably chosen loss function.

### 4.1 The domain of PMI vectors

PMI vector $\mathbf{p}^i \in \mathbb{R}^n$ has a component PMI$(w_i, c_j)$ for all context words $c_j \in \mathcal{E}$, given by:

$$\text{PMI}(w_i, c_j) = \log \tfrac{p(c_j, w_i)}{p(w_i)p(c_j)} = \log \tfrac{p(c_j|w_i)}{p(c_j)}. \quad (5)$$

Any difference in the probability of observing $c_j$ having observed $w_i$, relative to its marginal probability, can be thought of as *due to* $w_i$. Thus PMI$(w_i, c_j)$ captures the influence of one word on another. Specifically, by reference to marginal probability $p(c_j)$: PMI$(w_i, c_j) > 0$ implies $c_j$ is more likely to occur in the presence of $w_i$; PMI$(w_i, c_j) < 0$ implies $c_j$ is less likely to occur given $w_i$; and PMI$(w_i, c_j) = 0$ indicates that $w_i$ and $c_j$ occur independently, i.e. they are unrelated. PMI thus reflects the semantic property of *relatedness*, as previously noted [36, 6, 15]. A PMI *vector* thus reflects any change in the probability distribution over all words $p(\mathcal{E})$, given (or due to) $w_i$:

$$\mathbf{p}^i \triangleq \big\{\log \tfrac{p(c_j|w_i)}{p(c_j)}\big\}_{c_j \in \mathcal{E}} \triangleq \log \tfrac{p(\mathcal{E}|w_i)}{p(\mathcal{E})}. \quad (6)$$

While PMI values are unconstrained in $\mathbb{R}$, PMI vectors are constrained to an $n-1$ dimensional surface $\mathcal{S} \subset \mathbb{R}^n$, where each dimension corresponds to a word (Fig 1) (although technically a *hypersurface*, we refer to $\mathcal{S}$ simply as a "surface"). The geometry of $\mathcal{S}$ can be constructed step-wise from (6):

- the vector of numerator terms $\mathbf{q}^i = p(\mathcal{E}|w_i)$ lies on the simplex $\mathcal{Q} \subset \mathbb{R}^n$;
- dividing all $\mathbf{q} \in \mathcal{Q}$ (element-wise) by $\mathbf{p} = p(\mathcal{E}) \in \mathcal{Q}$, gives probability ratio vectors $\tfrac{\mathbf{q}}{\mathbf{p}}$ that lie on a "stretched simplex" $\mathcal{R} \subset \mathbb{R}^n$ (containing $\mathbf{1} \in \mathbb{R}^n$) that has a vertex at $\tfrac{1}{p(c_j)}$ on axis $j$, $\forall c_j \in \mathcal{E}$; and
- the natural logarithm transforms $\mathcal{R}$ to the surface $\mathcal{S}$, with $\mathbf{p}^i = \log \tfrac{p(\mathcal{E}|w_i)}{p(\mathcal{E})} \in \mathcal{S}$, $\forall w_i \in \mathcal{E}$.

Note, $\mathbf{p} = p(\mathcal{E})$ uniquely determines $\mathcal{S}$. Considering each point $\mathbf{s} \in \mathcal{S}$ as an element-wise log probability ratio vector $\mathbf{s} = \log \frac{\mathbf{q}}{\mathbf{p}} \in \mathcal{S}$ ($\mathbf{q} \in \mathcal{Q}$), shows $\mathcal{S}$ to have the properties (proofs in Appendix B):

P1 **$\mathcal{S}$, and any subsurface of $\mathcal{S}$, is non-linear.** PMI vectors are thus not constrained to a linear subspace, identifiable by low-rank factorisation of the PMI matrix, as may seem suggested by (2).

P2 **$\mathcal{S}$ contains the origin,** which can be considered the PMI vector of the *null word* $\emptyset$, i.e. $\mathbf{p}^\emptyset = \log \frac{p(\mathcal{E}|\emptyset)}{p(\mathcal{E})} = \log \frac{p(\mathcal{E})}{p(\mathcal{E})} = \mathbf{0} \in \mathbb{R}^n$.

P3 **Probability vector $\mathbf{q} \in \mathcal{Q}$ is normal to the tangent plane of $\mathcal{S}$** at $\mathbf{s} = \log \frac{\mathbf{q}}{\mathbf{p}} \in \mathcal{S}$.

P4 **$\mathcal{S}$ does not intersect with the fully positive or fully negative orthants** (excluding $\mathbf{0}$). Thus PMI vectors are not *isotropically* (i.e. uniformly) distributed in space (as assumed in [4]).

P5 **The sum of 2 points $\mathbf{s} + \mathbf{s}'$ lies in $\mathcal{S}$ only for certain $\mathbf{s}, \mathbf{s}' \in \mathcal{S}$.** That is, for any $\mathbf{s} \in \mathcal{S}$ ($\mathbf{s} \neq \mathbf{0}$), there exists a (strict) subset $\mathcal{S}_s \subset \mathcal{S}$, such that $\mathbf{s} + \mathbf{s}' \in \mathcal{S}$ iff $\mathbf{s}' \in \mathcal{S}_s$. Trivially $\mathbf{0} \in \mathcal{S}_s$, $\forall \mathbf{s} \in \mathcal{S}$.

Note that while all PMI vectors lie in $\mathcal{S}$, certainly not all (infinite) points in $\mathcal{S}$ correspond to the (finite) PMI vectors of words. Interestingly, P2 and P5 allude to properties of a *vector space*, often the desired structure for a *semantic space* [14]. Whilst the domain of PMI vectors is clearly not a vector space, addition and subtraction of PMI vectors do have *semantic meaning*, as we now show.

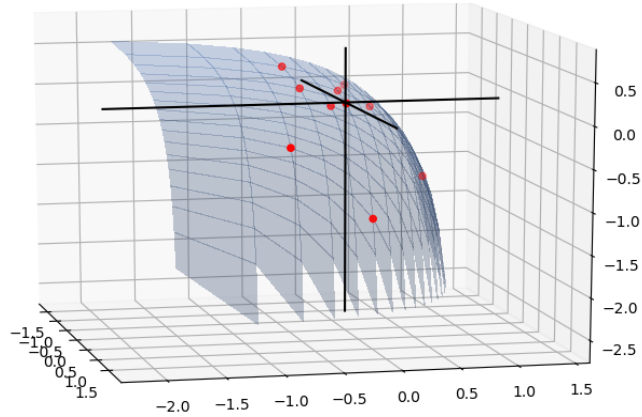

Figure 1: The PMI surface $\mathcal{S}$, showing sample PMI vectors of words (red dots)

## 4.2 Subtraction of PMI vectors finds similarity

Taking the definition from [2] (see S.2), we consider a word $w_i$ that *paraphrases* a word set $\mathcal{W} \in \mathcal{E}$, where $\mathcal{W} = \{w_j\}$ contains a single word. Since paraphrasing requires distributions of local context words (induced distributions) to be similar, this intuitively finds $w_i$ that are interchangeable with, or *similar* to, $w_j$: in the limit $w_j$ itself or, less trivially, a synonym. Thus, word similarity corresponds to a low KL divergence between $p(\mathcal{E}|w_i)$ and $p(\mathcal{E}|w_j)$. Interestingly, the difference between the associated PMI vectors:

$$\boldsymbol{\rho}^{i,j} = \mathbf{p}^i - \mathbf{p}^j = \log \frac{p(\mathcal{E}|w_i)}{p(\mathcal{E}|w_j)}, \qquad (7)$$

is a vector of un-weighted KL divergence components. Thus, if dimensions were suitably weighted, the sum of difference components (comparable to Manhattan distance but *directed*) would equate to a KL divergence between induced distributions. That is, if $\mathbf{q}^i = p(\mathcal{E}|w_i)$, then a KL divergence is given by $\mathbf{q}^{i\top} \boldsymbol{\rho}^{i,j}$. Furthermore, $\mathbf{q}^i$ is the normal to the surface $\mathcal{S}$ at $\mathbf{p}^i$ (with unit $l_1$ norm), by P3. The projection onto the normal (to $\mathcal{S}$) at $\mathbf{p}^j$, i.e. $-\mathbf{q}^{j\top} \boldsymbol{\rho}^{i,j}$, gives the other KL divergence. (Intuition for the semantic interpretation of each KL divergence is discussed in Appendix A of [2].)

### 4.3 Addition of PMI vectors finds paraphrases

From geometric arguments (P5), we know that only certain pairs of points in $\mathcal{S}$ sum to another point in the surface. We can also consider the *probabilistic* conditions for PMI vectors to sum to another:

$$\begin{aligned}
\mathbf{x} = \mathbf{p}^i + \mathbf{p}^j &= \log \tfrac{p(\mathcal{E}|w_i)}{p(\mathcal{E})} + \log \tfrac{p(\mathcal{E}|w_j)}{p(\mathcal{E})} \\
&= \underbrace{\log \tfrac{p(\mathcal{E}|w_i,w_j)}{p(\mathcal{E})}}_{\mathbf{p}^{i,j}} - \underbrace{\log \tfrac{p(w_i,w_j|\mathcal{E})}{p(w_i|\mathcal{E})p(w_j|\mathcal{E})}}_{\boldsymbol{\sigma}^{ij}} + \underbrace{\log \tfrac{p(w_i,w_j)}{p(w_i)p(w_j)}}_{\tau^{ij}} = \mathbf{p}^{i,j} - \boldsymbol{\sigma}^{ij} + \tau^{ij}\mathbf{1},
\end{aligned} \qquad (8)$$

where (overloading notation) $\mathbf{p}^{i,j}\!\in\!\mathcal{S}$ is a vector of PMI terms involving $p(\mathcal{E}|w_i,w_j)$, the induced distribution of $w_i$ and $w_j$ observed *together*;[2] and $\boldsymbol{\sigma}^{ij}\!\in\!\mathbb{R}^n$, $\tau^{ij}\!\in\!\mathbb{R}$ are the conditional and marginal dependence terms indicated (as seen in [2]). From (8), if $w_i, w_j \in \mathcal{E}$ occur both *independently and conditionally independently* given each and every word in $\mathcal{E}$, then $\mathbf{x}\!=\!\mathbf{p}^{i,j}\!\in\!\mathcal{S}$, and (from P5) $\mathbf{p}^j\!\in\!\mathcal{S}_{\mathbf{p}^i}$ and $\mathbf{p}^i\!\in\!\mathcal{S}_{\mathbf{p}^j}$. If not, error vector $\boldsymbol{\varepsilon}^{ij}\!=\!\boldsymbol{\sigma}^{ij}\!-\!\tau^{ij}\mathbf{1}$ separates $\mathbf{x}$ and $\mathbf{p}^{i,j}$ and $\mathbf{x}\!\notin\!\mathcal{S}$, unless by meaningless coincidence. (Note, whilst probabilistic aspects here mirror those of [2], we combine these with a geometric understanding.) Although certainly $\mathbf{p}^{i,j}\!\in\!\mathcal{S}$, the extent to which $\mathbf{p}^{i,j}\!\approx\!\mathbf{p}^k$ for some $w_k\!\in\!\mathcal{E}$ depends on paraphrase error $\boldsymbol{\rho}^{k,\{i,j\}}\!=\!\mathbf{p}^k\!-\!\mathbf{p}^{i,j}$, that compares the induced distributions of $w_k$ and $\{w_i, w_j\}$. Thus the PMI vector difference $(\mathbf{p}^i\!+\!\mathbf{p}^j)\!-\!\mathbf{p}^k$ for any words $w_i, w_j, w_k\!\in\!\mathcal{E}$ comprises: $\boldsymbol{\varepsilon}^{ij}$ a component between $\mathbf{p}^i\!+\!\mathbf{p}^j$ and the surface $\mathcal{S}$ (reflecting word dependence); and $\boldsymbol{\rho}^{k,\{i,j\}}$ a component *along* the surface (reflecting paraphrase error). The latter captures a semantic relationship with $w_k$, which the former may obscure, irrespective of $w_k$. (Further geometric and probabilistic implications are considered in Appendix C.)

### 4.4 Linear combinations of PMI vectors find analogies

PMI vectors of analogy relationships "$w_a$ is to $w_{a^*}$ as $w_b$ is to $w_{b^*}$" have been proven [2] to satisfy:

$$\mathbf{p}^{b^*} \approx \mathbf{p}^{a^*} - \mathbf{p}^a + \mathbf{p}^b. \qquad (9)$$

The proof builds on the concept of paraphrasing (with error terms similar to those in Section 4.3), comparing PMI vectors of analogous word pairs to show that $\mathbf{p}^a + \mathbf{p}^{b^*} \approx \mathbf{p}^{a^*} + \mathbf{p}^b$ and thus (9).

## 5  Encoding PMI: from PMI vectors to word embeddings

Understanding how high dimensional PMI vectors encode semantic properties desirable in word embeddings, we consider how they can be transferred to low dimensional representations. A key observation is that all PMI vector interactions, for similarity (7), paraphrases (8) and analogies (9), are *additive*, and are therefore preserved under *linear* projection. By comparison, the loss function of W2V (1) projects PMI vectors non-linearly, and that of GloVe (3) does project linearly, but not (necessarily) PMI vectors. Linear projection can be achieved by the least squares loss function:[3]

$$\ell_{LSQ} \;=\; \tfrac{1}{2} \sum_{i=1}^n \sum_{j=1}^n \bigl(\mathbf{w}_i^\top \mathbf{c}_j - \mathrm{PMI}(w_i, c_j)\bigr)^2. \qquad (10)$$

$\ell_{LSQ}$ is minimised when $\nabla_{\mathbf{W}^\top}\ell_{LSQ} = (\mathbf{W}^\top\mathbf{C} - \mathbf{P})\mathbf{C}^\top = 0$, or $\mathbf{W}^\top = \mathbf{P}\,\mathbf{C}^\dagger$, for $\mathbf{C}^\dagger = \mathbf{C}^\top(\mathbf{C}\mathbf{C}^\top)^{-1}$ the *Moore–Penrose pseudoinverse* of $\mathbf{C}$. This explicit linear projection allows interactions performed between word embeddings, e.g. dot product, to be mapped to interactions between PMI vectors, and thereby semantically interpreted. However, we do better still by considering how $\mathbf{W}$ and $\mathbf{C}$ relate.

### 5.1  The relationship between W and C

Whilst W2V and GloVe train two embedding matrices, typically only $\mathbf{W}$ is used and $\mathbf{C}$ discarded. Thus, although relationships are learned between $\mathbf{W}$ and $\mathbf{C}$, they are tested between $\mathbf{W}$ and $\mathbf{W}$. If $\mathbf{W}$ and $\mathbf{C}$ are equal, the distinction falls away, but that is not found to be the case in practice. Here, we consider why typically $\mathbf{W} \neq \mathbf{C}$ and, as such, what relationship between $\mathbf{W}$ and $\mathbf{C}$ does exist.

---

[2]Whilst $w_i, w_j$ are *both* target words, by symmetry we can interchange roles of context and target words to compute $p(\mathcal{E}|w,w')$ based on the distribution of target words for which $w_i$ and $w_j$ are both context words.

[3]We note that the W2V and GloVe loss functions include probability weightings (as considered in [35]), which we omit for simplicity.

If the symmetric PMI matrix $\mathbf{P}$ is positive semi-definite (PSD), its closest low-rank approximation (minimising $\ell_{LSQ}$) is given by the eigendecomposition $\mathbf{P} = \mathbf{\Pi}\mathbf{\Lambda}\mathbf{\Pi}^\top$, $\mathbf{\Pi}, \mathbf{\Lambda} \in \mathbb{R}^{n \times n}$, $\mathbf{\Pi}^\top\mathbf{\Pi} = \mathbf{I}$; and $\ell_{LSQ}$ is minimised by $\mathbf{W} = \mathbf{C} = \mathbf{S}^{1/2}\mathbf{U}^\top$, where $\mathbf{S} \in \mathbb{R}^{d \times d}$, $\mathbf{U} \in \mathbb{R}^{d \times n}$ are $\mathbf{\Lambda}, \mathbf{\Pi}$, respectively, truncated to their $d$ largest eigenvalue components. Any matrix pair $\mathbf{W}^* = \mathbf{M}^\top\mathbf{W}$, $\mathbf{C}^* = \mathbf{M}^{-1}\mathbf{W}$, also minimises $\ell_{LSQ}$ (for any invertible $\mathbf{M} \in \mathbb{R}^{d \times d}$), but of these $\mathbf{W}, \mathbf{C}$ are unique (up to rotation and permutation) in satisfying $\mathbf{W} = \mathbf{C}$, a preferred solution for learning word embeddings since the number of free parameters is halved and consideration of whether to use $\mathbf{W}, \mathbf{C}$ or both falls away.

However, $\mathbf{P}$ is not typically PSD in practice and this preferred (real) factorisation does not exist since $\mathbf{P}$ has negative eigenvalues, $\mathbf{S}^{1/2}$ is complex and any $\mathbf{W}, \mathbf{C}$ minimising $\ell_{LSQ}$ with $\mathbf{W} = \mathbf{C}$ must also be complex. (Complex word embeddings arise elsewhere, e.g. [16, 22], but since the word embeddings we examine are real we keep to the real domain.) By implication, any $\mathbf{W}, \mathbf{C} \in \mathbb{R}^{d \times n}$ that minimise $\ell_{LSQ}$ *cannot be equal*, contradicting the assumption $\mathbf{W} = \mathbf{C}$ sometimes made [14, 4]. Returning to the eigendecomposition, if $\mathbf{S}$ contains the $d$ largest *absolute* eigenvalues and $\mathbf{U}$ the corresponding eigenvectors of $\mathbf{P}$, we define $\mathbf{I}' = sign(\mathbf{S})$ (i.e. $\mathbf{I}'_{ii} = \pm 1$) such that $\mathbf{S} = |\mathbf{S}|\mathbf{I}'$. Thus, $\mathbf{W} = |\mathbf{S}|^{1/2}\mathbf{U}^\top$ and $\mathbf{C} = \mathbf{I}'\mathbf{W}$ can be seen to minimise $\ell_{LSQ}$ (i.e. $\mathbf{W}^\top\mathbf{C} \approx \mathbf{P}$) with $\mathbf{W} \neq \mathbf{C}$ but where corresponding *rows* of $\mathbf{W}, \mathbf{C}$ (denoted by superscript) satisfy $\mathbf{W}^i = \pm\mathbf{C}^i$ (recall word embeddings $\mathbf{w}_i, \mathbf{c}_i$ are columns of $\mathbf{W}, \mathbf{C}$). Such $\mathbf{W}, \mathbf{C}$ can be seen as *quasi*-complex conjugate. Again, $\mathbf{W}, \mathbf{C}$ can be used to define a family of matrix pairs that minimise $\ell_{LSQ}$, of which $\mathbf{W}, \mathbf{C}$ themselves are a most parameter efficient choice, with $(n+1)d$ free parameters compared to $2nd$.

## 5.2 Interpreting embedding interactions

Various word embedding interactions are used to predict semantic relationships, e.g. cosine similarity [26] and 3CosMult [19], although typically with little theoretical justification. With a semantic understanding of PMI vector interactions (S.4) and the derived relationship $\mathbf{C} = \mathbf{I}'\mathbf{W}$, we now interpret commonly used word embedding interactions and evaluate the effect of combining embeddings of $\mathbf{W}$ only (e.g. $\mathbf{w}_i^\top\mathbf{w}_j$), rather than $\mathbf{W}$ and $\mathbf{C}$ (e.g. $\mathbf{w}_i^\top\mathbf{c}_j$). For use below, we note that $\mathbf{W}^\top\mathbf{C} = \mathbf{U}\mathbf{S}\mathbf{U}^\top$, $\mathbf{C}^\dagger = \mathbf{U}|\mathbf{S}|^{-1/2}\mathbf{I}'$ and define: *reconstruction error matrix* $\mathbf{E} = \mathbf{P} - \mathbf{W}^\top\mathbf{C}$, i.e. $\mathbf{E} = \overline{\mathbf{U}}\,\overline{\mathbf{S}}\,\overline{\mathbf{U}}^\top$ where $\overline{\mathbf{U}}, \overline{\mathbf{S}}$ contain the $n-d$ smallest absolute eigenvalue components of $\mathbf{\Pi}, \mathbf{\Sigma}$ (as omitted from $\mathbf{U}, \mathbf{S}$); $\mathbf{F} = \mathbf{U}(\frac{\mathbf{S}-|\mathbf{S}|}{2})\mathbf{U}^\top$, comprising the negative eigenvalue components of $\mathbf{P}$; and *mean embeddings* $\mathbf{a}_i$ as the columns of $\mathbf{A} = \frac{\mathbf{W}+\mathbf{C}}{2} = \mathbf{U}|\mathbf{S}|^{1/2}\mathbf{I}'' \in \mathbb{R}^{d \times n}$, where $\mathbf{I}'' = \frac{\mathbf{I}+\mathbf{I}'}{2}$ (i.e. $\mathbf{I}''_{ii} \in \{0,1\}$).

**Dot Product:** We compare the following interactions, associated with predicting relatedness:

$$\begin{array}{llll}
\mathbf{W}, \mathbf{C} : & \mathbf{w}_i^\top\mathbf{c}_j = \mathbf{U}^i\,\mathbf{S}\,\mathbf{U}^{j\top} & & = \mathbf{P}_{i,j} - \mathbf{E}_{i,j} \\
\mathbf{W}, \mathbf{W} : & \mathbf{w}_i^\top\mathbf{w}_j = \mathbf{U}^i\,|\mathbf{S}|\,\mathbf{U}^{j\top} & = \mathbf{U}^i(\mathbf{S} - (\mathbf{S}-|\mathbf{S}|))\mathbf{U}^{j\top} & = \mathbf{P}_{i,j} - \mathbf{E}_{i,j} - 2\,\mathbf{F}_{i,j} \\
\mathbf{A}, \mathbf{A} : & \mathbf{a}_i^\top\mathbf{a}_j = \mathbf{U}^i\,|\mathbf{S}|\,\mathbf{I}''\,\mathbf{U}^{j\top} & = \mathbf{U}^i(\mathbf{S} - (\frac{\mathbf{S}-|\mathbf{S}|}{2}))\mathbf{U}^{j\top} & = \mathbf{P}_{i,j} - \mathbf{E}_{i,j} - \mathbf{F}_{i,j}
\end{array}$$

This shows that $\mathbf{w}_i^\top\mathbf{w}_j$ *overestimates* the PMI approximation given by $\mathbf{w}_i^\top\mathbf{c}_j$ by twice any component relating to negative eigenvalues – an overestimation that is halved using mean embeddings, $\mathbf{a}_i^\top\mathbf{a}_j$.

**Difference sum:** $(\mathbf{w}_i - \mathbf{w}_j)^\top\mathbf{1} = (\mathbf{p}^i - \mathbf{p}^j)\mathbf{C}^\dagger\mathbf{1} = \sum_{k=1}^n \mathbf{x}_k \log \frac{p(c_k|w_i)}{p(c_k|w_j)}$, $\mathbf{x} = \mathbf{U}|\mathbf{S}|^{-1/2}\mathbf{I}'\mathbf{1}$

Thus, summing over the difference of embedding components compares to a KL divergence between induced distributions (and so *similarity*) more so than for PMI vectors (S.4.2) as dimensions are weighted by $\mathbf{x}_k$. However, unlike a KL divergence, $\mathbf{x}$ is not a probability distribution and does not vary with $\mathbf{w}_i$ or $\mathbf{w}_j$. We speculate that between $\mathbf{x}$ and the omitted probability weighting of the loss function, the dimensions of low probability words are down-weighted, mitigating the effect of "outliers" to which PMI is known to be sensitive [37], and loosely reflecting a KL divergence.

**Euclidean distance:** $\|\mathbf{w}_i - \mathbf{w}_j\|_2 = \|(\log \frac{p(\mathcal{E}|w_i)}{p(\mathcal{E}|w_j)})\mathbf{C}^\dagger\|_2$ shows no obvious meaning.

**Cosine similarity:** Surprisingly, $\frac{\mathbf{w}_i^\top\mathbf{w}_j}{\|\mathbf{w}_i\|\|\mathbf{w}_j\|}$, as often used effectively to predict word relatedness and/or similarity [33, 5], has no immediate semantic interpretation. However, recent work [3] proposes a more holistic description of relatedness than $\text{PMI}(w_i, w_j) > 0$ (S.4.1): that related words $(w_i, w_j)$ have multiple positive PMI vector components in common, because all words associated with any common semantic "theme" are also more likely to co-occur. The *strength* of relatedness (*similarity* being the extreme case) is given by the number of common word associations, as reflected in the dimensionality of a common aggregate PMI vector component, which projects to a common embedding component. The *magnitude* of such common component is not directly meaningful, but as relatedness increases and $w_i, w_j$ share more common word associations, the *angle* between their PMI vectors, and so too their embeddings, narrows, justifying the widespread use of cosine similarity.

Table 1: Accuracy in semantic tasks using different loss functions on the text8 corpus [24].

| Model | Loss | Relationship | Relatedness [1] | Similarity [1] | Analogy [25] |
|---|---|---|---|---|---|
| $W2V$ | W2V | $\mathbf{W}^\top\mathbf{C} \approx \mathbf{P}$ | .628 | .703 | .283 |
| $W{=}C$ | LSQ | $\mathbf{W}^\top\mathbf{W} \approx \mathbf{P}$ | .721 | .786 | .411 |
| $LSQ$ | LSQ | $\mathbf{W}^\top\mathbf{C} \approx \mathbf{P}$ | **.727** | **.791** | **.425** |

Other statistical word embedding relationships assumed in [4] are considered in Appendix D.

## 6   Empirical evidence

Word embeddings (especially those of W2V) have been well empirically studied, with many experimental findings. Here we draw on previous results and run test experiments to provide empirical support for our main theoretical results:

1. Analogies form as linear relationships between linear projections of PMI vectors (S.4.4)

   Whilst previously explained in [2], we emphasise that their rationale for this well known phenomenon fits precisely within our broader explanation of W2V and GloVe embeddings. Further, re-ordering paraphrase questions is observed to materially affect prediction accuracy [23], which can be justified from the explanation provided in [2] (see Appendix E).

2. The linear projection of additive PMI vectors captures semantic properties more accurately than the non-linear projection of W2V (S.5).

   Several works consider alternatives to the W2V loss function [20, 21], but none isolates the effect of an equivalent linear loss function, which we therefore implement (detail below). Comparing models $W2V$ and $LSQ$ (Table 1) shows a material improvement across all semantic tasks from linear projection.

3. Word embedding matrices $\mathbf{W}$ and $\mathbf{C}$ are dissimilar (S.5.1).

   $\mathbf{W}, \mathbf{C}$ are typically found to differ, e.g. [26, 29, 28]. To demonstrate the difference, we include an experiment tying $\mathbf{W}{=}\mathbf{C}$. Comparing models $W{=}C$ and $LSQ$ (Table 1) shows a small but consistent improvement in the former despite a lower data-to-parameter ratio.

4. Dot products recover PMI with decreasing accuracy: $\mathbf{w}_i^\top \mathbf{c}_j \geq \mathbf{a}_i^\top \mathbf{a}_j \geq \mathbf{w}_i^\top \mathbf{w}_j$ (S.5.2).

   The use of average embeddings $\mathbf{a}_i^\top \mathbf{a}_j$ over $\mathbf{w}_i^\top \mathbf{w}_j$ is a well-known heuristic [29, 21]. More recently, [5] show that relatedness correlates noticeably better to $\mathbf{w}_i^\top \mathbf{c}_j$ than either of the "symmetric" choices ($\mathbf{a}_i^\top \mathbf{a}_j$ or $\mathbf{w}_i^\top \mathbf{w}_j$).

5. Relatedness is reflected by interactions between $\mathbf{W}$ and $\mathbf{C}$ embeddings, and similarity is reflected by interactions between $\mathbf{W}$ and $\mathbf{W}$. (S.5.2)

   Asr et al. [5] compare human judgements of similarity and relatedness to cosine similarity between combinations of $\mathbf{W}, \mathbf{C}$ and $\mathbf{A}$. The authors find a "very consistent" support for their conclusion that "WC ... best measures ... relatedness" and "similarity [is] best predicted by ... WW". An example is given for *house*: $\mathbf{w}_i^\top \mathbf{w}_j$ gives *mansion*, *farmhouse* and *cottage*, i.e. similar or synonymous words; $\mathbf{w}_i^\top \mathbf{c}_j$ gives *barn*, *residence*, *estate*, *kitchen*, i.e. related words.

**Models:**   As we perform a standard comparison of loss functions, similar to [20, 21], we leave experimental details to Appendix F. In summary, we learn 500 dimensional embeddings from word co-occurrences extracted from a standard corpus ("text8" [24]). We implement loss function (1) explicitly as model $W2V$. Models $W{=}C$ and $LSQ$ use least squares loss (10), with constraint $\mathbf{W}{=}\mathbf{C}$ in the latter (see point 3 above). Evaluation on popular data sets [1, 25] uses the Gensim toolkit [32].

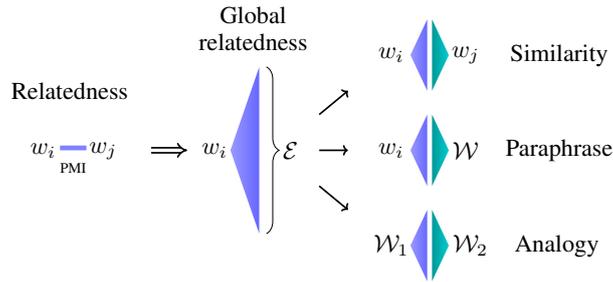

Figure 2: Interconnection between semantic relationships: relatedness is a base pairwise comparison (measured by PMI); *global relatedness* considers relatedness to all words (PMI vector); similarity, paraphrase and analogy depend on global relatedness between words ($w\in\mathcal{E}$) and word sets ($\mathcal{W}\subseteq\mathcal{E}$).

## 7 Discussion

Having established mathematical formulations for relatedness, similarity, paraphrase and analogy that explain how they are captured in word embeddings derived from PMI vectors (S.4), it can be seen that they also imply an interesting, hierarchical interplay between the semantic relationships themselves (Fig 2). At the core is *relatedness*, which correlates with PMI, both empirically [36, 6, 15] and intuitively (S.4.2). As a pairwise comparison of words, relatedness acts somewhat akin to a *kernel* (an actual kernel requires **P** to be PSD), allowing words to be considered numerically in terms of their relatedness to all words, as captured in a PMI vector, and compared according to how they each relate to all other words, or *globally relate*. Given this meta-comparison, we see that one word is *similar* to another if they are globally related (1-1); a *paraphrase* requires one word to globally relate to the joint occurrence of a set of words (1-$n$); and analogies arise when joint occurrences of word pairs are globally related ($n$-$n$). Continuing the "kernel" analogy, the PMI matrix mirrors a kernel matrix, and word embeddings the representations derived from *kernelised PCA* [34].

## 8 Conclusion

In this work, we take two previous results – the well known link between W2V embeddings and PMI [20], and a recent connection between PMI and analogies [2] – to show how the semantic properties of relatedness, similarity, paraphrase and analogy are captured in word embeddings that are linear projections of PMI vectors. The loss functions of W2V (2) and GloVe (3) approximate such a projection: non-linearly in the case of W2V and linearly projecting a variant of PMI in GloVe; explaining why their embeddings exhibit semantic properties useful in downstream tasks.

We derive a relationship between embedding matrices **W** and **C**, enabling word embedding interactions (e.g. dot product) to be semantically interpreted and justifying the familiar *cosine similarity* as a measure of relatedness and similarity. Our theoretical results explain several empirical observations, e.g. why **W** and **C** are not found to be equal despite representing the same words, their symmetric treatment in the loss function and a symmetric PMI matrix; why mean embeddings (**A**) are often found to outperform those from either **W** or **C**; and why relatedness corresponds to interactions between **W** and **C**, and similarity to interactions between **W** and **W**.

We discover an interesting hierarchical structure between semantic relationships: with *relatedness* as a basic pairwise comparison, *similarity*, *paraphrase* and *analogy* are defined according to how target words each relate to all words. Error terms arise in the latter higher order relationships due to statistical dependence between words. Such errors can be interpreted geometrically with respect to the hypersurface $\mathcal{S}$ on which all PMI vectors lie, and can, in principle, be evaluated from higher order statistics (e.g. trigram co-occurrences).

Several further details of W2V and GloVe remain to be explained that we hope to address in future work, e.g. the weighting of PMI components over the context window [31], the exponent $^3/_4$ often applied to unigram distributions [26], the probability weighting in the loss function (S.5), and an interpretation of the weight vector **x** in embedding differences (S.5.2).

**Acknowledgements**

We thank Ivan Titov, Jonathan Mallinson and the anonymous reviewers for helpful comments. Carl Allen and Ivana Balažević were supported by the Centre for Doctoral Training in Data Science, funded by EPSRC (grant EP/L016427/1) and the University of Edinburgh.

## A  The W2V shift

The number of negative samples per observed word pair arises in the optimum of the W2V loss function (4) as the so-called *shift* term, $-\log k$. The shift is of a comparable magnitude to empirical PMI values [2] and causes dot product interactions to take more negative values, distorting embeddings relative to there being no shift term.

Under certain word embedding interactions, e.g. the linear combination associated with analogies, the shift terms cancel and thus have no effect [2]. However, elsewhere the shift term has been seen to have a detrimental impact on downstream task performance that removing it corrects [28].

Stemming from an arbitrarily chosen hyper-parameter $k$, the shift term is an artefact of the W2V algorithm that vanishes only if $k\!=\!1$. Setting that explicitly reduces the number of negative samples and results in poorer performance of the embeddings. Alternatively, $k$ can be *effectively* set to 1 by averaging the loss function components of each set of $k$ negative samples, i.e. multiplying by $\frac{1}{k}$.

## B  Properties of the PMI surface: proofs (Sec 4.1)

P1 **$\mathcal{S}$, and any subsurface of $\mathcal{S}$, is non-linear.**  This follows directly from the construction of $\mathcal{S}$, in particualr the application of the natural logarithm to the linear surface $\mathcal{R}$.

P2 **$\mathcal{S}$ contains the origin**  Follows from construction: $\mathbf{p} = \mathbf{p} \in \mathcal{Q}$ implies $\mathbf{1} = \frac{\mathbf{P}}{p(\mathcal{E})} \in \mathcal{R}$, and therefore $\mathbf{0}\!=\!\log \mathbf{1} \!\in\! \mathcal{S}$

P3 **Probability vector $\mathbf{q} \in \mathcal{Q}$ is normal to the tangent plane of $\mathcal{S}$**  at $\mathbf{s} = \log \frac{\mathbf{q}}{\mathbf{p}} \in \mathcal{S}$. Consider $\mathbf{q} = (q_1, ..., q_n) \in \mathcal{Q}$ as having free parameters $q_{j<n}$ that determine $q_n$, and let $\mathbf{J} \in \mathbb{R}^{n \times (n-1)}$ define the tangent plane to $\mathcal{S}$ at $\mathbf{s}$, i.e. $\mathbf{J}_{i,j} = \frac{\partial s_i}{\partial q_j}$. It can be seen that for $i\!<\!n$, $\mathbf{J}_{i,j} = q_j^{-1}$ if $i\!=\!j$, and $\mathbf{J}_{i,j}\!=\!0$ otherwise; and that $\mathbf{J}_{n,j}\!=\!-(1 - \sum_{j=1}^{n-1} q_j)^{-1} = -q_n^{-1}$, $\forall j$. It follows that $\mathbf{q}^\top \mathbf{J} = \mathbf{0}$ and $\mathbf{q}$ is therefore normal to the tangent plane.

P4 **$\mathcal{S}$ does not intersect with the fully positive or fully negative orthants**  (excluding $\mathbf{0}$). This follows from the fact that components of one probability distribution, e.g. $p(\mathcal{E}|w_i)$, cannot *all* be greater (or *all* less) than their counterpart in another, e.g. $p(\mathcal{E})$. Any point in the fully positive or fully negative orthants would contradict this.

P5 **The sum of 2 points $\mathbf{s} + \mathbf{s}'$ lies in $\mathcal{S}$ only for certain $\mathbf{s}, \mathbf{s}' \in \mathcal{S}$.**  For probability vectors $\mathbf{p}$, $\mathbf{q}$, $\mathbf{q}' \in \mathcal{Q}$ and $\mathbf{s} = \log(\mathbf{q}/\mathbf{p})$, $\mathbf{s}' = \log(\mathbf{q}'/\mathbf{p}) \in \mathcal{S}$, we consider operations element-wise with corresponding vector elements denoted by lower case: $\mathbf{s} + \mathbf{s}' \in \mathcal{S}$ iff $s + s' = \log(q^*/p)$ for some probability vector $\mathbf{q}^* \in \mathcal{Q}$. Thus, $(q/p)(q'/p) = q^*/p$, or simply $(q/p)q' = q^*$, whereby components $(q/p)q'$ must sum to 1, or in vector terms $(\mathbf{q}/\mathbf{p})^\top \mathbf{q}' = 1$. since $\mathbf{q}'$ is a probability we can also say $(\mathbf{q}/\mathbf{p} - \mathbf{1})^\top \mathbf{q}' = 0$, and we have that $\mathbf{s} + \mathbf{s}' \in \mathcal{S}$ only if $\mathbf{s}' = \log(\mathbf{q}'/\mathbf{p}) \in \mathcal{S}$ with $\mathbf{q}'$ a probability vector orthogonal to $(\mathbf{q}/\mathbf{p}) - \mathbf{1}$. We see that the intersection of the hyperplane orthogonal to $(\mathbf{q}/\mathbf{p}) - \mathbf{1}$ and the simplex defines points $\mathbf{q}'$ that correspond to points in $\mathbf{s}' \in \mathcal{S}$ that can be added to $\mathbf{s}$, i.e. $\mathcal{S}_s$ (See Figs 3a and 3b). Trivially $\mathbf{0} \in \mathcal{S}_s$, $\forall \mathbf{s} \in \mathcal{S}$.

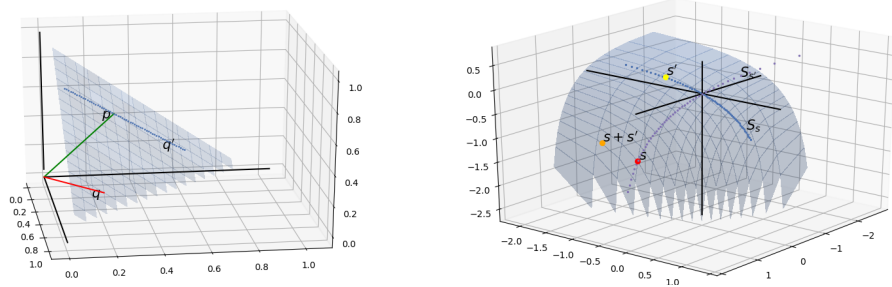

(a) Given point $\mathbf{s}' = \log \mathbf{q}'/\mathbf{p} \in \mathcal{S}$, those $\mathbf{q}'$ on simplex $\mathcal{Q}$ such that $\mathbf{s}' = \log \mathbf{q}'/\mathbf{p}$ satisfies $\mathbf{s} + \mathbf{s}' \in \mathcal{S}$.

(b) Subsurfaces $\mathcal{S}_s$ for a given point $\mathbf{s} \in \mathcal{S}$, and $\mathcal{S}_{s'}$ for any point $\mathbf{s}' \in \mathcal{S}_s$; showing also $\mathbf{s} + \mathbf{s}' \in \mathcal{S}$

Figure 3: Understanding the PMI surface $\mathcal{S}$.

## C  Further Geometric properties of the PMI surface

Combining both geometric and probabilistic arguments shows:

1. PMI vectors of words $w_j$ that are both conditionally and marginally independent of word $w_i$, lie in a strict subsurface $\mathcal{S}_{\mathbf{p}^i} \subset \mathcal{S}$;

2. only $\mathbf{p}^j \in \mathcal{S}_{\mathbf{p}^i}$ add to $\mathbf{p}^i$ to give another point on the surface, specifically $\mathbf{p}^i + \mathbf{p}^j = \mathbf{p}^{i,j}$ corresponding to the joint occurrence of $w_i$ and $w_j$;

3. for any $\mathbf{p}^j \notin \mathcal{S}_{\mathbf{p}^i}$, $\mathbf{p}^i + \mathbf{p}^j$ is off the surface, separated from $\mathbf{p}^{\{w,w'\}}$ by an error vector $\epsilon_{i,j}$, reflecting statistical dependence between $w_i$ and $w_j$.

4. By symmetry, $\mathbf{s}' \in \mathcal{S}_\mathbf{s}$ iff $\mathbf{s} \in \mathcal{S}_{\mathbf{s}'}$, thus subsurfaces occur in distinct pairs $(\mathcal{S}_\mathbf{s}, \mathcal{S}_{\mathbf{s}'})$ that partition all points in $\mathcal{S}$. Furthermore, for any pair of points $\mathbf{t} \in \mathcal{S}_\mathbf{s}, \mathbf{t}' \in \mathcal{S}_{\mathbf{s}'}$, their sum $\mathbf{t} + \mathbf{t}' \in \mathcal{S}$ and every $s \in \mathcal{S}$ is the sum of a unique such pair, which we denote $\mathcal{S}_\mathbf{s} \oplus \mathcal{S}_{\mathbf{s}'} = \mathcal{S}$, analogous to the Cartesian product.

## D  Comparison to embedding relationships of previous works

The following relationships between W2V embeddings and probabilities are assumed in [4]:

$$\mathbf{w}_i = \mathbf{c}_i, \quad \log p(w_i) \approx \tfrac{\|\mathbf{w}_i\|^2}{2d} - \log Z \quad \text{and} \quad \log p(w_i, c_j) \approx \tfrac{\|\mathbf{w}_i + \mathbf{w}_j\|^2}{2d} - 2\log Z,$$

By rearranging $\mathbf{w}_i^\top \mathbf{c}_j \approx \text{PMI}(w_i, c_j)$, as is claimed to follow from those above, we prove (below):

$$\log p(w_i) \approx \tfrac{-\mathbf{w}_i^\top \mathbf{c}_i}{2} + \tfrac{\log p(w_i, c_i)}{2} \quad \text{and} \quad \log p(w_i, c_j) \approx \tfrac{-(\mathbf{w}_i - \mathbf{w}_j)^\top (\mathbf{c}_i - \mathbf{c}_j)}{2} + \tfrac{\log p(w_i, c_i) p(w_j, c_j)}{2}.$$

Having previously shown that $\mathbf{w}_i \neq \mathbf{c}_i$ (Sec 5.1), if we nevertheless assume that equality for the sake of comparison, it can be seen that the relationships above differ fundamentally, e.g. having opposite sign. Also, the assumed *constant* $Z$ can be seen to vary arbitrarily with the extent to which each word co-occurs with itself.

### D.1  Proofs

Noting $p(w_i) = p(c_i)$, since the difference is only the role attributed to a word, shows:

$$\mathbf{w}_i^\top \mathbf{c}_j \approx \log \tfrac{p(w_i, c_j)}{p(w_i)p(c_j)} = \log p(w_i, c_j) - \log p(w_i) - \log p(w_j) \tag{11}$$

If $i = j$, i.e. target and context words are the same, it follows that:

$$\mathbf{w}_i^\top \mathbf{c}_i \approx \log p(w_i, c_i) - 2\log p(w_i)$$

$$\text{i.e.} \quad \log p(w_i) \approx \tfrac{-\mathbf{w}_i^\top \mathbf{c}_i}{2} + \tfrac{\log p(w_i, c_i)}{2} \tag{12}$$

In the general case:

$$\begin{aligned}
(\mathbf{w}_i - \mathbf{w}_j)^\top (\mathbf{c}_i - \mathbf{c}_j) &= \mathbf{w}_i^\top \mathbf{c}_i - \mathbf{w}_j^\top \mathbf{c}_i - \mathbf{w}_i^\top \mathbf{c}_j + \mathbf{w}_j^\top \mathbf{c}_j \\
&\stackrel{*}{=} \mathbf{w}_i^\top \mathbf{c}_i + \mathbf{w}_j^\top \mathbf{c}_j, -2\mathbf{w}_i^\top \mathbf{c}_j \\
&\stackrel{(11,12)}{\approx} (\log p(w_i, c_i) - 2\log p(w_i)) + (\log p(w_j, c_j) - 2\log p(w_j)) \\
&\quad - 2(\log p(w_i, c_j) - \log p(w_i) - \log p(w_j)) \\
&= \log p(w_i, c_i) + \log p(w_j, c_j) - 2\log p(w_i, c_j)
\end{aligned}$$

$$\text{thus} \quad \log p(w_i, c_j) \approx \tfrac{-(\mathbf{w}_i - \mathbf{w}_j)^\top (\mathbf{c}_i - \mathbf{c}_j)}{2} + \tfrac{\log p(w_i, c_i) p(w_j, c_j)}{2}. \tag{13}$$

The step marked * relies on $\mathbf{w}_i^\top \mathbf{c}_j = \mathbf{w}_i^\top (\mathbf{I}' \mathbf{w}_j) = (\mathbf{w}_i^\top \mathbf{I}')\mathbf{w}_j = \mathbf{c}_i^\top \mathbf{w}_j = \mathbf{w}_j^\top \mathbf{c}_i$, which follows from $\mathbf{C} = \mathbf{I}' \mathbf{W}$.

# E  Why order matters in analogies

Here, we develop the explanation of [2] to interpret the finding of Linzen [23] that some words within a particular analogy are more accurately predicted than others (see their "Reverse (add)").

From [2], we see that for analogy "$w_a$ is to $w_{a^*}$ as $w_b$ is to $w_{b^*}$", a "total error" term arises in the relationship $\mathbf{p}^{b^*} + \mathbf{p}^a = \mathbf{p}^{a^*} + \mathbf{p}^b$ between PMI vectors, and thus also word embeddings, due to statistical interactions between word pairs $\{w_a, w_{b^*}\}$ and $\{w_b, w_{a^*}\}$. Thus if $w_{b^*}$ is considered "missing" and to be predicted to complete the analogy, the statistical independence with $w_a$ is relevant, whereas if $w_b$ is to be predicted, statistical independence with $w_{a^*}$ is relevant. One of these may happen to exhibit higher independence, thus introduces less error and so be "easier to predict".

Separately, PMI vectors are unevenly distributed due to the non-uniform Zipf distribution of words. As such, some PMI vectors may happen to lie in more "cluttered" regions than others, an effect that may be exacerbated when projected to the far fewer dimensions of word embeddings. Thus, for the same magnitude error terms, words whose PMI vectors lie in more cluttered regions may be "harder to predict" due to many potential false positives nearby.

These two reasons explain (more concretely that the intuition of [23]) why the same analogy might more accurately be solved by predicting $w_b$ rather than $w_{b^*}$, or vice versa.

# F  Experimental details

## F.1  Training

PMI values are pre-computed from the corpus similarly to [29], substituting −1 for missing PMI values. We use the *text8* data set [24] containing $c.17$m tokens and $c.0.5$m unique words (sourced from the English Wikipedia dump, 03/03/06). 5 random word pairs (negative samples) are generated for each true word co-occurrence (positive sample) according to unigram word distributions. Dimensionality is 500. Words appearing less than 5 times are filtered and down-sampling is applied (see [26]). All models converged within 100 epochs (full passes over the PMI matrix). Learning rates that worked well were selected for each model: 0.01 for the least squares models, 0.007 for the W2V loss function. Results are averaged over 3 random seeds.

## F.2  Testing

Embeddings are evaluated on relatedness, similarity and analogy tasks using *WordSim353* [11, 1]. Ranking is by cosine similarity and evaluation compares Spearman's correlation between rankings and human-assigned similarity scores. Analogies use Google's analogy data set [25] of $c.20$k semantic and syntactic analogy questions "$w_a$ is to $w_{a^*}$ as $w_b$ is to ..?". Out-of-vocabulary words are filtered as standard [21]. Accuracy is computed by comparing $\operatorname{argmin}_{w_{b^*}} \|\mathbf{w}_a - \mathbf{w}_{a^*} - \mathbf{w}_b + \mathbf{w}_{b^*}\|$ to the labelled answer.

## 4.3   Impact

According to Google Scholar, the paper has received 12 citations as of August 2021. As with *Analogies Explained*, the potential impact of this paper may extend to the many other domains where the SGNS algorithm has been applied. The paper provides theoretical grounding for previous heuristics employed by practitioners, such as the use of *average* embeddings.

## 4.4   Discussion

*What the Vec* extends the previous findings of *Analogies Explained* to give a fuller understanding of PMI-based embeddings, their domain and how their interactions entail semantic meaning. The paper introduces two novel perspectives that may be useful to pursue in future work:

- a geometric view of the *PMI surface* $\mathcal{S}$, the domain of word embeddings, a $d$-dimensional hyper-surface in $|\mathcal{E}|$ dimensions (where $d$ and $|\mathcal{E}|$ are typically of order $10^2$–$10^3$ and $10^5$–$10^6$, resp.). This perspective allows meaningful probabilistic quantities, e.g. KL divergences, to be considered geometrically, which may lead to more principled word embedding comparison metrics than current heuristics, such as cosine similarity.

- a potential parallel to kernel theory: kernels allow objects to be numerically represented, $w_i \to \phi(w_i)$, and compared by inner product $\phi(w_i)^\top \phi(w_j)$; word embeddings enumerate words $\boldsymbol{w}_i = \phi(w_i)$, which are compared by a dot product $\boldsymbol{w}_i^\top \boldsymbol{w}_j$ (we note that the PMI matrix factorised by word embeddings is not positive semi-definite, as required of true kernels).

While the paper clarifies several points left outstanding in *Analogies Explained* and related works, various questions remain in terms of what word embeddings learn and how they capture semantics. We note these and other limitations we see in the paper:

- to consider interactions of word embeddings analytically, the paper assumes embeddings to be derived from PMI vectors under *unweighted, linear* projection, whereas projections are weighted in SGNS and GloVe and also non-linear for SGNS. More accurate analysis of the loss functions may be possible, e.g. by considering their Taylor approximations.

- several aspects of word embedding algorithms remain unexplained, e.g. the effect of context window size and the raising of marginal noise distribution probabilities to the power 0.75,

- two mathematical interpretations are given for the *relatedness* of words $w_i, w_j$:

    (i) the single value $\text{PMI}(w_i, w_j)$ that reflects their direct correlation, as might capture relationships between words such as *hot* and *dog*; and

    (ii) a relationship between full PMI-vectors, as might capture *thematic* relatedness of words that co-occur with many common words, e.g. *milk* and *cheese*.

    Differentiating types of relatedness may require more precise linguistic definitions and corresponding data sets, e.g. as considered by Kacmajor and Kelleher (2020);

- As noted previously for *Analogies Explained*, the variable performance of the vector offset method remains unexplained, it is also unclear whether the probabilistic definition of analogies, based on word transformations, reflects all semantic relation types.

Picking up on the last point, there are in fact several indications that the definition of analogies in *Analogies Explained*, as word pairs that share a common word transformation, may not be appropriate for all semantic relations:

(i) word transformations are 1-to-1 (subject to synonyms) whereas relations can also be 1-to-many, many-to-1 or many-to-many;

(ii) two entities may be related by non-equivalent relations that require distinct representations, e.g. $\langle kitten, young\_of, cat \rangle$, $\langle kitten, smaller\_than, cat \rangle$ (Hakami et al., 2018), whereas there is only one vector offset between their embeddings;

(iii) performance of the vector offset method is observed to vary significantly for relations of different semantic types (Levy and Goldberg, 2014a; Köper et al., 2015; Linzen, 2016; Drozd et al., 2016); and

(iv) the KGR model *TransE* (Bordes et al., 2013), which implements the vector offset method, is found to be outperformed by other KGR models, e.g. *MuRE* (Balažević et al., 2019b).

Although *Analogies Explained* describes a family of analogy relations that correspond to vector offsets, point (ii) above proves that a vector offset *cannot* apply to all semantic relation types if distinct relations between a common word pair are to have distinct representations. Furthermore, recent research in human cognition concludes that the vector offset is "limited in the range of semantic relations that it can capture" and in explaining human analogy judgements (Peterson et al., 2020). Hence, understanding more expressive relation representations, which subsume the vector offset, may help us consider how humans reason about concepts and their relations. Alternative functions for representing relations might differ *structurally*, e.g. vector offset vs a bi-linear operator (Hakami et al., 2018), or *parametrically*, as in knowledge graph representation models. However, as discussed in §3.1, there are limitations to how analogy relations can be modelled given only two word embeddings (of $a, a^*$), hence there may be an upper bound on how well analogy relations can be represented. To model each relation on a bespoke basis requires multiple related word pairs, as in *labelled* analogy sets (Bollegala et al., 2015; Drozd et al., 2016) or knowledge graphs.

# Chapter 5

# Interpreting Knowledge Graph Representation from Word Embeddings

The main contribution of this chapter is the paper *Interpreting Knowledge Graph Relation Representation from Word Embeddings* (**Interpreting KGs**), which was published at the *International Conference on Learning Representations* in May 2021. We first outline the motivation for this work (§5.1), before including the paper itself (§5.2), followed by a summary of its impact so far (§5.3) and a discussion (§5.4).

## 5.1 Motivation

The motivation for this paper is that of the overall thesis: to try to understand how knowledge graph entities and their semantic relations can be encoded in the geometry of embeddings and relation representations. This goal is inspired by well-performing knowledge graph representation models that compose entity embeddings and relation representations *linearly* (e.g. Nickel et al., 2011; Bordes et al., 2013; Yang et al., 2015; Trouillon et al., 2016; Balažević et al., 2019c) (§2.2). Such models beg the question how their relatively simple loss functions can capture every day knowledge and enable unknown facts to be predicted. Understanding the latent semantic structure behind these models is both of natural interest in itself and may lead to the development of representation models that achieve greater link prediction performance, are more interpretable or more uncertainty-aware.

The approach of the paper follows from recent understanding of how semantic relations between words can be encoded in the semantic space of PMI-based embeddings (§3, §4). In particular we build on the understanding of the vector offset method for analogies, cognisant of its apparent limitations for representing all semantic relations (§4.4).

## 5.2 The Paper

**Author Contributions**

The paper is co-authored by myself, Ivana Balažević and Timothy Hospedales, with Ivana and myself joint lead authors with equal contribution. The work lies at the intersection of my work towards theoretically understanding how semantics are encoded in the geometry of embeddings (see previous chapters) and Ivana's work on learning knowledge graph representations. Ivana and I co-wrote the paper and, through ongoing discussions, jointly developed the semantic relation types, their relation conditions and the corresponding theoretical score functions. I had the initial idea for the paper based on the findings of my earlier work on word embeddings, which developed into the central theory. Ivana and I discussed all experiments, which Ivana implemented and ran. Tim provided useful suggestions and helped in revising the final paper.

# INTERPRETING KNOWLEDGE GRAPH RELATION REPRESENTATION FROM WORD EMBEDDINGS

**Carl Allen**[1*], **Ivana Balažević**[1*] **& Timothy Hospedales**[1,2]
[1] University of Edinburgh, UK    [2] Samsung AI Centre, Cambridge, UK
{carl.allen, ivana.balazevic, t.hospedales}@ed.ac.uk

## ABSTRACT

Many models learn representations of knowledge graph data by exploiting its low-rank latent structure, encoding known relations between entities and enabling unknown facts to be inferred. To predict whether a relation holds between entities, embeddings are typically compared in the latent space following a relation-specific mapping. Whilst their predictive performance has steadily improved, how such models capture the underlying latent structure of semantic information remains unexplained. Building on recent theoretical understanding of word embeddings, we categorise knowledge graph relations into three types and for each derive explicit requirements of their representations. We show that empirical properties of relation representations and the relative performance of leading knowledge graph representation methods are justified by our analysis.

## 1 INTRODUCTION

Knowledge graphs are large repositories of binary relations between words (or entities) in the form of *(subject, relation, object)* triples. Many models for representing entities and relations have been developed, so that known facts can be recalled and previously unknown facts can be inferred, a task known as *link prediction*. Recent link prediction models (e.g. Bordes et al., 2013; Trouillon et al., 2016; Balažević et al., 2019b) learn entity representations, or *embeddings*, of far lower dimensionality than the number of entities, by capturing latent structure in the data. Relations are typically represented as a mapping from the embedding of a subject entity to those of related object entities. Although the performance of link prediction models has steadily improved for nearly a decade, relatively little is understood of the low-rank latent structure that underpins them, which we address in this work. The outcomes of our analysis can be used to aid and direct future knowledge graph model design.

We start by drawing a parallel between the entity embeddings of knowledge graphs and context-free word embeddings, e.g. as learned by Word2Vec (W2V) (Mikolov et al., 2013a) and GloVe (Pennington et al., 2014). Our motivating premise is that the same latent word features (e.g. meaning(s), tense, grammatical type) give rise to the patterns found in different data sources, i.e. manifesting in word co-occurrence statistics and determining which words relate to which. Different embedding approaches may capture such structure in different ways, but if it is fundamentally the same, an understanding gained from one embedding task (e.g. word embedding) may benefit another (e.g. knowledge graph representation). Furthermore, the relatively limited but accurate data used in knowledge graph representation differs materially from the highly abundant but statistically noisy text data used for word embeddings. As such, theoretically reconciling the two embedding methods may lead to unified and improved embeddings learned jointly from both data sources.

Recent work (Allen & Hospedales, 2019; Allen et al., 2019) theoretically explains how semantic properties are encoded in word embeddings that (approximately) factorise a matrix of *pointwise mutual information* (PMI) from word co-occurrence statistics, as known for W2V (Levy & Goldberg, 2014). *Semantic* relationships between words, specifically similarity, relatedness, paraphrase and analogy, are proven to manifest as linear *geometric* relationships between rows of the PMI matrix (subject to known error terms), of which word embeddings can be considered low-rank projections. This explains, for example, the observations that similar words have similar embeddings and that embeddings of analogous word pairs share a common "vector offset" (e.g. Mikolov et al., 2013b).

---
[*]Equal contribution

**Table 1:** Score functions of representative linear link prediction models. $\boldsymbol{R} \in \mathbb{R}^{d_e \times d_e}$ and $\boldsymbol{r} \in \mathbb{R}^{d_e}$ are the relation matrix and translation vector, $\mathbf{W} \in \mathbb{R}^{d_e \times d_r \times d_e}$ is the core tensor and $b_s, b_o \in \mathbb{R}$ are the entity biases.

| Model    |                          | Linear Subcategory                 | Score Function                                                      |
|----------|--------------------------|------------------------------------|---------------------------------------------------------------------|
| TransE   | (Bordes et al., 2013)    | additive                           | $-\|\boldsymbol{e}_s + \boldsymbol{r} - \boldsymbol{e}_o\|_2^2$     |
| DistMult | (Yang et al., 2015)      | multiplicative (diagonal)          | $\boldsymbol{e}_s^\top \boldsymbol{R} \boldsymbol{e}_o$             |
| TuckER   | (Balažević et al., 2019b)| multiplicative                     | $\mathbf{W} \times_1 \boldsymbol{e}_s \times_2 \boldsymbol{r} \times_3 \boldsymbol{e}_o$ |
| MuRE     | (Balažević et al., 2019a)| multiplicative (diagonal) + additive | $-\|\boldsymbol{R}\boldsymbol{e}_s + \boldsymbol{r} - \boldsymbol{e}_o\|_2^2 + b_s + b_o$ |

We extend this insight to identify geometric relationships between PMI-based word embeddings that correspond to other relations, i.e. those of knowledge graphs. Such *relation conditions* define relation-specific mappings between entity embeddings (i.e. *relation representations*) and so provide a "blue-print" for knowledge graph representation models. Analysing the relation representations of leading knowledge graph representation models, we find that various properties, including their relative link prediction performance, accord with predictions based on these relation conditions, supporting the premise that a *common latent structure* is learned by word and knowledge graph embedding models, despite the significant differences between their training data and methodology.

In summary, the key contributions of this work are:
* to use recent understanding of PMI-based word embeddings to derive geometric attributes of a relation representation for it to map subject word embeddings to all related object word embeddings (*relation conditions*), which partition relations into three *types* (§3);
* to show that both per-relation ranking as well as classification performance of leading link prediction models corresponds to the model satisfying the appropriate relation conditions, i.e. how closely its relation representations match the geometric form derived theoretically (§4.1); and
* to show that properties of knowledge graph representation models fit predictions based on relation conditions, e.g. the strength of a relation's *relatedness* aspect is reflected in the eigenvalues of its relation matrix (§4.2).

## 2 Background

**Knowledge graph representation:** Recent knowledge graph models typically represent entities $e_s, e_o$ as vectors $\boldsymbol{e}_s, \boldsymbol{e}_o \in \mathbb{R}^{d_e}$, and relations as transformations in the latent space from subject to object entity embedding, where the dimension $d_e$ is far lower (e.g. 200) than the number of entities $n_e$ (e.g. $>10^4$). Such models are distinguished by their *score function*, which defines (i) the form of the relation transformation, e.g. matrix multiplication and/or vector addition; and (ii) the measure of proximity between a transformed subject embedding and an object embedding, e.g. dot product or Euclidean distance. Score functions can be non-linear (e.g. Dettmers et al., 2018), or linear and sub-categorised as *additive*, *multiplicative* or both. We focus on linear models due to their simplicity and strong performance at link prediction (including state-of-the-art). Table 1 shows the score functions of competitive linear knowledge graph embedding models spanning the sub-categories: TransE (Bordes et al., 2013), DistMult (Yang et al., 2015), TuckER (Balažević et al., 2019b) and MuRE (Balažević et al., 2019a).

*Additive models* apply a relation-specific translation to a subject entity embedding and typically use Euclidean distance to evaluate proximity to object embeddings. A generic additive score function is given by $\phi(e_s, r, e_o) = -\|\boldsymbol{e}_s + \boldsymbol{r} - \boldsymbol{e}_o\|_2^2 + b_s + b_o$. A simple example is TransE, where $b_s = b_o = 0$.

*Multiplicative models* have the generic score function $\phi(e_s, r, e_o) = \boldsymbol{e}_s^\top \boldsymbol{R} \boldsymbol{e}_o$, i.e. a bilinear product of the entity embeddings and a relation-specific matrix $\boldsymbol{R}$. DistMult is a simple example with $\boldsymbol{R}$ diagonal and so cannot model asymmetric relations (Trouillon et al., 2016). In TuckER, each relation-specific $\boldsymbol{R} = \mathbf{W} \times_3 \boldsymbol{r}$ is a linear combination of $d_r$ "prototype" relation matrices in a core tensor $\mathbf{W} \in \mathbb{R}^{d_e \times d_r \times d_e}$ ($\times_n$ denoting tensor product along mode $n$), facilitating *multi-task learning* across relations.

Some models, e.g. MuRE, combine both multiplicative ($\boldsymbol{R}$) and additive ($\boldsymbol{r}$) components.

**Word embedding:** Algorithms such as Word2Vec (Mikolov et al., 2013a) and GloVe (Pennington et al., 2014) generate low-dimensional word embeddings that perform well on downstream tasks (Baroni et al., 2014). Such models predict the context words ($c_j$) observed around a target word ($w_i$) in a text corpus using shallow neural networks. Whilst recent language models (e.g. Devlin et al., 2018; Peters et al., 2018) achieve strong performance using *contextualised* word embeddings, we focus on "context-free" embeddings since knowledge graph entities have no obvious context and, importantly, they offer insight into embedding interpretability.

Levy & Goldberg (2014) show that, for a dictionary of $n_e$ unique words and embedding dimension $d_e \ll n_e$, W2V's loss function is minimised when its embeddings $\boldsymbol{w}_i, \boldsymbol{c}_j$ form matrices $\boldsymbol{W}, \boldsymbol{C} \in \mathbb{R}^{d_e \times n_e}$ that factorise a *pointwise mutual information* (PMI) matrix of word co-occurrence statistics (PMI$(w_i, c_j) = \log \frac{P(w_i, c_j)}{P(w_i)P(c_j)}$), subject to a shift term. This result relates W2V to earlier count-based embeddings and specifically PMI, which has a history in linguistic analysis (Turney & Pantel, 2010). From its loss function, GloVe can be seen to perform a related factorisation.

Recent work (Allen & Hospedales, 2019; Allen et al., 2019) shows how the semantic relationships of *similarity*, *relatedness*, *paraphrase* and *analogy* are encoded in PMI-based word embeddings by recognising such embeddings as low-rank projections of high dimensional rows of the PMI matrix, termed *PMI vectors*. Those semantic relationships are described in terms of *multiplicative* interactions between co-occurrence probabilities (subject to defined error terms), that correspond to *additive* interactions between (logarithmic) PMI statistics, and hence PMI vectors. Thus, under a sufficiently linear projection, those semantic relationships correspond to linear relationships between word embeddings. Note that although the relative geometry reflecting semantic relationships is preserved, the direct interpretability of dimensions, as in PMI vectors, is lost since the embedding matrices can be arbitrarily scaled/rotated if the other is inversely transformed. We state the relevant semantic relationships on which we build, denoting the set of unique dictionary words by $\mathcal{E}$:

- **Paraphrase**: word subsets $\mathcal{W}, \mathcal{W}^* \subseteq \mathcal{E}$ are said to *paraphrase* if they induce similar distributions over nearby words, i.e. $p(\mathcal{E}|\mathcal{W}) \approx p(\mathcal{E}|\mathcal{W}^*)$, e.g. {*king*} paraphrases {*man, royal*}.
- **Analogy**: a common example of an *analogy* is "*woman* is to *queen* as *man* is to *king*" and can be defined as any set of word pairs $\{(w_i, w_i^*)\}_{i \in \mathcal{I}}$ for which it is semantically meaningful to say "$w_a$ is to $w_a^*$ as $w_b$ is to $w_b^*$" $\forall a, b \in \mathcal{I}$.

Where one word subset paraphrases another, the sums of their embeddings are shown to be equal (subject to the independence of words within each set), e.g. $\boldsymbol{w}_{king} \approx \boldsymbol{w}_{man} + \boldsymbol{w}_{royal}$. An interesting connection is established between the two semantic relationships: a set of word pairs $\mathcal{A} = \{(w_a, w_a^*), (w_b, w_b^*)\}$ is an analogy if $\{w_a, w_b^*\}$ paraphrases $\{w_a^*, w_b\}$, in which case the embeddings satisfy $\boldsymbol{w}_{a^*} - \boldsymbol{w}_a \approx \boldsymbol{w}_{b^*} - \boldsymbol{w}_b$ ("vector offset").

## 3 FROM ANALOGIES TO KNOWLEDGE GRAPH RELATIONS

Analogies from the field of word embeddings are our starting point for developing a theoretical basis for representing knowledge graph relations. The relevance of analogies stems from the observation that for an analogy to hold (see §2), its word pairs, e.g {*(man, king), (woman, queen), (girl, princess)*}, must be *related* in the same way, comparably to subject-object entity pairs under a common knowledge graph relation. Our aim is to develop the understanding of PMI-based word embeddings (henceforth *word embeddings*), to identify the mathematical properties necessary for a relation representation to map subject word embeddings to all related object word embeddings.

Considering the paraphrasing word sets {*king*} and {*man, royal*} corresponding to the word embedding relationship $\boldsymbol{w}_{king} \approx \boldsymbol{w}_{man} + \boldsymbol{w}_{royal}$ (§2), *royal* can be interpreted as the semantic difference between *man* and *king*, fitting intuitively with the relationship $\boldsymbol{w}_{royal} \approx \boldsymbol{w}_{king} - \boldsymbol{w}_{man}$. Fundamentally, this relationship holds because the difference between words that co-occur (i.e. occur more frequently than if independent) with *king* and those that co-occur with *man*, reflects those words that co-occur with *royal*. We refer to this difference in co-occurrence distribution as a "context shift", from *man* (subject) to *king* (object). Allen & Hospedales (2019) effectively show that where multiple word pairs share a common context shift, they form an analogy whose embeddings satisfy the vector offset relationship. This result seems obvious where the context shift mirrors an identifiable word, the embedding of which is approximated by the common vector offset, e.g. *queen* and *woman* are related by the same context shift, i.e. $\boldsymbol{w}_{queen} \approx \boldsymbol{w}_{woman} + \boldsymbol{w}_{royal}$, thus $\boldsymbol{w}_{queen} - \boldsymbol{w}_{woman} \approx \boldsymbol{w}_{king} - \boldsymbol{w}_{man}$. However, the same result holds, i.e. an analogy is formed with a common vector offset between embeddings, for an arbitrary (common) context shift that may reflect no particular word. Importantly, these context shift relations evidence a case in which it is known how a relation can be represented, i.e. by an additive vector (comparable to TransE) *if* entities are represented by word embeddings. More generally, this provides an interpretable foothold into relation representation.

Note that not all sets of word pairs considered analogies exhibit a clear context shift relation, e.g. in the analogy {(*car,engine*), (*bus,seats*)}, the difference between words co-occurring with *engine* and *car* is not expected to reflect the corresponding difference between *bus* and *seats*. This illustrates how analogies are a loosely defined concept, e.g. their implicit relation may be semantic or syntactic, with several sub-categories of each (e.g. see Gladkova et al. (2016)). The same is readily observed for the relations of knowledge graphs. This likely explains the observed variability in "solving" analogies by use of vector offset (e.g. Köper et al., 2015; Karpinska et al., 2018; Gladkova et al., 2016) and suggests that further consideration is required to represent relations (or solve analogies) in general.

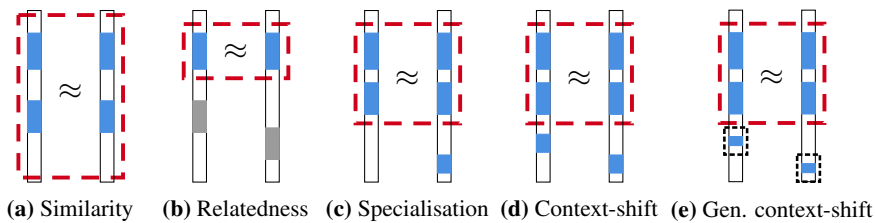

(a) Similarity  (b) Relatedness  (c) Specialisation  (d) Context-shift  (e) Gen. context-shift

**Figure 1:** Relationships between PMI vectors (black rectangles) of subject/object words for different relation *types*. PMI vectors capture co-occurrence with every dictionary word: strong associations (PMI > 0) are shaded (blue define the relation, grey are random other associations); red dash = *relatedness*; black dash = *context sets*.

We have seen that the existence of a context shift relation between a subject and object word implies a (relation-specific) geometric relationship between word embeddings, thus the latter provides a *necessary condition for the relation to hold*. We refer to this as a "relation condition" and aim to identify relation conditions for other classes of relation. Once identified, relation conditions define a mapping from subject embeddings to all related object embeddings, by which related entities might be identified with a proximity measure (e.g. Euclidean distance or dot product). This is the precise aim of a knowledge graph representation model, but loss functions are typically developed heuristically. Given the existence of many representation models, we can verify identified relation conditions by contrasting the per-relation performance of various models with the extent to which their loss function reflects the appropriate relation conditions. Note that since relation conditions are necessary rather than sufficient, they do not guarantee a relation holds, i.e. false positives may arise.

Whilst we seek to establish relation conditions based on PMI word embeddings, the data used to train knowledge graph embeddings differs significantly to the text data used by word embeddings, and the relevance of conditions ultimately based on PMI statistics may seem questionable. However, where a knowledge graph representation model implements relation conditions and measures proximity between embeddings, the parameters of word embeddings necessarily provide *a potential* solution that minimises the loss function. Many equivalent solutions may exist due to symmetry as typical for neural network architectures. We now define relation types and identify their relation conditions (underlined); we then consider the completeness of this categorisation.

- **Similarity:** Semantically similar words induce similar distributions over the words they co-occur with. Thus their PMI vectors and <u>word embeddings are similar</u> (Fig 1a).

- **Relatedness:** The relatedness of two words can be considered in terms of the words $\mathcal{S} \subseteq \mathcal{E}$ with which both co-occur similarly. $\mathcal{S}$ defines the *nature* of relatedness, e.g. *milk* and *cheese* are related by $\mathcal{S} = \{dairy, breakfast, ...\}$; and $|\mathcal{S}|$ reflects the *strength* of relatedness. Since PMI vector components corresponding to $\mathcal{S}$ are similar (Fig 1b), embeddings of $\mathcal{S}$-*related* words <u>have similar components in the subspace $\mathbb{V}_\mathcal{S}$</u> that spans the projected PMI vector dimensions corresponding to $\mathcal{S}$. The rank of $\mathbb{V}_\mathcal{S}$ is thus anticipated to reflect relatedness strength. Relatedness can be seen as a weaker and more variable generalisation of similarity, its limiting case where $\mathcal{S} = \mathcal{E}$, hence $\text{rank}(\mathbb{V}_\mathcal{S}) = d_e$.

- **Context-shift:** As discussed above, words related by a common difference between their distributions of co-occurring words, defined as *context-shifts*, share a <u>common vector offset between word embeddings</u>. Context might be considered *added* (e.g. *man* to *king*), termed **specialisation** (Fig 1c), *subtracted* (e.g. *king* to *man*) or both (Fig 1d). These relations are 1-to-1 (subject to synonyms) and include an aspect of *relatedness* due to the word associations in common. Note that, specialisations include hyponyms/hypernyms and context shifts include meronyms.

- **Generalised context-shift:** Context-shift relations generalise to 1-to-many, many-to-1 and many-to-many relations where the added/subtracted context may be from a (relation-specific) *context set* (Fig 1e), e.g. *any* city or *anything* bigger. The potential scope and size of context sets adds variability to these relations. The limiting case in which the context set is "small" reduces to a 1-to-1 context-shift (above) and the <u>embedding difference is a known vector offset</u>. In the limiting case of a "large" context set, the added/subtracted context is essentially unrestricted such that only the relatedness aspect of the relation, and thus a <u>common subspace component of embeddings</u>, is fixed.

**Categorisation completeness:** Taking intuition from Fig 1 and considering PMI vectors as *sets of word features*, these relation types can be interpreted as set operations: similarity as set equality; relatedness as subset equality; and context-shift as a relation-specific set difference. Since for any relation each feature must either remain unchanged (relatedness), change (context shift) or else be irrelevant, we conjecture that the above relation types give a complete partition of semantic relations.

Table 2: Categorisation of WN18RR relations.

| Type | Relation | Examples *(subject entity, object entity)* |
|---|---|---|
| R | verb_group | (trim_down_VB_1, cut_VB_35), (hatch_VB_1, incubate_VB_2) |
|   | derivationally_related_form | (lodge_VB_4, accommodation_NN_4), (question_NN_1, inquire_VB_1) |
|   | also_see | (clean_JJ_1, tidy_JJ_1), (ram_VB_2, screw_VB_3) |
| S | hypernym | (land_reform_NN_1, reform_NN_1), (prickle-weed_NN_1, herbaceous_plant_NN_1) |
|   | instance_hypernym | (yellowstone_river_NN_1, river_NN_1), (leipzig_NN_1, urban_center_NN_1) |
| C | member_of_domain_usage | (colloquialism_NN_1, figure_VB_5), (plural_form_NN_1, authority_NN_2) |
|   | member_of_domain_region | (rome_NN_1, gladiator_NN_1), (usa_NN_1, multiple_voting_NN_1) |
|   | member_meronym | (south_NN_2, sunshine_state_NN_1), (genus_carya_NN_1, pecan_tree_NN_1) |
|   | has_part | (aircraft_NN_1, cabin_NN_3), (morocco_NN_1, atlas_mountains_NN_1) |
|   | synset_domain_topic_of | (quark_NN_1, physics_NN_1), (harmonize_VB_3, music_NN_4) |

### 3.1 CATEGORISING REAL KNOWLEDGE GRAPH RELATIONS

Analysing the relations of popular knowledge graph datasets, we observe that they indeed imply (i) a relatedness aspect reflecting a common theme (e.g. both entities are animals or geographic terms); and (ii) contextual themes specific to the subject and/or object entities. Further, relations fall under a hierarchy of three *relation types*: highly related (**R**); generalised specialisation (**S**); and generalised context-shift (**C**). As above, "generalised" indicates that context differences are not restricted to be 1-1. From Fig 1, it can be seen that type R relations are a special case of S, which are a special case of C. Thus type C encompasses all considered relations. Whilst there are many ways to classify relations, e.g. by hierarchy, transitivity, the proposed relation conditions delineate relations by the required mathematical form (and complexity) of their representation. Table 2 shows a categorisation of the relations of the WN18RR dataset (Dettmers et al., 2018) comprising 11 relations and 40,943 entities.[1] An explanation for the category assignment is in Appx. A. Analysing the commonly used FB15k-237 dataset (Toutanova et al., 2015) reveals relations to be almost exclusively of type C, precluding a contrast of performance per relation type and hence that dataset is omitted from our analysis. Instead, we categorise a random subsample of 12 relations from the NELL-995 dataset (Xiong et al., 2017), containing 75,492 entities and 200 relations (see Tables 8 and 9 in Appx. B).

### 3.2 RELATIONS AS MAPPINGS BETWEEN EMBEDDINGS

Given the relation conditions of a relation type, we now consider mappings that satisfy them and thereby loss functions able to identify relations of each type, evaluating proximity between mapped entity embeddings by dot product or Euclidean distance. We then contrast our theoretically derived loss functions, specific to a relation type, with those of several knowledge graph models (Table 1) to predict identifiable properties and the relative performance of different knowledge graph models for each relation type.

**R:** Identifying $\mathcal{S}$-relatedness requires testing both entity embeddings $e_s, e_o$ for a common subspace component $\mathbb{V}_\mathcal{S}$, which can be achieved by projecting both embeddings onto $\mathbb{V}_\mathcal{S}$ and comparing their images. Projection requires multiplication by a matrix $P_r \in \mathbb{R}^{d \times d}$ and cannot be achieved additively, except in the trivial limiting case of similarity ($P_r = I$) when $r \approx 0$ can be added.
Comparison by dot product gives $(P_r e_s)^\top (P_r e_o) = e_s^\top P_r^\top P_r e_o = e_s^\top M_r e_o$ (for relation-specific symmetric $M_r = P_r^\top P_r$). Euclidean distance gives $\|P_r e_s - P_r e_o\|^2 = (e_s - e_o)^\top M_r (e_s - e_o) = \|P_r e_s\|^2 - 2 e_s^\top M_r e_o + \|P_r e_o\|^2$.

**S/C:** The relation conditions require testing for both $\mathcal{S}$-relatedness and relation-specific entity component(s) ($v_r^s, v_r^o$). This is achieved by (i) multiplying both entity embeddings by a relation-specific projection matrix $P_r$ that projects onto the subspace that spans the low-rank projection of dimensions corresponding to $\mathcal{S}$, $v_r^s$ and $v_r^o$ (i.e. testing for $\mathcal{S}$-relatedness while preserving relation-specific entity components); and (ii) adding a relation-specific vector $r = v_r^o - v_r^s$ to the transformed subject entity embeddings.

---
[1] We omit the relation "similar_to" since its instances have no discernible structure, and only 3 occur in the test set, all of which are the inverse of a training example and trivial to predict.

Comparing the transformed entity embeddings by dot product equates to $(\boldsymbol{P}_r\boldsymbol{e}_s + \boldsymbol{r})^\top \boldsymbol{P}_r\boldsymbol{e}_o$; and by Euclidean distance to $\|\boldsymbol{P}_r\boldsymbol{e}_s + \boldsymbol{r} - \boldsymbol{P}_r\boldsymbol{e}_o\|^2 = \|\boldsymbol{P}_r\boldsymbol{e}_s + \boldsymbol{r}\|^2 - 2(\boldsymbol{P}_r\boldsymbol{e}_s + \boldsymbol{r})^\top \boldsymbol{P}_r\boldsymbol{e}_o + \|\boldsymbol{P}_r\boldsymbol{e}_o\|^2$ (*cf* MuRE: $\|\boldsymbol{R}\boldsymbol{e}_s + \boldsymbol{r} - \boldsymbol{e}_o\|^2$).

Contrasting these theoretically derived loss functions with those of knowledge graph models (Table 1), we make the following predictions:

**P1:** The ability to learn the representation of a relation is expected to reflect:
    (a) the complexity of its type (R<S<C) independently of model choice; and
    (b) whether relation conditions (e.g. additive/multiplicative interactions) are met by the model.

**P2:** Knowledge graph relation representations reflect the following type-specific properties:
    (a) relation matrices for relatedness (type R) relations are highly symmetric;
    (b) offset vectors for relatedness relations have low norm; and
    (c) as a proxy to the rank of $\mathbb{V}_\mathcal{S}$, the eigenvalues of a relation matrix reflect relatedness strength.

To elaborate, our core prediction P1(b) anticipates that: (i) additive-only models (e.g. TransE) are not suited to identifying the relatedness aspect of relations, except in limiting cases of similarity, requiring a zero vector); (ii) multiplicative-only models (e.g. DistMult) should perform well on type R relations, but are not suited to identifying entity-specific features of type S/C (an asymmetric relation matrix in TuckER may help compensate); and (iii) the loss function of MuRE closely resembles that derived for type C relations, which generalise all others, and is thus expected to perform best overall.

## 4 EVIDENCE LINKING KNOWLEDGE GRAPH AND WORD EMBEDDINGS

We test whether the predictions P1 and P2, made on the basis of word embeddings, apply to knowledge graph relations by analysing the performance and properties of competitive knowledge graph models. We compare TransE, DistMult, TuckER and MuRE, which entail different forms of relation representation, on all WN18RR relations and a similar number of NELL-995 relations (spanning all relation types). All models have a comparable number of free parameters.

Since for TransE, the logistic sigmoid cannot be applied to the score function to give a probabilistic interpretation comparable to other models, for fair comparison we include MuRE$_I$, a constrained variant of MuRE with $\boldsymbol{R}_s = \boldsymbol{R}_o = \boldsymbol{I}$, as a proxy to TransE. Implementation details are included in Appx. D. For evaluation, we generate $2n_e$ *evaluation triples* for each test triple ($n_e = |\mathcal{E}|$ denoting the number of entities) by fixing the subject entity $e_s$ and relation $r$ and replacing the object entity $e_o$ with each entity in turn and then keeping $e_o$ and $r$ fixed and varying $e_s$. Each model's scores for the evaluation triples are ranked to give the standard metric Hits@10 (Bordes et al., 2013), i.e. the fraction of times a true triple appears in the top 10 ranked evaluation triples.

### 4.1 P1: JUSTIFYING THE RELATIVE PERFORMANCE OF KNOWLEDGE GRAPH MODELS

**Ranking performance:** Tables 3 and 4 report Hits@10 for each relation and include the relation type as well as known confounding influences: percentage of relation instances in the training and test sets (approximately equal), the actual number of instances in the test set (causing some results to be highly granular), Krackhardt hierarchy score (see Appx. E) (Krackhardt, 2014; Balažević et al., 2019a) and maximum and average shortest path between any two related nodes. A further confounding effect is dependence between relations: Lacroix et al. (2018) and Balažević et al. (2019b) independently show that constraining the rank of relation representations is beneficial for datasets with many relations due to *multi-task learning*, particularly when the number of instances per relation is low. This is expected to benefit TuckER on the NELL-995 dataset (200 relations).

As predicted in P1(a), all models tend to perform best at type R relations, with a clear performance gap to other relation types. Also, performance on type S relations appears higher in general than type C. In accordance with P1(b), additive-only models (TransE, MuRE$_I$) perform worst on average since all relation types involve (multiplicative) relatedness. Best performance is achieved on type R relations, which can be represented by a small/zero additive vector. Multiplicative-only DistMult performs well, sometimes best, on type R relations, fitting expectation as it can represent those relations and has no inessential parameters, e.g. that may overfit to noise, which may explain instances where MuRE performs slightly worse. As expected, MuRE, performs best overall (particularly on WN18RR), and most strongly on S and C type relations, predicted to require both multiplicative and additive components. Comparable performance of TuckER on NELL-995 may be explained by its multi-task learning ability.

Table 3: Hits@10 per relation on WN18RR.

| Relation Name | Type | % | # | Khs | Max/Avg Path | | TransE | MuRE$_I$ | DistMult | TuckER | MuRE |
|---|---|---|---|---|---|---|---|---|---|---|---|
| verb_group | R | 1% | 39 | 0.00 | - | - | 0.87 | 0.95 | **0.97** | **0.97** | **0.97** |
| derivationally_related_form | R | 34% | 1074 | 0.04 | - | - | 0.93 | 0.96 | 0.96 | 0.96 | **0.97** |
| also_see | R | 2% | 56 | 0.24 | 44 | 15.2 | 0.59 | **0.73** | 0.67 | 0.72 | **0.73** |
| instance_hypernym | S | 4% | 122 | 1.00 | 3 | 1.0 | 0.22 | 0.52 | 0.47 | 0.53 | **0.54** |
| synset_domain_topic_of | C | 4% | 114 | 0.99 | 3 | 1.1 | 0.19 | 0.43 | 0.42 | 0.45 | **0.53** |
| member_of_domain_usage | C | 1% | 24 | 1.00 | 2 | 1.0 | 0.42 | 0.42 | 0.48 | 0.38 | **0.50** |
| member_of_domain_region | C | 1% | 26 | 1.00 | 2 | 1.0 | 0.35 | 0.40 | 0.40 | 0.35 | **0.46** |
| member_meronym | C | 8% | 253 | 1.00 | 10 | 3.9 | 0.04 | 0.38 | 0.30 | **0.39** | **0.39** |
| has_part | C | 6% | 172 | 1.00 | 13 | 2.2 | 0.04 | 0.31 | 0.28 | 0.29 | **0.35** |
| hypernym | S | 40% | 1251 | 0.99 | 18 | 4.5 | 0.02 | 0.20 | 0.19 | 0.20 | **0.28** |
| all | | 100% | 3134 | | | | 0.38 | 0.52 | 0.51 | 0.53 | **0.57** |

Table 4: Hits@10 per relation on NELL-995.

| Relation Name | Type | % | # | Khs | Max/Avg Path | | TransE | MuRE$_I$ | DistMult | TuckER | MuRE |
|---|---|---|---|---|---|---|---|---|---|---|---|
| teamplaysagainstteam | R | 2% | 243 | 0.11 | 10 | 3.5 | 0.76 | 0.84 | **0.90** | 0.89 | 0.89 |
| clothingtogowithclothing | R | 1% | 132 | 0.17 | 5 | 2.6 | 0.72 | 0.80 | **0.88** | 0.85 | 0.84 |
| professionistypeofprofession | S | 1% | 143 | 0.38 | 7 | 2.5 | 0.37 | 0.55 | 0.62 | 0.65 | **0.66** |
| animalistypeofanimal | S | 1% | 103 | 0.68 | 9 | 3.1 | 0.50 | 0.56 | 0.64 | **0.68** | 0.65 |
| athleteplayssport | C | 1% | 113 | 1.00 | 1 | 1.0 | 0.54 | 0.58 | 0.58 | 0.60 | **0.64** |
| chemicalistypeofchemical | S | 1% | 115 | 0.53 | 6 | 3.0 | 0.23 | 0.43 | 0.49 | 0.51 | **0.60** |
| itemfoundinroom | C | 2% | 162 | 1.00 | 1 | 1.0 | 0.39 | 0.57 | 0.53 | 0.56 | **0.59** |
| agentcollaborateswithagent | R | 1% | 119 | 0.51 | 14 | 4.7 | 0.44 | 0.58 | **0.64** | 0.61 | 0.58 |
| bodypartcontainsbodypart | C | 1% | 103 | 0.60 | 7 | 3.2 | 0.30 | 0.38 | 0.54 | **0.58** | 0.55 |
| atdate | C | 10% | 967 | 0.99 | 4 | 1.1 | 0.41 | 0.50 | 0.48 | 0.48 | **0.52** |
| locationlocatedwithinlocation | C | 2% | 157 | 1.00 | 6 | 1.9 | 0.35 | 0.37 | 0.46 | **0.48** | **0.48** |
| atlocation | C | 1% | 294 | 0.99 | 6 | 1.4 | 0.22 | 0.33 | 0.39 | 0.43 | **0.44** |
| all | | 100% | 20000 | | | | 0.36 | 0.48 | 0.51 | **0.52** | **0.52** |

Other anomalous results also closely align with confounding factors. For example, all models perform poorly on the *hypernym* relation, despite it having a relative abundance of training data (40% of all instances), which may be explained by its *hierarchical* nature (Khs ≈ 1 and long paths). The same may explain the reduced performance on relations *also_see* and *agentcollaborateswithagent*. As found previously (Balažević et al., 2019a), none of the models considered are well suited to modelling hierarchical structures. We also note that the percentage of training instances of a relation is not a dominant factor on performance, as would be expected if all relations could be equally represented.

**Classification performance:** We further evaluate whether P1 holds when comparing knowledge graph models by classification accuracy on WN18RR. Independent predictions of whether a given triple is true or false are not commonly evaluated, instead metrics such as mean reciprocal rank and Hits@$k$ are reported that compare the prediction of a test triple against all evaluation triples. Not only is this computationally costly, the evaluation is flawed if an entity is related to $l > k$ others ($k$ is often 1 or 3). A correct prediction validly falling within the top $l$ but not the top $k$ would appear incorrect under the metric. Some recent works also note the importance of standalone predictions (Speranskaya et al., 2020; Pezeshkpour et al., 2020).

Since for each relation there are $n_e^2$ possible entity-entity relationships, we sub-sample by computing predictions only for those $(e_s, r, e_o)$ triples for which the $e_s, r$ pairs appear in the test set. We split positive predictions ($\sigma(\phi(e_s, r, e_o)) > 0.5$) between (i) *known truths* – training or test/validation instances; and (ii) *other*, the truth of which is not known. We then compute per-relation accuracy over the true training instances ("train") and true test/validation instances ("test"); and the average number of "other" triples predicted true per $e_s, r$ pair. Table 5 shows results for MuRE$_I$, DistMult, TuckER and MuRE. All models achieve near perfect training accuracy. The additive-multiplicative MuRE gives best test set performance, followed (surprisingly) closely by MuRE$_I$, with multiplicative models (DistMult and TuckER) performing poorly on all but type R relations in line with P1(b), with near-zero performance on most type S/C relations. Since the ground truth labels for "other" triples predicted to be true are not in the dataset, we analyse a sample of "other" true predictions for one relation of each type (see Appx. G). From this, we estimate that TuckER is relatively accurate but pessimistic (∼0.3 correct of the 0.5 predictions ≈ 60%), MuRE$_I$ is optimistic but inaccurate (∼2.3 of 7.5 ≈ 31%), whereas MuRE is both optimistic and accurate (∼1.1 of 1.5 ≈ 73%).

**Summary:** Our analysis identifies the best performing model per relation type as predicted by P1(b): multiplicative-only DistMult for type R, additive-multiplicative MuRE for types S/C; providing a basis for *dataset-dependent model selection*. The per-relation insight into where models perform poorly, e.g. hierarchical or type C relations, can be used to aid and direct future model design. Analysis of the classification performance: (i) shows that MuRE is the most reliable fact prediction model; and (ii) emphasises the poorer ability of multiplicative-only models to represent S/C relations.

Table 5: Per relation prediction accuracy for MuRE$_I$ (M$_I$), (D)istMult, (T)uckER and (M)uRE (WN18RR).

| Relation Name | Type | #$_{train}$ | #$_{test}$ | Accuracy (train) | | | | Accuracy (test) | | | | # Other "True" | | | |
|---|---|---|---|---|---|---|---|---|---|---|---|---|---|---|---|
| | | | | M$_I$ | D | T | M | M$_I$ | D | T | M | M$_I$ | D | T | M |
| verb_group | R | 15 | 39 | 1.00 | 1.00 | 1.00 | 1.00 | 0.97 | 0.97 | 0.97 | 0.97 | 8.3 | 1.7 | 0.9 | 2.7 |
| derivationally_related_form | R | 1714 | 1127 | 1.00 | 1.00 | 1.00 | 1.00 | 0.96 | 0.94 | 0.95 | 0.95 | 8.8 | 0.5 | 0.6 | 1.7 |
| also_see | R | 95 | 61 | 1.00 | 1.00 | 1.00 | 1.00 | 0.64 | 0.64 | 0.61 | 0.59 | 7.9 | 1.6 | 0.9 | 1.9 |
| instance_hypernym | S | 52 | 122 | 1.00 | 1.00 | 1.00 | 1.00 | 0.32 | 0.32 | 0.23 | 0.43 | 1.3 | 0.4 | 0.3 | 0.9 |
| member_of_domain_usage | C | 545 | 43 | 0.98 | 1.00 | 1.00 | 1.00 | 0.02 | 0.00 | 0.02 | 0.00 | 1.5 | 0.6 | 0.0 | 0.3 |
| member_of_domain_region | C | 543 | 42 | 0.88 | 0.89 | 1.00 | 0.93 | 0.02 | 0.02 | 0.00 | 0.02 | 1.0 | 0.4 | 0.8 | 0.7 |
| synset_domain_topic_of | C | 13 | 115 | 1.00 | 1.00 | 1.00 | 1.00 | 0.42 | 0.10 | 0.14 | 0.47 | 0.7 | 0.6 | 0.1 | 0.2 |
| member_meronym | C | 1402 | 307 | 1.00 | 1.00 | 1.00 | 1.00 | 0.22 | 0.02 | 0.01 | 0.22 | 7.9 | 3.4 | 1.5 | 5.6 |
| has_part | C | 848 | 196 | 1.00 | 1.00 | 1.00 | 1.00 | 0.24 | 0.05 | 0.09 | 0.22 | 7.1 | 2.4 | 1.3 | 3.9 |
| hypernym | S | 57 | 1254 | 1.00 | 1.00 | 1.00 | 1.00 | 0.15 | 0.02 | 0.02 | 0.22 | 3.7 | 1.2 | 0.0 | 1.7 |
| all | | 5284 | 3306 | 0.99 | 0.99 | 1.00 | 0.99 | 0.47 | 0.37 | 0.37 | 0.50 | 5.9 | 1.2 | 0.5 | 2.1 |

### 4.2 P2: Properties of relation representation

**P2(a)-(b):** Table 6 shows the symmetry score ($\in [-1, 1]$ indicating perfect anti-symmetry to symmetry; see Appx. F) for the relation matrix of TuckER and the norm of relation vectors of TransE, MuRE$_I$ and MuRE on the WN18RR dataset. As expected, type R relations have materially higher symmetry than both other relation types, fitting the prediction of how TuckER compensates for having no additive component. All additive models learn relation vectors of a noticeably lower norm for type R relations, which in the limiting case (similarity) require no additive component, than for types S or C.

**P2(c):** Fig 2 shows eigenvalue magnitudes (scaled relative to the largest and ordered) of relation-specific matrices $R$ of MuRE, labelled by relation type, as predicted to reflect the strength of a relation's *relatedness* aspect. As expected, values are highest for type R relations. For relation types S and C the profiles are more varied, supporting the understanding that relatedness of such relations is highly variable, both in its nature (components of $\mathcal{S}$) and strength (cardinality of $\mathcal{S}$).

Table 6: Relation matrix symmetry score [-1,1] for TuckER; and relation vector norm for TransE, MuRE$_I$ and MuRE (WN18RR).

| Relation | Type | Symmetry Score TuckER | Vector Norm | | |
|---|---|---|---|---|---|
| | | | TransE | MuRE$_I$ | MuRE |
| verb_group | R | 0.52 | 5.65 | 0.76 | 0.89 |
| derivationally_related_form | R | 0.54 | 2.98 | 0.45 | 0.69 |
| also_see | R | 0.50 | 7.20 | 0.97 | 0.97 |
| instance_hypernym | S | 0.13 | 18.26 | 2.98 | 1.88 |
| member_of_domain_usage | C | 0.10 | 11.24 | 3.18 | 1.88 |
| member_of_domain_region | C | 0.06 | 12.52 | 3.07 | 2.11 |
| synset_domain_topic_of | C | 0.12 | 23.29 | 2.65 | 1.52 |
| member_meronym | C | 0.12 | 4.97 | 1.91 | 1.97 |
| has_part | C | 0.13 | 6.44 | 1.69 | 1.25 |
| hypernym | S | 0.04 | 9.64 | 1.53 | 1.03 |

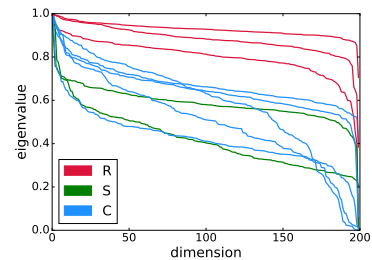

**Figure 2:** Eigenvalue magnitudes of relation-specific matrices $R$ for MuRE by relation type (WN18RR).

## 5 Conclusion

Many low-rank knowledge graph representation models have been developed, yet little is known of the latent structure they learn. We build on recent understanding of PMI-based word embeddings to theoretically establish a set of geometric properties of relation representations (relation conditions) required to map PMI-based word embeddings of subject entities to related object entities under knowledge graph relations. These conditions partition relations into three types and provide a basis to consider the loss functions of existing knowledge graph models. Models that satisfy the relation conditions of a particular type have a known set of model parameters that minimise the loss function, i.e. the parameters of PMI embeddings, together with potentially many equivalent solutions. We show that the better a model's architecture satisfies a relation's conditions, the better its performance at link prediction, evaluated under both rank-based metrics and accuracy. Overall, we generalise recent theoretical understanding of how particular semantic relations, e.g. similarity and analogy, are encoded between PMI-based word embeddings to the general relations of knowledge graphs. In doing so, we provide evidence in support of our initial premise: that common latent structure is exploited by both PMI-based word embeddings (e.g. W2V) and knowledge graph representation.

ACKNOWLEDGEMENTS

Carl Allen and Ivana Balažević were supported by the Centre for Doctoral Training in Data Science, funded by EPSRC (grant EP/L016427/1) and the University of Edinburgh.

## A  CATEGORISING WN18RR RELATIONS

Table 7 describes how each WN18RR relation was assigned to its respective category.

Table 7: Explanation for the WN18RR relation category assignment.

| Type | Relation | Relatedness | Subject Specifics | Object Specifics |
|---|---|---|---|---|
| R | verb_group | both verbs; similar in meaning | - | - |
|   | derivationally_related_form | different syntactic categories; semantically related | - | - |
|   | also_see | semantically similar | - | - |
| S | hypernym | semantically similar | instance of the object | - |
|   | instance_hypernym | semantically similar | instance of the object | - |
| C | member_of_domain_usage | loosely semantically related (word usage features) | usage descriptor | broad feature set |
|   | member_of_domain_region | loosely semantically related (regional features) | region descriptor | broad feature set |
|   | member_meronym | semantically related | collection of objects | part of the subject |
|   | has_part | semantically related | collection of objects | part of the subject |
|   | synset_domain_topic_of | semantically related | broad feature set | domain descriptor |

## B  CATEGORISING NELL-995 RELATIONS

Categorisation of NELL-995 relations and the explanation for the category assignment of are shown in Tables 8 and 9 respectively.

Table 8: Categorisation of NELL-995 relations.

| Type | Relation | Examples *(subject entity, object entity)* |
|---|---|---|
| R | teamplaysagainstteam | *(rangers, mariners), (phillies, tampa_bay_rays)* |
|   | clothingtogowithclothing | *(shirts, trousers), (shoes, black_shirt)* |
|   | agentcollaborateswithagent | *(white_stripes, jack_white), (barack_obama, hillary_clinton)* |
| S | professionistypeofprofession | *(trial_lawyers, attorneys), (engineers, experts)* |
|   | animalistypeofanimal | *(cats, small_animals), (chickens, livestock)* |
|   | chemicalistypeofchemical | *(moisture, gas), (oxide, materials)* |
| C | athleteplayssport | *(joe_smith, baseball), (chris_cooley, football)* |
|   | itemfoundinroom | *(bed, den), (refrigerator, kitchen_area)* |
|   | bodypartcontainsbodypart | *(system002, eyes), (blood, left_ventricle)* |
|   | atdate | *(scotland, n2009), (wto, n2003)* |
|   | locationlocatedwithinlocation | *(medellin, colombia), (jackson, wyoming)* |
|   | atlocation | *(ogunquin, maine), (palmer_lake, colorado)* |

Table 9: Explanation for the NELL-995 relation category assignment.

| Type | Relation | Relatedness | Subject Specifics | Object Specifics |
|---|---|---|---|---|
| R | teamplaysagainstteam | both sport teams | - | - |
|   | clothingtogowithclothing | both items of clothing that go together | - | - |
|   | agentcollaborateswithagent | both people or companies; related industries | - | - |
| S | professionistypeofprofession | semantically related (both profession types) | instance of the object | - |
|   | animalistypeofanimal | semantically related (both animals) | instance of the object | - |
|   | chemicalistypeofchemical | semantically related (both chemicals) | instance of the object | - |
| C | athleteplayssport | semantically related (sports features) | athlete descriptor | sport descriptor |
|   | itemfoundinroom | semantically related (room/furniture features) | item descriptor | room descriptor |
|   | bodypartcontainsbodypart | emantically related (specific body part features) | collection of objects | part of the subject |
|   | atdate | loosely semantically related (start date features) | broad feature set | date descriptor |
|   | locationlocatedwithinlocation | semantically related (geographical features) | part of the subject | collection of objects |
|   | atlocation | semantically related (geographical features) | part of the subject | collection of objects |

## C  SPLITTING THE NELL-995 DATASET

The test set of NELL-995 created by Xiong et al. (2017) contains only 10 out of 200 relations present in the training set. To ensure a fair representation of all training set relations in the validation and test sets, we create new validation and test set splits by combining the initial validation and test sets with the training set and randomly selecting 10,000 triples each from the combined dataset.

## D  IMPLEMENTATION DETAILS

All algorithms are re-implemented in PyTorch with the Adam optimizer (Kingma & Ba, 2015) that minimises binary cross-entropy loss, using hyper-parameters that work well for all models (learning rate: 0.001, batch size: 128, number of negative samples: 50). Entity and relation embedding dimensionality is set to $d_e = d_r = 200$ for all models except TuckER, for which $d_r = 30$ (Balažević et al., 2019b).

## E  KRACKHARDT HIERARCHY SCORE

The Krackhardt hierarchy score measures the proportion of node pairs $(x, y)$ where there exists a directed path $x \to y$, but not $y \to x$; and it takes a value of one for all directed acyclic graphs, and zero for cycles and cliques (Krackhardt, 2014; Balažević et al., 2019a).

Let $\boldsymbol{M} \in \mathbb{R}^{n \times n}$ be the binary *reachability matrix* of a directed graph $\mathcal{G}$ with $n$ nodes, with $\boldsymbol{M}_{i,j} = 1$ if there exists a directed path from node $i$ to node $j$ and 0 otherwise. The Krackhardt hierarchy score of $\mathcal{G}$ is defined as:

$$\text{Khs}_{\mathcal{G}} = \frac{\sum_{i=1}^{n} \sum_{j=1}^{n} \mathbb{1}(\boldsymbol{M}_{i,j} == 1 \wedge \boldsymbol{M}_{j,i} == 0)}{\sum_{i=1}^{n} \sum_{j=1}^{n} \mathbb{1}(\boldsymbol{M}_{i,j} == 1)}. \quad (1)$$

## F  SYMMETRY SCORE

The symmetry score $\in [-1, 1]$ (Hubert & Baker, 1979) for a relation matrix $\boldsymbol{R} \in \mathbb{R}^{d_e \times d_e}$ is defined as:

$$s = \frac{\sum \sum_{i \neq j} \boldsymbol{R}_{ij} \boldsymbol{R}_{ji} - \frac{(\sum \sum_{i \neq j} \boldsymbol{R}_{ij})^2}{d_e(d_e-1)}}{\sum \sum_{i \neq j} \boldsymbol{R}_{ij}^2 - \frac{(\sum \sum_{i \neq j} \boldsymbol{R}_{ij})^2}{d_e(d_e-1)}}, \quad (2)$$

where 1 indicates a symmetric and -1 an anti-symmetric matrix.

## G  "OTHER" PREDICTED FACTS

Tables 10 to 13 shows a sample of the unknown triples (i.e. those formed using the WN18RR entities and relations, but not present in the dataset) for the *derivationally_related_form* (R), *instance_hypernym* (S) and *synset_domain_topic_of* (C) relations at a range of probabilities $(\sigma(\phi(e_s, r, e_o)) \approx \{0.4, 0.6, 0.8, 1\})$, as predicted by each model. True triples are indicated in bold; instances where a model predicts an entity is related to itself are indicated in blue.

Table 10: "Other" facts as predicted by MuRE$_I$.

| Relation (Type) | $\sigma(\phi(e_s, r, e_o)) \approx 0.4$ | $\sigma(\phi(e_s, r, e_o)) \approx 0.6$ | $\sigma(\phi(e_s, r, e_o)) \approx 0.8$ | $\sigma(\phi(e_s, r, e_o)) \approx 1$ |
|---|---|---|---|---|
| derivationally_related_form (R) | (equalizer_NN_2, set_off_VB_5) (constellation_NN_2, satellite_NN_3) (**shrink_VB_3, subtraction_NN_2**) (continue_VB_10, proceed_VB_1) (support_VB_6, defend_VB_5) (shutter_NN_1, fill_up_VB_3) (yawning_NN_1, patellar_reflex_NN_1) (**yaw_NN_1, spiral_VB_1**) (stratum_NN_2, social_group_NN_1) (duel_VB_1, scuffle_NN_3) | (extrapolation_NN_1, maths_NN_1) (spread_VB_5, circularize_VB_3) (flaunt_NN_1, showing_NN_2) (**extrapolate_VB_3, synthesis_NN_3**) (strategist_NN_1, machination_NN_1) (crush_VB_4, grind_VB_2) (spike_VB_5, steady_VB_2) (licking_NN_1, vanquish_VB_1) (**synthetical_JJ_1, synthesizer_NN_2**) (realization_NN_2, embodiment_NN_3) | (sewer_NN_2, stitcher_NN_1) (lard_VB_1, vegetable_oil_NN_1) (**snuggle_NN_1, draw_close_VB_3**) (**train_VB_3, training_NN_1**) (**scratch_VB_3, skin_sensation_NN_1**) (scheme_NN_5, schematization_NN_1) (ordain_VB_3, vest_VB_1) (lie_VB_1, front_end_NN_1) (tread_NN_1, step_NN_9) (**register_NN_3, file_away_VB_1**) | (trail_VB_2, trail_VB_2) (**worship_VB_1, worship_VB_1**) (**steer_VB_1, steer_VB_1**) (**sort_out_VB_1, sort_out_VB_1**) (**make_full_VB_1, make_full_VB_1**) (utilize_VB_1, utilize_VB_1) (geology_NN_1, geology_NN_1) (**zoology_NN_2, zoology_NN_2**) (**uranology_NN_1, uranology_NN_1**) (travel_VB_1, travel_VB_1) |
| instance_hypernym (S) | (thomas_aquinas_NN_1, martyr_NN_2) (volcano_islands_NN_1, volcano_NN_2) (cape_horn_NN_1, urban_center_NN_1) (bergen_NN_1, national_capital_NN_1) (marshall_NN_2, generalship_NN_1) (**nansen_NN_1, venturer_NN_2**) (wisconsin_NN_2, state_capital_NN_1) (prussia_NN_1, stockade_NN_2) (**de_mille_NN_1, dancing-master_NN_1**) (aegean_sea_NN_1, aegean_island_NN_1) | (**taiwan_NN_1, asian_nation_NN_1**) (**st._gregory_of_n._NN_1, canonization_NN_1**) (st._gregory_of_n._NN_1, saint_VB_2) (mccormick_NN_1, find_VB_8) (**st._gregory_i_NN_1, bishop_NN_1**) (richard_buckminster_f._NN_1, technological_JJ_2) (thomas_aquinas_NN_1, archbishop_NN_1) (**marshall_NN_2, general_officer_NN_1**) (newman_NN_2, primateship_NN_1) (thomas_the_apostle_NN_1, sanctify_VB_1) | (**prophets_NN_1, gospels_NN_1**) (malcolm_x_NN_1, passive_resister_NN_1) (taiwan_NN_1, national_capital_NN_1) (truth_NN_2, abolitionism_NN_1) (**thomas_aquinas_NN_1, saint_VB_2**) (central_america_NN_1, s._am._nation_NN_1) (de_mille_NN_1, dance_NN_1) (st._gregory_i_NN_1, apostle_NN_3) (fertile_crescent_NN_1, asian_nation_NN_1) (robert_owen_NN_1, industry_NN_1) | (**helsinki_NN_1, urban_center_NN_1**) (mannheim_NN_1, stockade_NN_2) (**nippon_NN_1, nippon_NN_1**) (victor_hugo_NN_1, novel_NN_1) (regiomontanus_NN_1, uranology_NN_1) (**prophets_NN_1, book_NN_10**) (thomas_aquinas_NN_1, church_father_NN_1) (woody_guthrie_NN_1, minstrel_VB_1) (central_america_NN_1, c._am._nation_NN_1) (aegean_sea_NN_1, island_NN_1) |
| synset_domain_topic_of (C) | (write_VB_8, tape_VB_3) (introvert_NN_1, scientific_discipline_NN_1) (**libel_NN_1, slur_NN_2**) (etymologizing_NN_1, law_NN_1) (**temple_NN_4, place_of_worship_NN_1**) (trial_impression_NN_1, proof_VB_1) (friend_of_the_court_NN_1, war_machine_NN_1) (**multiv._analysis_NN_1, applied_math_NN_1**) (**sell_VB_1, transaction_NN_1**) (draw_VB_6, represent_VB_9) | (draw_VB_6, creative_person_NN_1) (suborder_NN_1, taxonomic_group_NN_1) (draw_VB_6, draw_VB_6) (**first_sacker_NN_1, ballgame_NN_2**) (alchemize_VB_1, modify_VB_3) (sermon_NN_1, sermon_NN_1) (**saint_VB_2, catholic_church_NN_1**) (male_JJ_1, masculine_JJ_2) (fire_VB_3, zoology_NN_2) (sell_VB_1, sell_VB_1) | (libel_NN_1, sully_VB_3) (relationship_NN_4, relationship_NN_4) (**etymologizing_NN_1, linguistics_NN_1**) (**turn_VB_12, cultivation_NN_2**) (brynhild_NN_1, mythologize_VB_2) (**brynhild_NN_1, myth_NN_1**) (**assist_NN_2, am._football_game_NN_1**) (mitzvah_NN_2, human_activity_NN_1) (drive_NN_12, drive_VB_8) (**relationship_NN_4, biology_NN_1**) | (**libel_NN_1, disparagement_NN_1**) (**roll-on_roll-off_NN_1, transport_NN_1**) (**prance_VB_4, equestrian_sport_NN_1**) (**libel_NN_1, traducement_NN_1**) (**sell_VB_1, selling_NN_1**) (trot_VB_2, ride_horseback_VB_1) (prance_VB_4, ride_horseback_VB_1) (gallop_VB_1, ride_horseback_VB_1) (**brynhild_NN_1, mythology_NN_2**) (**drive_NN_12, badminton_NN_1**) |

Table 11: "Other" facts as predicted by DistMult.

| Relation (Type) | $\sigma(\phi(e_s, r, e_o)) \approx 0.4$ | $\sigma(\phi(e_s, r, e_o)) \approx 0.6$ | $\sigma(\phi(e_s, r, e_o)) \approx 0.8$ | $\sigma(\phi(e_s, r, e_o)) \approx 1$ |
|---|---|---|---|---|
| derivationally_ related_form (R) | (stag_VB_3, undercover_work_NN_1) (print_VB_4, publisher_NN_2) (crier_NN_3, pitchman_NN_2) (play_VB_26, turn_NN_10) (count_VB_4, recite_VB_2) (vividness_NN_2, imbue_VB_3) (sea_mew_NN_1, larus_NN_1) (alkali_NN_2, acidify_VB_2) (see_VB_17, understand_VB_2) (shun_VB_1, hedging_NN_2) | (dish_NN_2, stew_NN_2) (expose_VB_3, show_NN_1) (system_NN_9, orderliness_NN_1) (spread_NN_4, strew_VB_1) (take_down_VB_2, put_VB_2) (wrestle_VB_4, wrestler_NN_1) (autotr._organism_NN_1, epiphytic_JJ_1) (duel_VB_1, slugfest_NN_1) (vocal_NN_2, rock_star_NN_1) (smelling_NN_1, scent_VB_1) | (shrink_NN_1, pedology_NN_1) (finish_VB_6, finishing_NN_2) (play_VB_26, playing_NN_3) (centralization_NN_1, unite_VB_6) (existence_NN_1, living_NN_3) (mouth_VB_3, sassing_NN_1) (constellation_NN_2, star_NN_1) (print_VB_4, publishing_house_NN_1) (puzzle_VB_2, secret_NN_3) (uranology_NN_1, tt_NN_1) | (alliterate_VB_1, versifier_NN_1) (geology_NN_1, structural_JJ_5) (reset_VB_1, amputation_NN_2) (nutrition_NN_3, man_NN_4) (saint_NN_3, sanctify_VB_1) (right_fielder_NN_1, leftfield_NN_1) (list_VB_4, slope_NN_2) (lieutenancy_NN_1, captain_NN_1) (tread_NN_1, step_VB_7) (exenteration_NN_1, enucleate_VB_2) |
| instance_ hypernym (S) | (wisconsin_NN_2, urban_center_NN_1) (marshall_NN_2, lieutenant_general_NN_1) (abidjan_NN_1, cote_d'ivoire_NN_1) (world_war_i_NN_1, urban_center_NN_1) (st._paul_NN_2, evangelist_NN_2) (deep_south_NN_1, urban_center_NN_1) (nuptse_NN_1, urban_center_NN_1) (ticino_NN_1, urban_center_NN_1) (aegean_sea_NN_1, aegean_island_NN_1) (cowpens_NN_1, war_of_am._ind._NN_1) | (mississippi_river_NN_1, american_state_NN_1) (r_e._byrd_NN_1, commissioned_officer_NN_1) (kobenhavn_NN_1, urban_center_NN_1) (the_gambia_NN_1, africa_NN_1) (tirich_mir_NN_1, urban_center_NN_1) (r_e._byrd_NN_1, military_advisor_NN_1) (tampa_bay_NN_1, urban_center_NN_1) (tidewater_region_NN_1, south_NN_1) (r_e._byrd_NN_1, executive_officer_NN_1) | (deep_south_NN_1, south_NN_1) (capital_of_gambia_NN_1, urban_center_NN_1) (south_west_africa_NN_1, africa_NN_1) (brandenburg_NN_1, urban_center_NN_1) (sierra_nevada_NN_1, urban_center_NN_1) (malcolm_x_NN_1, emancipationist_NN_1) (north_platte_river_NN_1, urban_center_NN_1) (oslo_NN_1, urban_center_NN_1) (zaire_river_NN_1, urban_center_NN_1) (transylvanian_alps_NN_1, urban_center_NN_1) | (helsinki_NN_1, urban_center_NN_1) (the_nazarene_NN_1, save_VB_7) (irish_capital_NN_1, urban_center_NN_1) (r_e._byrd_NN_1, inspector_general_NN_1) (r_e._byrd_NN_1, chief_of_staff_NN_1) (central_america_NN_1, c._am._nation_NN_1) (malcolm_x_NN_1, environmentalist_NN_1) (the_nazarene_NN_1, christian_JJ_1) (thomas_aquinas_NN_1, church_father_NN_1) (the_nazarene_NN_1, el_nino_NN_2) |
| synset_domain_ topic_of (C) | (limitation_NN_4, trammel_VB_2) (light_colonel_NN_1, colonel_NN_1) (nurse_VB_1, nursing_NN_1) (sermon_NN_1, prophesy_VB_2) (libel_NN_1, practice_of_law_NN_1) (slugger_NN_1, baseball_player_NN_1) (rna_NN_1, chemistry_NN_1) (metrify_VB_1, versify_VB_1) (trial_impression_NN_1, publish_VB_1) (turn_VB_12, plowman_NN_1) | (roll-on_roll-off_NN_1, transport_NN_1) (hizb_ut-tahrir_NN_1, asia_NN_1) (slugger_NN_1, softball_game_NN_1) (sermon_NN_1, sermonize_VB_1) (draw_VB_6, drawer_NN_3) (turn_VB_12, plow_NN_1) (assist_NN_2, softball_game_NN_1) (council_NN_2, assembly_NN_4) (throughput_NN_1, turnout_NN_4) (cream_VB_1, cream_NN_2) | (etymologizing_NN_1, explanation_NN_1) (ferry_VB_3, travel_VB_1) (public_prosecutor_NN_1, prosecute_VB_2) (alchemize_VB_1, modify_VB_3) (libel_NN_1, libel_VB_1) (turn_VB_12, till_VB_1) (hit_NN_1, hit_VB_1) (fire_VB_3, flaming_NN_1) (ring_NN_4, chemical_chain_NN_1) (ibidinal_energy_NN_1, charge_NN_9) | (flat_JJ_5, matte_NN_2) (etymologizing_NN_1, derive_VB_3) (hole_out_VB_1, hole_NN_3) (relationship_NN_4, relation_NN_1) (drive_NN_12, badminton_NN_2) (etymologizing_NN_1, etymologize_VB_2) (matrix_algebra_NN_1, diagonalization_NN_1) (cabinetwork_NN_2, woodworking_NN_1) (cabinetwork_NN_2, bottom_NN_1) (cabinetwork_NN_2, upholster_VB_1) |

Table 12: "Other" facts as predicted by TuckER.

| Relation (Type) | $\sigma(\phi(e_s, r, e_o)) \approx 0.4$ | $\sigma(\phi(e_s, r, e_o)) \approx 0.6$ | $\sigma(\phi(e_s, r, e_o)) \approx 0.8$ | $\sigma(\phi(e_s, r, e_o)) \approx 1$ |
|---|---|---|---|---|
| derivationally_ related_form (R) | (tympanist_NN_1, gong_NN_2) (indication_NN_1, signalize_VB_2) (turn_over_VB_3, rotation_NN_3) (date_VB_5, geological_dating_NN_1) (set_VB_23, emblem_NN_2) (tyro_NN_1, start_VB_5) (identification_NN_1, name_VB_5) (stabber_NN_1, thrust_VB_5) (justification_NN_1, apology_NN_2) (manufacture_VB_1, prevarication_NN_1) | (mash_NN_2, mill_VB_2) (walk_VB_9, zimmer_frame_NN_1) (use_VB_5, utility_NN_2) (musical_instrument_NN_1, write_VB_6) (lining_NN_3, wrap_up_VB_1) (scrap_VB_2, struggle_NN_2) (tape_VB_3, tape_recorder_NN_1) (vindicate_VB_2, justification_NN_2) (leaching_NN_1, percolate_VB_3) (synchronize_VB_2, synchroscope_NN_1) | (take_chances_VB_1, venture_NN_1) (shutter_NN_1, fill_up_VB_3) (exit_NN_3, leave_VB_1) (trembler_NN_1, vibrate_VB_1) (motivator_NN_1, trip_VB_4) (support_VB_6, indorsement_NN_1) (federate_VB_2, confederation_NN_1) (take_over_VB_6, return_NN_2) (wait_on_VB_1, supporter_NN_3) (denote_VB_3, promulgation_NN_1) | (venturer_NN_2, venturer_NN_2) (dynamitist_NN_1, dynamitist_NN_1) (love_VB_3, lover_NN_2) (snuggle_NN_1, squeeze_VB_8) (departed_NN_1, die_VB_2) (position_VB_1, placement_NN_1) (repentant_JJ_1, repentant_JJ_1) (tread_NN_1, step_VB_7) (stockist_NN_1, stockist_NN_1) (philanthropist_NN_1, philanthropist_NN_1) |
| instance_ hypernym (S) | (deep_south_NN_1, south_NN_1) (st_paul_NN_2, organist_NN_1) (helsinki_NN_1, urban_center_NN_1) (malcolm_x_NN_1, emancipationist_NN_1) (thomas_the_apostle_NN_1, church_father_NN_1) (st_gregory_of_n._NN_1, sermonizer_NN_1) (robert_owen_NN_1, movie_maker_NN_1) (theresa_NN_1, monk_NN_1) (st_paul_NN_2, philosopher_NN_1) (ibn-roshd_NN_1, pedagogue_NN_1) | (thomas_aquinas_NN_1, bishop_NN_1) (irish_capital_NN_1, urban_center_NN_1) (thomas_the_apostle_NN_1, apostle_NN_2) (st_paul_NN_2, apostle_NN_3) (mccormick_NN_1, painter_NN_1) (thomas_the_apostle_NN_1, troglodyte_NN_1) (mccormick_NN_1, electrical_engineer_NN_1) (mississippi_river_NN_1, american_state_NN_1) | (cowpens_NN_1, siege_NN_1) (mccormick_NN_1, arms_manufacturer_NN_1) (thomas_the_apostle_NN_1, evangelist_NN_1) (mccormick_NN_1, technologist_NN_1) (st_gregory_i_NN_1, church_father_NN_1) | (r_e_byrd_NN_1, siege_NN_1) (shaw_NN_3, women's_rightist_NN_1) (aegean_sea_NN_1, aegean_island_NN_1) (thomas_aquinas_NN_1, church_father_NN_1) |
| synset_domain_ topic_of (C) | (roll-on_roll-off_NN_1, motorcar_NN_1) (libel_NN_1, legislature_NN_1) (roll-on_roll-off_NN_1, passenger_vehicle_NN_1) | (drive_NN_12, badminton_NN_1) | | |

Table 13: "Other" facts as predicted by MuRE.

| Relation (Type) | $\sigma(\phi(e_s, r, e_o)) \approx 0.4$ | $\sigma(\phi(e_s, r, e_o)) \approx 0.6$ | $\sigma(\phi(e_s, r, e_o)) \approx 0.8$ | $\sigma(\phi(e_s, r, e_o)) \approx 1$ |
|---|---|---|---|---|
| derivationally_related_form (R) | (surround_VB_1, wall_NN_1)<br>(unpleasant_JJ_1, unpalatableness_NN_1)<br>(love_VB_3, enjoyment_NN_2)<br>(magnitude_NN_1, tall_JJ_1)<br>(testify_VB_2, information_NN_1)<br>(connect_VB_6, converging_NN_1)<br>(connect_VB_6, connexion_NN_4)<br>(operate_VB_4, psyop_NN_1)<br>(market_VB_1, trade_NN_4)<br>(operate_VB_4, mission_NN_2) | (word_picture_NN_1, sketch_VB_2)<br>(develop_VB_10, adjustment_NN_4)<br>(gloss_VB_3, commentary_NN_1)<br>(violate_VB_2, violation_NN_3)<br>(suffocate_VB_1, strangler_tree_NN_1)<br>(number_VB_3, point_NN_12)<br>(develop_VB_10, organic_process_NN_1)<br>(plication_NN_1, twist_VB_4)<br>(split_up_VB_3, separation_NN_5)<br>(plication_NN_1, wrinkle_VB_2) | (smelling_NN_1, wind_VB_4)<br>(try_out_VB_1, somatic_cell_nuclear_transplantation_NN_1)<br>(lighting_NN_4, set_on_fire_VB_1)<br>(symphalangus_NN_1, one-half_NN_1)<br>(just_JJ_3, validity_NN_1)<br>(reprove_VB_1, talking_to_NN_1)<br>(sustain_VB_5, beam_NN_2)<br>(spring_NN_6, hurdle_VB_1)<br>(spark_NN_1, scintillate_VB_1)<br>(utility_NN_2, functional_JJ_1) | (spoliation_NN_2, sack_VB_1)<br>(desire_NN_2, hope_VB_2)<br>(snuffle_VB_3, whine_NN_1)<br>(nasalization_NN_1, sound_out_VB_1)<br>(tread_NN_1, step_VB_7)<br>(yearn_VB_1, pining_NN_1)<br>(unreliableness_NN_1, arbitrary_JJ_1)<br>(travesty_NN_2, travesty_NN_2)<br>(spark_NN_1, sparkle_VB_1)<br>(stockist_NN_1, stockist_NN_1) |
| instance_hypernym (S) | (malcolm_x_NN_1, hipster_NN_1)<br>(the_nazarene_NN_1, judaism_NN_2)<br>(old_line_state_NN_1, river_NN_1)<br>(r_e_byrd_NN_1, commissioned_officer_NN_1)<br>(south_korea_NN_1, peninsula_NN_1)<br>(st_gregory_of_n_NN_1, vicar_of_christ_NN_1)<br>(nippon_NN_1, italian_region_NN_1)<br>(robert_owen_NN_1, tycoon_NN_1)<br>(mandalay_NN_1, national_capital_NN_1)<br>(nan_ling_NN_1, urban_center_NN_1) | (central_america_NN_1, central_america_NN_1)<br>(st_gregory_i_NN_1, church_father_NN_1)<br>(south_korea_NN_1, african_nation_NN_1)<br>(malcolm_x_NN_1, passive_resister_NN_1)<br>(malcolm_x_NN_1, birth-control_reformer_NN_1)<br>(los_angeles_NN_1, port_NN_1)<br>(great_lakes_NN_1, canadian_province_NN_1)<br>(transylvanian_alps_NN_1, urban_center_NN_1)<br>(gettysburg_NN_2, siege_NN_1)<br>(wisconsin_NN_2, geographical_region_NN_1) | (theresa_NN_1, monk_NN_1)<br>(nippon_NN_1, european_nation_NN_1)<br>(great_lakes_NN_1, river_NN_1)<br>(r_e_byrd_NN_1, noncommissioned_officer_NN_1)<br>(world_war_i_NN_1, pitched_battle_NN_1)<br>(irish_capital_NN_1, urban_center_NN_1)<br>(volcano_islands_NN_1, urban_center_NN_1)<br>(nippon_NN_1, american_state_NN_1)<br>(helsinki_NN_1, urban_center_NN_1)<br>(capital_of_gambia_NN_1, urban_center_NN_1) | |
| synset_domain_topic_of (C) | (libel_NN_1, criminal_law_NN_1)<br>(brynhild_NN_1, mythology_NN_2)<br>(slugger_NN_1, sport_NN_1)<br>(sell_VB_1, law_NN_1)<br>(semitic_deity_NN_1, mythology_NN_1)<br>(nuclear_deterrence_NN_1, law_NN_1)<br>(reception_NN_5, baseball_game_NN_1)<br>(photosynthesis_NN_1, chemistry_NN_1)<br>(isolde_NN_1, parable_NN_1)<br>(assist_NN_2, court_game_NN_1) | (write_VB_8, transcription_NN_5)<br>(temple_NN_4, muslimism_NN_2)<br>(assist_NN_2, hockey_NN_1)<br>(relationship_NN_4, biology_NN_1)<br>(apostle_NN_3, western_church_NN_1)<br>(assist_NN_2, sport_NN_1)<br>(trot_VB_2, equestrian_sport_NN_1)<br>(rna_NN_1, chemistry_NN_1)<br>(assist_NN_2, soccer_NN_1)<br>(assist_NN_2, football_game_NN_1) | (assist_NN_2, am._football_game_NN_1)<br>(drive_NN_12, court_game_NN_1)<br>(sell_VB_1, offense_NN_3)<br>(slugger_NN_1, softball_game_NN_1)<br>(drive_NN_12, badminton_NN_1) | |

## 5.3 Impact

According to Google Scholar, the paper has received 13 citations as of August 2021, including from a recent survey of knowledge graphs and "explainable AI" (Bianchi et al., 2020).

## 5.4 Discussion

*Interpreting KGs* presents a model for the latent semantic structure of knowledge graph representation. In doing so, the paper brings together representations of words and entities learned from text corpora and knowledge graphs. This is appealing since it might be expected that representations of words, or the concepts they themselves represent, are independent of the learning method. Where, previously, we noted parallels between word embeddings and knowledge graph representations, in particular between analogy and knowledge graph relations (§3.1), *Interpreting KGs* develops this into a unifying theoretical model for knowledge graph representations and word embeddings.

Although the latent structure of knowledge graph representation models is analysed by consideration of PMI-based embeddings, we re-emphasise that we do not expect that KGR models learn PMI statistics. Instead, by deriving geometric relationships between PMI vectors of entities that satisfy different semantic relation types, we identify latent structure that can be learned by a suitably designed model. This might be seen analogously to first theoretically determining that data is, say, linear or Gaussian and then applying a linear regression model or principal component analysis, respectively, because those models are known to "fit" such data.

*Analogies Explained* shows that the vector offset method captures analogies involving a particular type of semantic relation, but several indications have suggested it may not represent *all* semantic relations: the 1-to-1 nature of the vector offset method; the presence of multiple relations between certain pairs of words; the variable performance of the vector offset method across different relation types; and the outperformance of *TransE* by other KGR models (§4.4). While the minimal data available to learn analogy relations – a single related word pair $a$, $a^*$ – restrict options for representing them to relatively simple functions, such as the vector offset, having multiple related entity pairs allows more expressive parameterised functions to be learned, e.g. for knowledge graphs. *Interpreting KGs* offers the first explanation based on latent semantic structure for why the vector offset is indeed insufficient for modelling all relations. The paper illustrates how embeddings relate to one another for different types of semantic relation, showing where the vector offset fits within a hierarchy of relation types.

The paper culminates in justifying why the *additive and multiplicative* knowledge graph representation model *MuRE* (Balažević et al., 2019b) outperforms models that are strictly additive *or* multiplicative. We note that this justification is not retrospective, rather the *MuRE* model was developed (by the same authors) contemporaneously with *Interpreting KGs*, inspired by early theoretical insight. As such, *Interpreting KGs* can be seen to justify the performance of *MuRE*, or the *MuRE* model can be viewed as a practical implementation and empirical validation of the paper.

We note several limitations of *Interpreting KGs*, which may be fruitful to address in future work:

- the paper considers relationships between PMI vectors based on co-occurrence

   statistics that use a fixed context window, abstracting away word order and failing to account for this likely important information;

- although the paper implicitly attributes semantic meaning to components of relation representations, e.g. the vector offset and the common subspace of related embeddings, it does not explore this explicitly or verify it empirically; and

- although the paper evaluates the performance of several recent competitive knowledge graph representation models, there are many other KGR models, some of which may potentially have entirely different and unexplained rationale.

# Chapter 6

# Conclusion

This thesis presents three works that develop our understanding of the latent semantic space that underpins well-known word embedding algorithms and knowledge graph representation models.

- *Analogies Explained* (§3) provides a mathematical justification of why word embeddings learned by algorithms such as SGNS and GloVe can be used to "solve analogies" by taking linear combinations corresponding to parallelograms. The result follows from recognising that such word embeddings are low rank projections of PMI vectors; and that PMI vectors of analogies form parallelograms subject to semantically interpretable error terms.

- *What the Vec* (§4) develops an understanding of the space of PMI vectors, considers more fully the semantic relationships of similarity, relatedness, analogy and paraphrase and their inter-connection, and justifies previous empirical findings such as the improved performance of *average* embeddings $\frac{\boldsymbol{w}_i + \boldsymbol{c}_i}{2}$.

- *Interpreting KGs* (§5) extends the correspondence between embedding geometry and semantics for certain relation types (similarity, paraphrases, analogies) to include the specific relations of knowledge graphs, enabling geometric properties of relation representations to be derived and justifying the relative per-relation performance of a range of knowledge graph representation models.

Throughout, we have drawn parallels between (a) word embeddings and semantic relationships between them, such as similarity and analogies, and (b) the representations of entities and relations of knowledge graphs. Although these two paradigms are often treated separately, the same words/entities and relations may appear in a text corpus or knowledge graph, hence it seems both intuitive and useful to represent them in a common way since representations may then be learned jointly and utilised more widely. This thesis takes positive steps, providing theoretical and empirical support for the premise that common semantic structure can underpin word embeddings and knowledge graph representations.

While some recent works question whether the vector offset method does or does not represent analogies (Rogers et al., 2017; Schluter, 2018), we find that the answer is nuanced since analogies are not homogeneous. Some analogies may be well represented by a vector offset, others less so and some very poorly, justifying the observed variable performance in analogical reasoning for different semantic relations. *Analogies Explained* describes the positive cases, i.e. those semantic relations that are represented well by a vector offset; *Interpreting KGs* extends that to more general semantic relations, where it can be seen that the vector offset alone is insufficient. For analogies, however, the options for representing relations are highly limited due to the availability of only one training instance. Relation representations cannot easily be parameterised on a bespoke basis without over-fitting, hence it may be that some other (unparameterised) function outperforms the vector offset where the latter is insufficient (averaged over a given dataset). A better function might also account for known trends in error terms (e.g. Paperno and Baroni, 2016). As such, future work may improve on the vector offset method or explain heuristics such as 3CosMul (Levy et al., 2015).

Having linked word embedding and knowledge graph representation, future algorithms might be developed that learn word/entity embeddings *jointly* from text corpora *and* knowledge graphs in a principled manner. This is appealing since the data sources complement one another: text is abundant but its co-occurrence statistics are noisy, whereas knowledge graph data is largely accurate but more difficult to acquire. Bringing together text and knowledge graphs may also lead to improved methods for *extracting relations from text*.

We hope that future work can build more generally on the insights into latent semantic structure presented in this thesis to develop algorithms that perform better, offer greater interpretability and allow unwanted statistical biases in the data to be mitigated. Future work may also extend the understanding of *un-contextualised* embeddings to *contextualised* word embeddings that presently achieve impressive performance in many downstream NLP tasks (Devlin et al., 2019; Brown et al., 2020) and may become increasingly pervasive across numerous applications. Such models contain vast numbers of parameters, requiring significant time and energy to train. A clearer understanding of what such parameters learn may allow more succinct models to be developed and their limitations, such as biases picked up from the data, to be understood and potentially mitigated.